\documentclass{article}

\usepackage[main, final]{neurips_2026}

\usepackage[utf8]{inputenc}
\usepackage[T1]{fontenc}
\usepackage{hyperref}
\usepackage{url}
\usepackage{booktabs}
\usepackage{amsfonts}
\usepackage{nicefrac}
\usepackage{microtype}
\usepackage{framed}
\usepackage{xcolor}

\usepackage[utf8]{inputenc} 
\usepackage[T1]{fontenc}    
\usepackage{hyperref}       
\usepackage{url}            
\usepackage{booktabs}       
\usepackage{amsfonts}       
\usepackage{nicefrac}       
\usepackage{microtype}      
\usepackage{xcolor}         
\usepackage{colortbl}

\usepackage{amsthm}
\theoremstyle{plain}

\theoremstyle{definition}

\theoremstyle{remark}

\usepackage[flushleft]{threeparttable}

\title{Plausibility Is Not Prediction: Contrastive Evidence for LLM-Based Cellular Perturbation Reasoning}

\author{%
  Xinyu Yuan\textsuperscript{1,2,*}, Xixian Liu\textsuperscript{1,2,*}, Jianan Zhao\textsuperscript{4,5}, Yashi Zhang\textsuperscript{1,2}, Hongyu Guo\textsuperscript{4,5}, Jian Tang\textsuperscript{1,3,6,\#}  \\
  \textsuperscript{1}{Mila - Qu\'ebec AI Institute} \textsuperscript{2}{University of Montr\'eal} \textsuperscript{3}{HEC Montr\'eal}\\ \textsuperscript{4}{University of Ottawa} \textsuperscript{5}{National Research Council of Canada} \textsuperscript{6}{CIFAR AI Chair}\\
\textsuperscript{*}{Equal contribution.} 
\textsuperscript{\#}{Correspondence to: tangjian@mila.quebec}
}

\usepackage{xspace}
\usepackage{minted}
\usepackage{wrapfig}
\usepackage{caption}
\usepackage{amssymb}
\usepackage{enumitem}
\usepackage{multirow}
\usepackage{adjustbox}
\usepackage{algorithm}
\usepackage{subcaption}
\usepackage{algpseudocode}


\newcommand{\corereasoning}{CORE-Reasoning\xspace}
\newcommand{\corevoting}{CORE-Voting\xspace}
\newcommand{\geneprior}{Gene-prior Voting\xspace}
\newcommand{\core}{CORE\xspace}

\definecolor{visualization_blue}{rgb}{0.00784313725490196, 0.24313725490196078, 1.0}
\definecolor{visualization_orange}{rgb}
{1.0, 0.48627450980392156, 0.0}

\definecolor{visualization_green}{rgb}{0.10196078431372549, 0.788235294117647, 0.2196078431372549}
 
\definecolor{visualization_red}{rgb}{0.9098039215686274, 0.0, 0.043137254901960784}
 
\definecolor{visualization_purple}{rgb}{0.5450980392156862, 0.16862745098039217, 0.8862745098039215}
 
\definecolor{visualization_brown}{rgb}{0.6235294117647059, 0.2823529411764706, 0.0}
 
\definecolor{visualization_pink}{rgb}{0.9450980392156862, 0.2980392156862745, 0.7568627450980392}
 
\definecolor{visualization_grey}{rgb}{0.6392156862745098, 0.6392156862745098, 0.6392156862745098}
 
\definecolor{visualization_yellow}{rgb}{1.0, 0.7686274509803922, 0.0}

\algnewcommand{\LineComment}[1]{\State \(\triangleright\) #1}

\newcommand{\multilinecomment}[1]{}

\usepackage{amsmath,amsfonts,bm}

\def\eqref#1{equation~\ref{#1}}

\def\1{\bm{1}}

\DeclareMathAlphabet{\mathsfit}{\encodingdefault}{\sfdefault}{m}{sl}
\SetMathAlphabet{\mathsfit}{bold}{\encodingdefault}{\sfdefault}{bx}{n}

\begin{document}

\maketitle

\vspace{-14pt}
\begin{abstract}

Perturbation experiments are central to understanding cellular mechanisms, but remain costly and sparse, motivating prediction of gene expression responses for unobserved conditions. 
A promising recent direction leverages large language models (LLMs) as “virtual cell” simulators—using stepwise, knowledge-grounded mechanistic reasoning to infer differential expression—pointing toward an interpretable, knowledge-driven paradigm that transcends purely data-driven approaches. 
However, we find that plausibility is not prediction: despite producing biologically plausible explanations, these methods fail to capture perturbation-specific effects: systematically overestimating differential expression, often underperforming  a simple gene‑frequency baseline in aggregate evaluations, and collapsing to chance-level performance at the per-gene level. This reveals a reliance  on intrinsic gene response tendencies rather than true perturbation reasoning.
We trace this failure to how evidence is presented: existing methods evaluate perturbation–gene pairs in isolation, without exposing how related perturbations differ in their effects on the same gene. 
To address this limitation, 
we introduce \textbf{CORE} (\textbf{C}ontrastive \textbf{O}rganization of \textbf{R}elational \textbf{E}vidence), which reframes prediction as a comparison task by organizing evidence into positive and negative outcomes from related perturbations. Using a biomedical knowledge graph for evidence retrieval, CORE improves calibration and substantially boosts  perturbation-specific prediction in both LLM-based and non-LLM settings: for example, on drug-perturbation data, {\corereasoning} improves Qwen3.5-9B aggregate metrics
by up to 28.6\%, while on generic perturbation data, {\corevoting} raises macro-per-gene
AUROC from chance to 0.703 in average across four cell lines. This highlights contrastive evidence organization as essential  to reliable LLM-based perturbation reasoning. 


\end{abstract}

\vspace{-12pt}
\section{Introduction}
\vspace{-5pt}
\if false
Perturbation experiments, including drug treatment and genetic editing, are important for studying cellular mechanisms~\cite{dixit2016perturbseq,adamson2016multiplexed,lamb2006connectivity}. 
By measuring cellular responses after a defined intervention, these experiments provide a way to connect perturbations with downstream molecular changes. 
Differential expression (DE) is one of the most widely used molecular readouts in such studies: genes that are significantly up- or down-regulated after perturbation can support target discovery, mechanism-of-action hypothesis generation, and drug repurposing~\cite{lamb2006connectivity,ni2024pertkge}.

However, experimental cost and finite budgets limit perturbation profiling to only a small fraction of possible perturbations and cellular contexts.  
In practice, observed outcomes are therefore often used to prioritize which perturbations should be tested next. 
This makes it important to predict transcriptional responses for perturbations that have not been experimentally profiled~\cite{lotfollahi2023predicting}~\cite{roohani2024predicting}. 
\fi 

Perturbation experiments, including drug treatments and genetic editing, are central to understanding cellular mechanisms~\cite{adamson2016multiplexed,dixit2016perturbseq,lamb2006connectivity}. By measuring gene expression changes after controlled interventions, these experiments link perturbations to downstream molecular responses. However, their cost and scale limit coverage across the vast space of perturbations and cellular contexts, motivating  prediction of transcriptional responses for unobserved conditions, a modeling task traditionally addressed by purely data-driven methods  that learn statistical associations from observed data~\cite{cui2024scgpt,lotfollahi2023predicting,roohani2024predicting}. 

\if false
\textcolor{red}{Mention popular virtual cell modeling approaches here, and explain why LLMs stand out and have significant potential.}
Recent LLM-based methods formulate perturbation prediction as a natural-language reasoning task~\cite{wu2025perturbqa}~\cite{wei2026vcworld}. Given a cell line $c$, a perturbation $p$, and a gene $g$, the model predicts whether the gene should be differentially expressed, often with a generated biological rationale reasoning chains. 
This paradigm is attractive because much biological knowledge is semantic and relational: gene functions, drug mechanisms, pathway memberships, disease associations, and observed experimental DE outcomes are often described in text or structured knowledge resources. LLMs therefore provide a flexible way to condition predictions on such evidence and to reason about perturbations with limited direct DE measurement outcomes.
\fi 

In contrast, recent work  explores large language models (LLMs) as ``virtual cell'' simulators~\cite{wei2026vcworld,wu2025perturbqa,zhang2025cellverse}, recasting perturbation prediction as a natural-language reasoning task. Given a cell line $c$, a perturbation $p$, and a gene $g$, these models are prompted to predict differential expression with  biological rationales. This paradigm is particularly appealing because biological knowledge—such as gene functions, drug mechanisms, and pathway relationships—is inherently semantic and relational, aligning naturally with LLM reasoning~\cite{chandak2023building,himmelstein2017systematic,wei2026vcworld}. Consequently, LLMs offer an interpretable, knowledge-driven framework with the potential to  transcend purely data-driven methods. 

\if false
However, we observe that current LLM perturbation reasoning methods fail to effectively understand the observed experimental DE outcomes and can hardly capture the data's distribution from two perspectives. First, under standard binary prompting, LLMs show systematic tendency to over-predict differential expression: their predicted yes-rate is much higher than the empirical positive rate in the data (Fig.~\ref{}).
Second, as shown in Fig.~\ref{}, we find that LLMs can be substantially outperformed by a simple gene-frequency predictor (\textbf{{\geneprior}}), which uses only how often each gene is differentially expressed in the observed training data.

We note this biological realistic patten behind the perturbation reasoning task formulation: many genes exhibit stable response tendencies across perturbations: some genes are differentially expressed under many conditions, while others rarely respond. This leads to the seemingly "success" for {\textbf{{\geneprior}}}, even though this predictor ignores perturbation identity entirely and its performance does not indicate perturbation reasoning. We therefore expand the evaluation protocol with per-gene metrics in addition to the aggregate metrics, to measure the more useful capability--perturbation-specific discrimination--which determins which perturbations affect a given gene after accounting for its background response tendency. However, as shown in Fig.~\ref{}, both \textbf{{\geneprior}} and current LLMs reasoning methods stays at random for per-gene metrics (AUROC ~0.5).

In this work, we not only diagnose this systematic failure for current LLM perturbation reasoning methods, we also propose a fundamental insight that, the main bottleneck is not the LLM model capacity, but how observed DE evidence is organized before prediction.

We propose to address this bottleneck through contrastive evidence organization.
Current LLM prompting pipelines often ask the model to judge whether a perturbation--gene pair is biologically plausible in isolation, and in-context evidences are just retrieved by textual or KG-structural proximity and listed in order without clarifying their internal relationships. 
In such prompts, the model is not explicitly shown the empirical response tendency of the target gene, nor contrasting cases where related perturbations did and did not affect that same gene. 
We instead organize observed DE evidence around support cases that are conditional on the target gene. 
For each test triple $(c,p,g)$, we retrieve training perturbations that are relationally close to $p$ and have observed response labels for the same gene $g$. 
We then separate these retrieved examples into positive and negative support cases. 
The prediction is no longer based only on whether $(p,g)$ appears biologically plausible, but on how differently these related perturbations affected $g$ in observed experiments by comparison.

We call this evidence organization procedure \textbf{CORE}, for \textbf{Contrastive Organization of Relational Evidence}. 
CORE turns isolated pairwise prediction into cross-perturbation comparison. 
Because relationally similar perturbations can still produce different transcriptional outcomes, CORE does not assume that related perturbations should always share the same label. 
Instead, it explicitly presents both positive and negative support cases when available, allowing the model to compare the test perturbation against observed perturbations with known outcomes for the same target gene. 
This design anchors prediction in empirical support evidence and makes perturbation-specific discrimination more explicit.

To instantiate CORE, we construct a biomedical knowledge graph for perturbation reasoning, \textbf{ReasonKG}. 
ReasonKG harmonizes entities and relations among perturbations, genes, diseases, pathways, and biological processes, and improves the coverage of benchmark perturbations and genes compared with previous resources (Sec.~\ref{}). 
The graph is used as a retrieval structure rather than as a prediction target by itself: it provides interpretable relational paths between perturbations and genes, and these paths serve as mechanism-relevant evidence for organizing support cases. 
This structure allows CORE to retrieve perturbations that are biologically related to the test perturbation while grounding the final decision in observed response labels for the target gene.

We demonstrate CORE evidence's effectiveness in two complementary ways. 
\textbf{{\corevoting}} is a lightweight non-LLM classifier that directly aggregates the observed labels of graph-organized support cases for the target gene. 
It demonstrates whether structured support evidence from ReasonKG and training outcomes contains perturbation-specific signal, independent of LLM reasoning. 
\textbf{{\corereasoning}} provides the same contrastive support cases to an LLM as in-context examples. 
By exposing positive and negative outcomes for related perturbations, {\corereasoning} guides the LLM away from isolated plausibility judgments and toward evidence-grounded comparison. 
Compared with prior LLM reasoning pipelines, {\corereasoning} reduces positive-rate miscalibration and improves perturbation-specific prediction under per-gene evaluation.

Our contributions are as follows. 
First, we identify two empirical failure modes in representative LLM-based perturbation reasoning pipelines: positive-rate miscalibration and insufficient control for gene-level response frequency. 
Second, we construct \textbf{ReasonKG}, a harmonized biomedical knowledge graph that improves perturbation and gene grounding coverage and provides relational paths for evidence retrieval. 
Third, we propose \textbf{CORE}, a contrastive evidence organization procedure that conditions prediction on same-gene support cases from related perturbations. 
Fourth, we introduce \textbf{{\corevoting}} and \textbf{{\corereasoning}}, showing that graph-organized contrastive support evidence improves perturbation-specific differential-expression prediction beyond gene-frequency shortcuts.

\fi 

Despite this promise, current LLM-based methods~\cite{wei2026vcworld,wu2025perturbqa,zhang2025cellverse} fail to capture perturbation-specific effects. They systematically overestimate differential expression (Fig.~\ref{fig:yes_bias_c32}), often underperform simple gene-frequency and support-voting baselines under aggregate metrics (Figs.~\ref{fig:c32_core_voting_sweep} and~\ref{fig:c32_llm_metrics}), and collapse toward chance-level performance at the per-gene level (Fig.~\ref{fig:c32_llm_metrics}). These findings reveal reliance on intrinsic gene response tendencies rather than true perturbation reasoning.

This failure arises from how evidence is presented during reasoning. Specifically, existing methods evaluate perturbation–gene pairs in isolation, retrieving supporting evidence without exposing how similar perturbations differ in their effects on the same gene. Thereby, models are not given the contrastive signals needed to distinguish perturbation-specific outcomes.

\begin{wrapfigure}[13]{r}[0pt]{0.4\textwidth}
\vspace{-16pt}
    \centering
    \includegraphics[width=\linewidth]{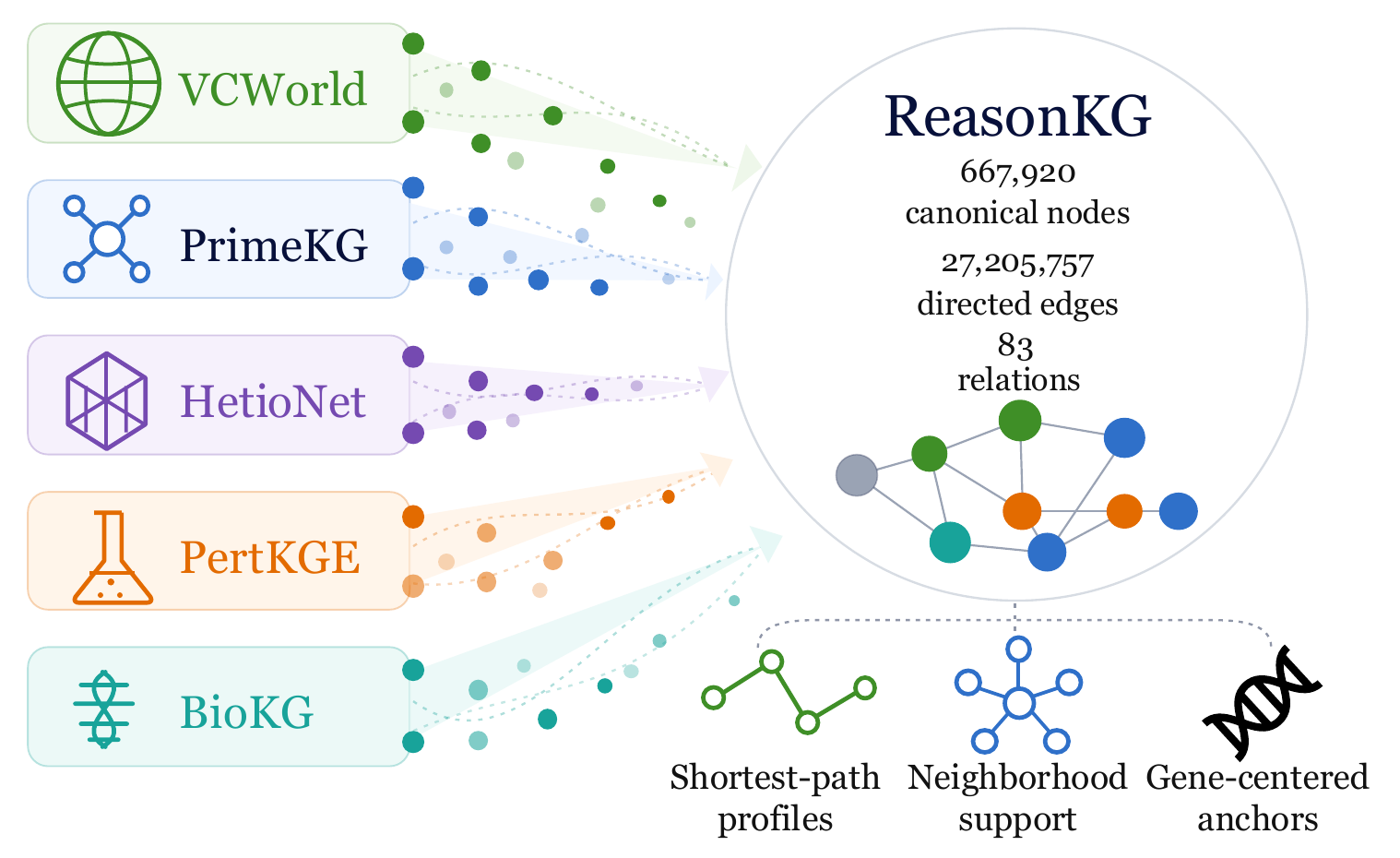}
    \vspace{-15pt}
    \caption{ReasonKG harmonizes five biomedical resources into a canonical  graph for CORE retrieval.}
    \label{fig:reasonkg}
\end{wrapfigure}

To address this limitation, we propose {\core} (Contrastive Organization of Relational Evidence), which reframes prediction as comparative task. As illustrated in Fig.~\ref{fig:core-framework}, for each query, {\core} retrieves related perturbations with observed outcomes for the same gene and organizes them into positive and negative support sets. This enables models to compare how similar perturbations affect the gene, anchoring predictions in empirical evidence rather than isolated plausibility. We instantiate {\core} using newly constructed ReasonKG (Fig.~\ref{fig:reasonkg}), a biomedical knowledge graph for relational retrieval, and evaluate it in both non-LLM and LLM-based settings, where it improves calibration and substantially boosts  perturbation-specific prediction.

We summarize our main contributions as follows. First, we identify two failure modes in LLM-based cellular perturbation reasoning—miscalibration and reliance on gene-level frequency. On Tahoe100M C32, a standard VCWorld~\cite{wei2026vcworld} prompt can
predict ``yes'' for 92.1\% of answered queries despite a 29.0\% positive rate. Second, we introduce per-gene evaluation to measure perturbation-specific discrimination. Third, we propose CORE with a newly constructed knowledge graph ReasonKG, showing that contrastive evidence organization boosts perturbation-specific prediction. Without an LLM, for drug-perturbation data,  \corevoting improves macro-per-gene
AUROC from 0.500 to 0.574 at the fixed support budget and to 0.627 with full
support. With LLM reasoning, \corereasoning raises Mistral-7B aggregate AUROC
from 0.365 to 0.862 and improves Qwen3.5-9B aggregate AUROC, accuracy, and F1
by 26.0\%, 19.8\%, and 28.6\%, respectively. On generic perturbation data, \corevoting
improves macro-per-gene AUROC from 0.500 to 0.711 and \corereasoning similarly improves across LLM backbones, supporting transfer beyond drug perturbations.



\vspace{-6.5pt}
\section{Preliminaries}
\label{sec:preliminaries}
\vspace{-5pt}
\subsection{Perturbation Differential-Expression Prediction}
\label{sec:prelim-task}
\vspace{-4pt}
\textbf{Task definition.} We study binary differential-expression (DE) prediction under perturbation. Each example is a triple \((c,p,g)\), where \(c\) denotes a cellular context (e.g., C32 melanoma cells or K562 leukemia cells), \(p\) is a perturbation such as a drug or genetic intervention, and \(g\) is a target gene. The label \(y_{c,p,g}\in\{0,1\}\) indicates whether \(g\) is called differentially expressed after applying \(p\) in context \(c\). Since our benchmarks evaluate a fixed cellular context at a time~\cite{wu2025perturbqa,zhang2025tahoe100m}, we omit \(c\) when unambiguous and write \(y_{p,g}\).

\textbf{Limited-support held-out perturbations.} We focus on an out-of-distribution setting in which test perturbations are held out from the observed support perturbations. Let \(\mathcal{P}_{\mathrm{sup}}\) denote support perturbations and \(\mathcal{P}_{\mathrm{test}}\) denote held-out test perturbations. Prediction uses only a finite collection of observed DE outcomes, reflecting the practical setting where profiling covers a limited panel of perturbations and genes. A support budget \((k,m)\) provides labels for \(k\) support perturbations and \(m\) measured genes per support perturbation:
\begin{equation}
\mathcal{S}_{k,m}
=
\big\{(p_i,g_j,y_{p_i,g_j}) :
p_i\in\mathcal{P}_{\mathrm{sup}}^{(k)},\,
g_j\in\mathcal{G}_{p_i}^{(m)}
\big\},
\end{equation}
where \(\mathcal{G}_{p_i}^{(m)}\) denotes the set of measured genes selected for perturbation \(p_i\).

\subsection{Perturbation-Specific Effects Evaluation}
\label{sec:prelim-evaluation}
\vspace{-3pt}

\textbf{Aggregate metrics.} Following prior perturbation-reasoning benchmarks~\cite{wu2025perturbqa,wei2026vcworld}, aggregate metrics are computed over all test perturbation--gene pairs. These metrics summarize global predictive performance, but they can be dominated by gene-level response frequencies and therefore do not by themselves show whether a model distinguishes among perturbations for the same target gene.

\textbf{Per-gene metrics.} To measure perturbation-specific discrimination, we also report macro-per-gene metrics. For each gene \(g\), metrics are computed over the held-out set \(\{(p,g):p\in\mathcal{P}_{\mathrm{test}}\}\), considering only genes with both positive and negative labels; the resulting scores are then averaged across genes. Because \(g\) is fixed within each group, these metrics require the model to rank perturbations differently for the same gene, matching the practical question of which perturbations are likely to affect a given target. They are therefore better aligned with perturbation prioritization than aggregate metrics dominated by broadly responsive or unresponsive genes.

\vspace{-6pt}
\section{Motivation: Why Plausibility Fails}
\vspace{-5pt}
\label{sec:motivation}


Recent LLM-based virtual-cell methods cast perturbation prediction as a natural-language reasoning problem~\cite{wei2026vcworld,wu2025perturbqa}. Given a cellular context, a perturbation, and a target gene, the model is prompted to combine biomedical knowledge, retrieved examples, and step-by-step mechanistic reasoning to decide whether the gene is differentially expressed. This formulation is appealing because it produces interpretable rationales and can incorporate semantic biological knowledge that is difficult to represent in purely data-driven models.



However, a plausible mechanistic explanation is not necessarily predictive evidence. Existing prompt formulations largely evaluate each perturbation--gene pair in isolation. They encourage the model to explain whether the pair is biologically plausible, but they do not require the model to compare the query against observed outcomes for related perturbations on the same target gene. Consequently, the model may produce a convincing positive rationale whenever a drug, pathway, or disease association can be connected to the gene, even if the measured cellular response is negative. We next show that this mismatch between plausibility and prediction leads to three empirical failure modes.

\vspace{-4pt}
\subsection{Systematic Overestimation of Positive Predictions}
\vspace{-3pt}

\begin{wrapfigure}[14]{r}[0pt]{0.35\textwidth}
\vspace{-16pt}
    \centering
    \includegraphics[width=\linewidth]{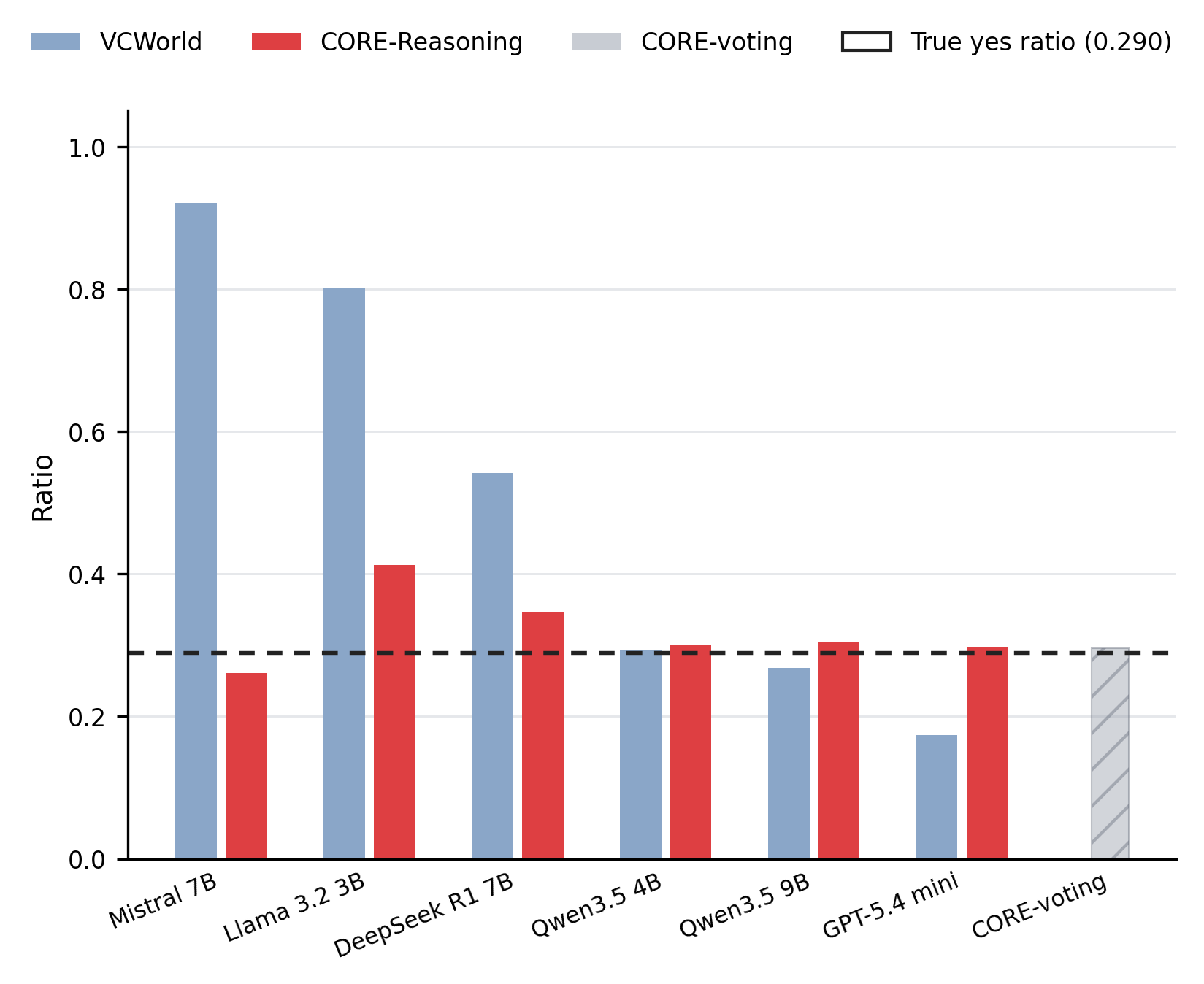}
        \vspace{-15pt}
    \vspace{-2pt}\caption{Predicted positive-rate on Tahoe100M C32, showing VCWorld’s yes bias and CORE’s closer calibration to the true yes ratio.}
    \label{fig:yes_bias_c32}
\end{wrapfigure}

We first examine whether isolated virtual-cell prompts yield calibrated binary decisions. Across VCWorld-style prompts using different base LLMs, we observe a consistent positive-rate bias (Fig.~\ref{fig:yes_bias_c32}): on the drug-perturbation data (Tahoe100M C32), where the empirical test positive rate is \(29.0\%\), several VCWorld runs predict ``yes'' for most queries, with some yes-rates exceeding \(80\%\). This suggests that isolated biological reasoning can be over-permissive: when prompted for a mechanistic rationale, LLMs often find a plausible story and treat it as evidence for differential expression.

This calibration failure is especially problematic for perturbation prediction because negative labels are not merely missing evidence. A gene can be biologically connected to a perturbation through broad stress, pathway, or disease context, yet fail to pass the differential-expression threshold in a specific cell line. Reliable prediction thus requires evidence that distinguishes measured responses, not only evidence that a connection could exist.



\vspace{-2pt}
\subsection{Aggregate Metrics Are Confounded by Gene Response Frequency}
\vspace{-3pt}

The second failure mode is more subtle: high aggregate performance can be achieved without perturbation-specific reasoning. Many genes have stable background response tendencies across perturbations. Some genes are differentially expressed under many conditions, whereas others are rarely called differentially expressed. A predictor can exploit this marginal gene frequency while ignoring the identity of the test perturbation.

To make this confound explicit, we define a diagnostic baseline, {\geneprior}, using only the support set \(\mathcal{S}_{k,m}\). For each gene \(g\) with at least one observed support label, the baseline estimates
\begin{equation}
\hat{\pi}(g)
=
\frac{
\sum\limits_{(p',g',y_{p',g'})\in \mathcal{S}_{k,m}}
\mathbf{1}\{g'=g\}\,y_{p',g'}
}{
\sum\limits_{(p',g',y_{p',g'})\in \mathcal{S}_{k,m}}
\mathbf{1}\{g'=g\}
}.
\label{eq:gene-prior}
\end{equation}
For every held-out perturbation \(p\in\mathcal{P}_{\mathrm{test}}\), it assigns the score
\begin{equation}
s_{\mathrm{prior}}(p,g)=\hat{\pi}(g).
\label{eq:gene-prior-score}
\end{equation}
{\geneprior} is deliberately blind to perturbation identity: all test perturbations receive the same score for a fixed gene. Yet comparison across Fig.~\ref{fig:c32_core_voting_sweep} and Fig.~\ref{fig:c32_llm_metrics} shows that it can match or exceed isolated LLM prompting on standard aggregate metrics. This further exposes an evaluation confound. Aggregate AUROC and AUPRC reward a model for ranking broadly responsive genes above rarely responsive genes, even when the model does not learn which perturbations affect a given gene.

\vspace{-2pt}
\subsection{Per-Gene Discrimination Remains Near Chance}
\vspace{-3pt}

Third, when we remove the gene-frequency shortcut by evaluating within each gene, isolated LLM reasoning provides little perturbation-specific discrimination as analyzed below. Macro-per-gene metrics compute performance over held-out perturbations for a fixed target gene and then average across genes with both positive and negative labels. Because the gene is fixed, success requires assigning different scores to different perturbations for the same gene.

Under this evaluation, {\geneprior} is uninformative by construction, yielding chance-level per-gene AUROC. More importantly, VCWorld-style LLM prompting also remains near chance across model variants (Fig.~\ref{fig:c32_llm_metrics}). The generated rationales may identify plausible biological links, but they do not reliably determine whether this perturbation changes this gene in the measured context.

\textbf{Summary of motivation.} Together, these observations indicate that the bottleneck is not only biomedical knowledge coverage or model capacity. The central issue is how evidence is organized before reasoning. Existing prompts present an isolated query and encourage plausibility assessment; they do not expose the empirical contrast that matters for prediction: among biologically related perturbations, which ones did and did not alter the same target gene? This motivates {\core}, which reframes perturbation prediction as evidence-grounded comparison by retrieving related support perturbations and organizing their same-gene outcomes into positive and negative evidence.


\vspace{-5pt}
\section{Method:  {\core}}
\label{sec:method}
\vspace{-5pt}

\begin{figure*}[t]
\vspace{-10pt}
    \centering
    \includegraphics[width=0.98\textwidth]{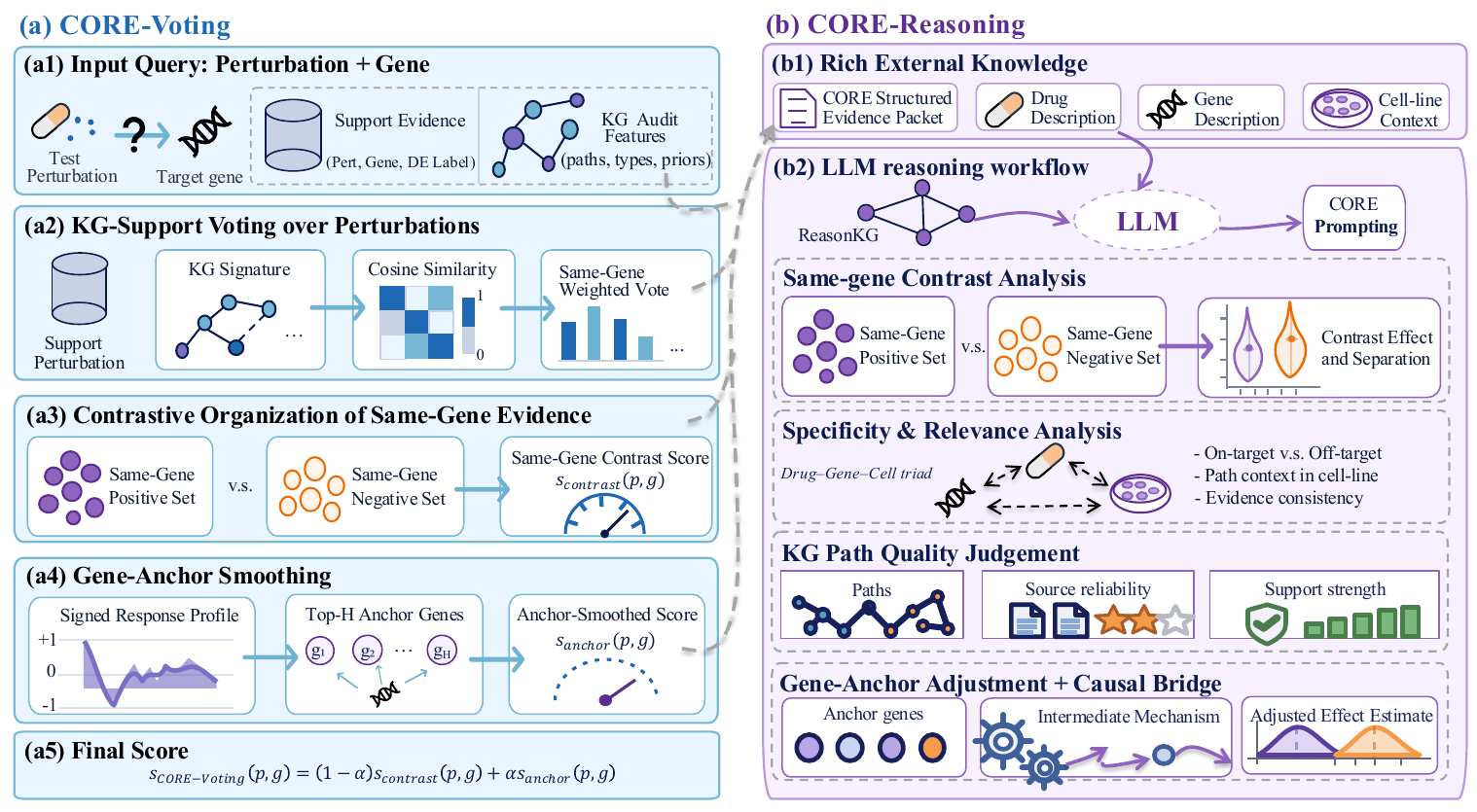}
    \caption{CORE framework. KG-based perturbation signatures retrieve same-gene positive and negative support cases (Sec.~\ref{sec:method-core}); CORE-Voting aggregates them directly (Sec.~\ref{sec:method-core-voting}), while CORE-Reasoning presents the structured evidence packet to an LLM for reasoning (Sec.~\ref{sec:method-core-reasoning}).}
    \label{fig:core-framework}
    \vspace{-10pt}
\end{figure*}

\core is a knowledge-grounded framework for perturbation differential-expression prediction. Its central design choice is to replace isolated pairwise plausibility judgments with \emph{contrastive same-gene evidence}. For a held-out query \((p,g)\), where \(p\) is a perturbation and \(g\) is the query gene, \core retrieves support perturbations related to \(p\), collects their observed labels for the same gene \(g\), and separates the evidence into positive and negative support cases. Since \(g\) is fixed throughout this comparison, the model must distinguish among perturbations rather than relying solely on abstract mechanistic plausibility. 


{\core} is model-agnostic. We instantiate it in two complementary settings to
validate the framework (Fig.~\ref{fig:core-framework}). {\corevoting} is a non-LLM deterministic evidence aggregator designed to test
whether the contrastive support evidence itself contains perturbation-specific signal,
independent of LLM reasoning (Sec.~\ref{sec:method-core-voting}). {\corereasoning} reformulates the same evidence
into a structured prompt
for LLMs, enabling us to evaluate
whether LLMs can leverage
contrastive evidence more effectively than standard prompting strategies (Sec.~\ref{sec:method-core-reasoning}).

Next, we describe the biomedical knowledge graph, the construction of the contrastive evidence sets, and the two framework instantiations in detail. 


\vspace{-3pt}
\subsection{ReasonKG: A Biomedical Knowledge Graph for Contrastive Perturbation Reasoning}
\label{sec:method-reasonkg}
\vspace{-3pt}

CORE requires knowledge-grounded evidence: support perturbations should be biologically related to the query
perturbation via structured biomedical relationships rather than surface textual similarity alone.
However, we find that existing knowledge graphs used in this setting are
often sparse or fragmented, limiting their ability to support gene-conditioned perturbation comparison.
For example, in VCWorld, many drug and gene entities lack traversable paths that can
connect perturbations through shared biological context (Fig.~\ref{fig:reasonkg-vcworld-connectivity}). We therefore construct a biomedical
knowledge graph $\mathcal{G}$, termed ReasonKG. ReasonKG is used to identify support perturbations related to the query
perturbation (Sec.~\ref{sec:method-core}) and to provide interpretable relational paths for downstream reasoning (Sec.~\ref{sec:method-core-reasoning}).

ReasonKG integrates and harmonizes
multiple biomedical resources~\cite{wei2026vcworld,chandak2023building,himmelstein2017systematic,ni2024pertkge,walsh2020biokg},
providing broad and biologically actionable coverage. Most nodes are centered on genes, proteins,
pathways, GO terms, and diseases (Fig.~\ref{fig:reasonkg}), while multi-source merged nodes create
reliable anchors for connecting heterogeneous biomedical knowledge. Its diverse relation vocabulary
links perturbations to targets, pathways, functions, and biological contexts,
enabling biologically grounded relational reasoning.
Additional analyses of node and relation distributions, graph connectivity, and comparisons with the VCWorld knowledge graph are provided in App.~\ref{app:kg-resources}.

\vspace{-2pt}
\subsection{Construction of Contrastive Evidence Sets}
\label{sec:method-core}
\vspace{-3pt}

Given a query \((c, p, g)\), {\core} 
retrieves perturbations biologically related to \(p\) using the biomedical knowledge graph introduced in Sec.~\ref{sec:method-reasonkg}, collects their observed outcomes for the target gene \(g\), and organizes the retrieved evidence into positive and negative support sets, as detailed below.

\textbf{KG-derived perturbation signatures.} For each KG-covered perturbation \(p\), {\core} constructs a sparse signature \(\phi(p)\). The signature summarizes the perturbation's relation to KG-covered genes using shortest-path distance buckets and training-estimated proximity statistics (details in App.~\ref{app:kg-signatures}).

\textbf{Retrieving structurally similar perturbations.} Given a query perturbation \(p\) and a support perturbation \(p' \in \mathcal{P}_{\mathrm{sup}}\), {\core} computes their perturbation-side similarity as clipped cosine similarity:
\begin{equation}
    \mathrm{sim}(p,p')
=
\max\{0,\cos(\phi(p),\phi(p'))\}.
\end{equation}
The top-k most similar support perturbations are then retrieved: 
\begin{equation}
    \mathcal{R}_k(p) = \mathrm{TopK}_{p'\in\mathcal{P}_{\mathrm{sup}}} \mathrm{sim}(p,p').
\end{equation}
To quantify support strength, similarity scores are 
converted into a nonnegative voting weight by a monotone power transformation:
\begin{equation}
w(p,p')
=
\mathrm{sim}(p,p')^\eta.
\end{equation}

\textbf{Contrastive organization of same-gene evidence.} {\core} partitions retrieved support perturbations by their observed outcome for the same gene: 
\begin{equation}
\mathcal{H}^{+}_{p,g} = \{p'\in\mathcal{R}_k(p): y_{p',g}=1\}, \qquad \mathcal{H}^{-}_{p,g} = \{p'\in\mathcal{R}_k(p): y_{p',g}=0\}.    
\end{equation}
Positive support cases contain  related perturbations that affected \(g\), while negative cases contain related perturbations that did not. This same-gene contrast is central to {\core}. Because \(g\) is fixed, the evidence compares perturbations rather than genes.  Both 
positive and negative evidence 
are necessary because biological relatedness does not imply identical transcriptional response. 

Finally, {\core} summarizes the retrieved evidence using positive and negative vote masses: 
\begin{equation}
    V^{+}_{p,g} = \sum_{p'\in\mathcal{H}^{+}_{p,g}} w(p,p'), \qquad V^{-}_{p,g} = \sum_{p'\in\mathcal{H}^{-}_{p,g}} w(p,p').
\end{equation}
\vspace{-10pt}

The resulting evidence packet contains the query perturbation, 
positive and negative support sets, and the corresponding 
vote masses. {\corevoting} uses this packet numerically 
while  {\corereasoning} serializes it into an LLM prompt for reasoning,
 as detail next. 

\vspace{-2pt}
\subsection{{\corevoting}} 
\label{sec:method-core-voting}
\vspace{-3pt}
{\corevoting} is a non-LLM, deterministic predictor that directly aggregates the contrastive support evidence, to test whether the contrastive support evidences inherently contains perturbation-specific signals. We define the weighted positive support fraction as a contrast score for the same fixed gene: 
\begin{equation}
s_{\mathrm{contrast}}(p,g)
=
\begin{cases}
\frac{V^{+}_{p,g}} {V^{+}_{p,g}+V^{-}_{p,g}}, & V^{+}_{p,g}+V^{-}_{p,g} > 0,\\[1.2ex]
\hat{\pi}(g), & \mathrm{otherwise}.
\end{cases}
\end{equation}
\vspace{-5pt}

This score transfers evidence from observed support examples \(y_{p',g}\) for the same target gene \(g\). A high value means that under perturbations similar to \(p\), relevant support evidence predicts \(g\) to be more likely differentially expressed. When no same-gene support examples are observed, we fall back to the support gene prior defined in Eqn.~\ref{eq:gene-prior}.

\textbf{Gene-anchor smoothing.} To improve robustness under limited support, {\core} also applies gene-anchor smoothing to borrow insights from genes similar to the query gene \(g\).  A signed support response profile $r_g$ over support perturbations is defined as
\begin{equation}
    r_g(p')
=
\begin{cases}
2y_{p',g}-1, & \text{if } (p', g, y_{p',g}) \in \mathcal{S}_{k,m},\\
0, & \text{otherwise}.
\end{cases}
\end{equation}
\vspace{-1.2pt}
Gene-gene similarity is computed by clipped cosine similarity:
\begin{equation}
\mathrm{sim}(g,g')
=
\max\{0,\cos(r_g,r_{g'})\}. 
\end{equation}
We identify top $H$ response-similar anchor genes \begin{equation}
    \mathcal A_H(g) = \mathrm{TopH}_{g'} \mathrm{sim}(g, g').
\end{equation}
\vspace{-1.2pt}
The anchor-smoothed score averages the same-gene contrast scores over the anchor genes:
\begin{equation}
s_{\mathrm{anchor}}(p,g)
=
\frac{
\sum_{g'\in\mathcal A_H(g)}
\mathrm{sim}(g,g') s_{\mathrm{base}}(p,g')
}{
\sum_{g'\in\mathcal A_H(g)}
\mathrm{sim}(g,g')
}.
\end{equation}
\vspace{-1.2pt}
\textbf{Final {\corevoting} score}. We compute a convex combination of the same-gene contrast score and the anchor-smoothed score:
\begin{equation}
s_{\text{{\corevoting}}}(p,g)
=
(1-\alpha)s_{\mathrm{contrast}}(p,g)
+
\alpha s_{\mathrm{anchor}}(p,g).    
\end{equation}

\subsection{{\corereasoning}}
\label{sec:method-core-reasoning}
\vspace{-3pt}

{\corereasoning} turns {\core} evidence into a structured LLM prompt. For each query \((p,g)\), the evidence contains: 
\textbf{(i)} the gene prior $\hat{\pi}(g)$, marked as a bias to avoid over-reliance; 
\textbf{(ii)} same-gene contrast evidence, including $s_{\mathrm{contrast}}(p,g)$, positive and negative support perturbations ($\mathcal{H}^+_{p,q}$,   $\mathcal{H}^-_{p,q}$), and the corresponding vote masses ($V^+_{p,q}$, $V^-_{p,q}$);
\textbf{(iii)} shortest KG paths between the grounded or mapped query perturbation and gene (path retrieval detailed in App.~\ref{app:kg-path-evidence}); and 
\textbf{(iv)} gene-anchor smoothing evidence, including \(s_{\mathrm{anchor}}(p,g)\), top anchor genes $\mathcal{A}_H(g)$, and the final score $s_{\text{\corevoting}}(p,g)$.

The LLM then adjudicates evidence quality: whether positive and negative supports differ in biological relevance, whether KG paths provide specific biological mechanistic bridges or only generic connectivity, whether endpoint mappings are reliable, and whether anchor-gene smoothing is consistent with the same-gene contrast. 
Thus, {\corereasoning} uses the LLM for structured evidence interpretation---resolving conflicts, assessing path quality, and explaining query-specific deviations---instead of unconstrained plausibility generation.

\if false
\section{Related Work}

\textbf{Single-cell perturbation response prediction.}
Perturbation prediction has traditionally been formulated as learning how cellular expression states transform under genetic or chemical interventions. Methods such as CPA~\cite{lotfollahi2023predicting}, GEARS~\cite{roohani2024predicting}, scGPT~\cite{cui2024scgpt}, and State~\cite{adduri2025state} have substantially advanced transcriptome-level response modeling, covering compositional perturbation autoencoding, graph-enhanced multigene perturbation prediction, single-cell foundation modeling, and large-scale context-aware virtual-cell modeling. However, these objectives are primarily defined over reconstructed or predicted expression profiles, distributional shifts, or transcriptome-level summary metrics. They do not require the model to explicitly distinguish why one perturbation should alter a specific target gene while a related perturbation should not. Recent benchmarks have exposed this gap: deep perturbation models and single-cell foundation models can fail to outperform simple baselines in expression-change prediction~\cite{ahlmanneltze2025deep}, and PerturbQA shows that differential-expression-oriented perturbation reasoning remains difficult even when the task is cast directly as binary or directional prediction over perturbation--gene pairs~\cite{wu2025perturbqa}.

\textbf{LLM-based perturbation reasoning.}
Recent work has reframed perturbation prediction as knowledge-grounded biological reasoning with LLMs. PerturbQA introduced QA tasks for differential expression, direction of change, and gene-set enrichment under unseen perturbations~\cite{wu2025perturbqa}, and CellVerse extended language-centric evaluation to single-cell multi-omics settings, including drug-response and gene-level perturbation-analysis tasks~\cite{zhang2025cellverse}. SUMMER improved LLM performance on PerturbQA by combining biological summaries, retrieved experimental evidence, and prompting within an inference-time framework~\cite{wu2025perturbqa}. VCWorld further pushes this paradigm toward virtual-cell simulation by integrating structured biological knowledge with iterative LLM reasoning to generate stepwise, interpretable drug-perturbation predictions~\cite{wei2026vcworld}. However, these methods largely reason over perturbation--gene pairs in isolation, which can yield biologically plausible explanations without true perturbation-specific discrimination. {\core} targets this limitation by organizing retrieved evidence contrastively, comparing positive and negative outcomes from related perturbations on the same target gene.

\fi 

\vspace{-5pt}
\section{Experiments}
\label{sec:experiment}
\vspace{-5pt}

We evaluate whether \core improves differential-expression prediction beyond
two common failure modes of LLM-based perturbation reasoning: high aggregate
scores driven by gene response frequency, and poorly calibrated
yes-biased predictions from isolated biological plausibility judgments. Our
experiments address four questions: \textbf{(1)} how strong is the gene response
prior under standard aggregate metrics? \textbf{(2)} does contrastive same-gene
evidence contain perturbation-specific signal even without an LLM? \textbf{(3)} can
LLMs use the same evidence to improve prediction and calibration under a fixed
support budget? \textbf{(4)} do the results transfer from drug perturbations to a
generic perturbations?


\vspace{-3pt}
\subsection{Experimental Setup}
\vspace{-3pt}

\textbf{Datasets.}
We evaluate binary differential-expression (DE) prediction in two benchmark families across five fixed cellular contexts: Tahoe100M drug perturbations~\cite{zhang2025tahoe100m} in C32 melanoma cells and PerturbQA CRISPRi gene perturbations~\cite{wu2025perturbqa} in K562, HepG2, Jurkat, and RPE1 cells.
 Split statistics and preprocessing
details are in Tab.~\ref{tab:dataset_statistics} and App.~\ref{app:dataset-details}.

\textbf{Compared methods.}
Tab.~\ref{tab:method-taxonomy} summarizes the method families. {\geneprior}
gene response probabilities from the support split, ignoring perturbation identity. Standard baseline prompt uses task-native biological prompts without labeled same-gene contrast: VCWorld-style
prompting~\cite{wei2026vcworld} for Tahoe100M C32 and SUMMER-style prompting~\cite{wu2025perturbqa} for PerturbQA K562.
{\corevoting} aggregates KG-retrieved same-gene support labels without an LLM. {\corereasoning} gives the same contrastive evidence along with interpretable KG paths to an
LLM for reasoning. We use SGLang inference~\cite{zheng2024sglang} for all LLM variants except for GPT-5.4-mini. We also compare against STATE~\cite{adduri2025predicting} and PerturbDiff~\cite{yuan2026perturbdiff}, two recent state-of-the-art baselines for single-cell perturbation prediction. 
Including these baselines allows us to test the advantage of LLM reasoning.

\begin{table}[t]
\vspace{-10pt}
\centering
\small
\setlength{\tabcolsep}{5pt}
\caption{
Taxonomy of evaluated method families.
The key distinction is whether the model use same-gene supports as positive and negative evidence for the query perturbation as {\core} proposed.
}
\begin{tabular}{lcccc}
\toprule
Method & KG retrieval & Support labels & Same-gene contrast & LLM reasoning \\
\midrule
{\geneprior} & No & Yes & No & No \\
Baseline prompt & No & Yes & No & Yes \\
\corevoting & Yes & Yes & Yes & No \\
\corereasoning & Yes & Yes & Yes & Yes \\
\bottomrule
\end{tabular}
\label{tab:method-taxonomy}
\vspace{-6pt}
\end{table}

\textbf{Metrics.}
We report AUROC, AUPRC, accuracy, and F1 at instance and macro-per-gene levels. Per-gene AUROC and AUPRC average per-gene rankings, testing whether methods distinguish perturbations for the same gene rather than exploit gene-prior shortcuts. For LLMs, we also report effective answer rate and predicted-yes rate, and rescale results by effective answer rate following~\cite{wei2026vcworld}.

\textbf{Support budgets.}
On Tahoe100M C32, we sweep
\((k,m)\), $k$ support perturbations and $m$ maximum observed genes per perturbation. For LLM experiments,
we fix \(k=50,m=200\), which yields a moderate, prompt-feasible support pool and
uses the same held-out test perturbations as full support. All prompt variants are evaluated on the same support pool and held-out test set. All PerturbQA-DE cell lines use the full-support setting.

\begin{figure*}[t]
\vspace{-10pt}
    \centering
    \begin{subfigure}[t]{0.62\textwidth}
        \centering
        \includegraphics[width=\linewidth]{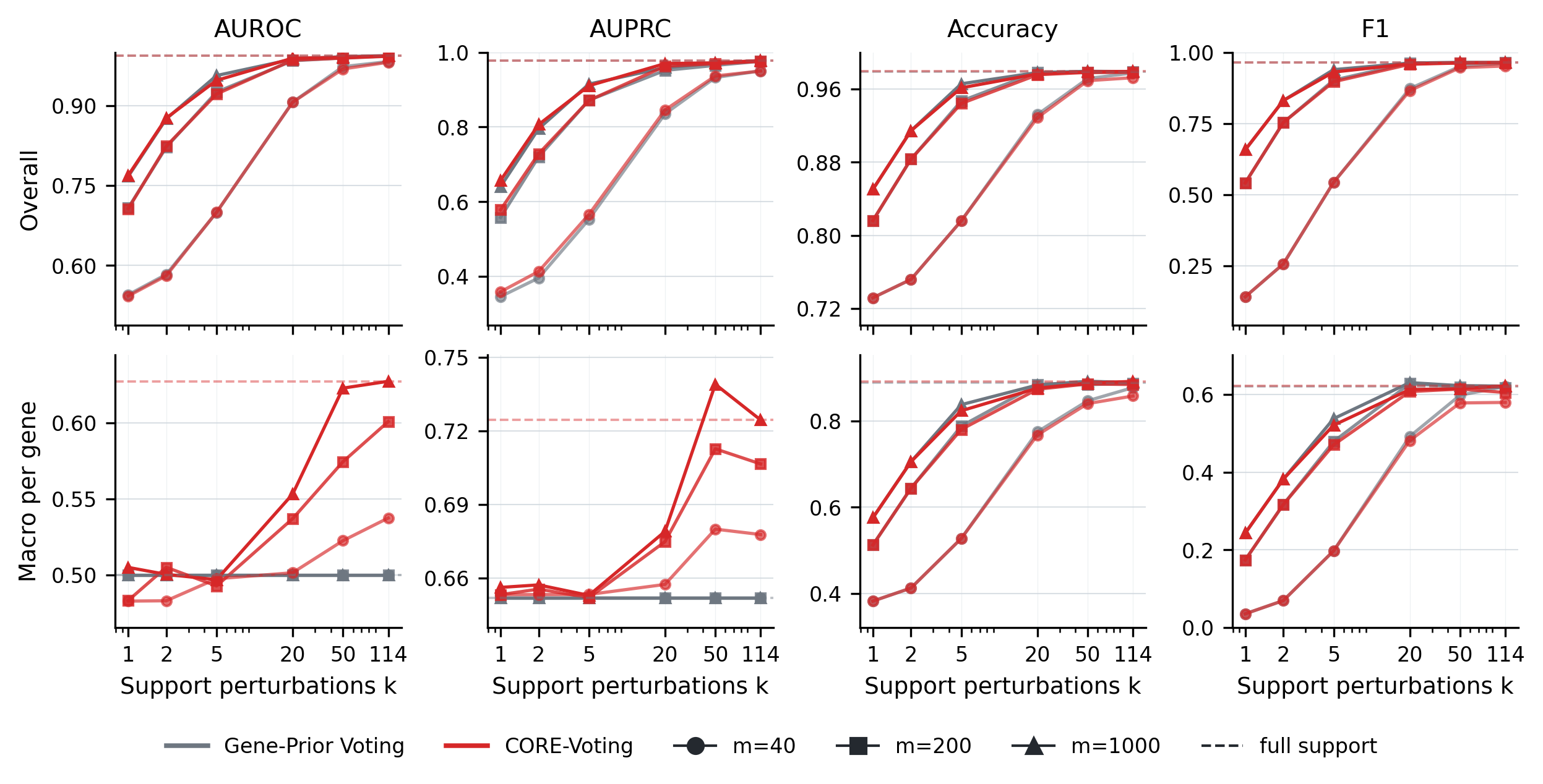}
        \caption{Support-budget sweep on Tahoe100M C32.}
        \label{fig:c32_core_voting_sweep}
    \end{subfigure}
    \hfill
    \begin{subfigure}[t]{0.35\textwidth}
        \centering
        \includegraphics[width=\linewidth]{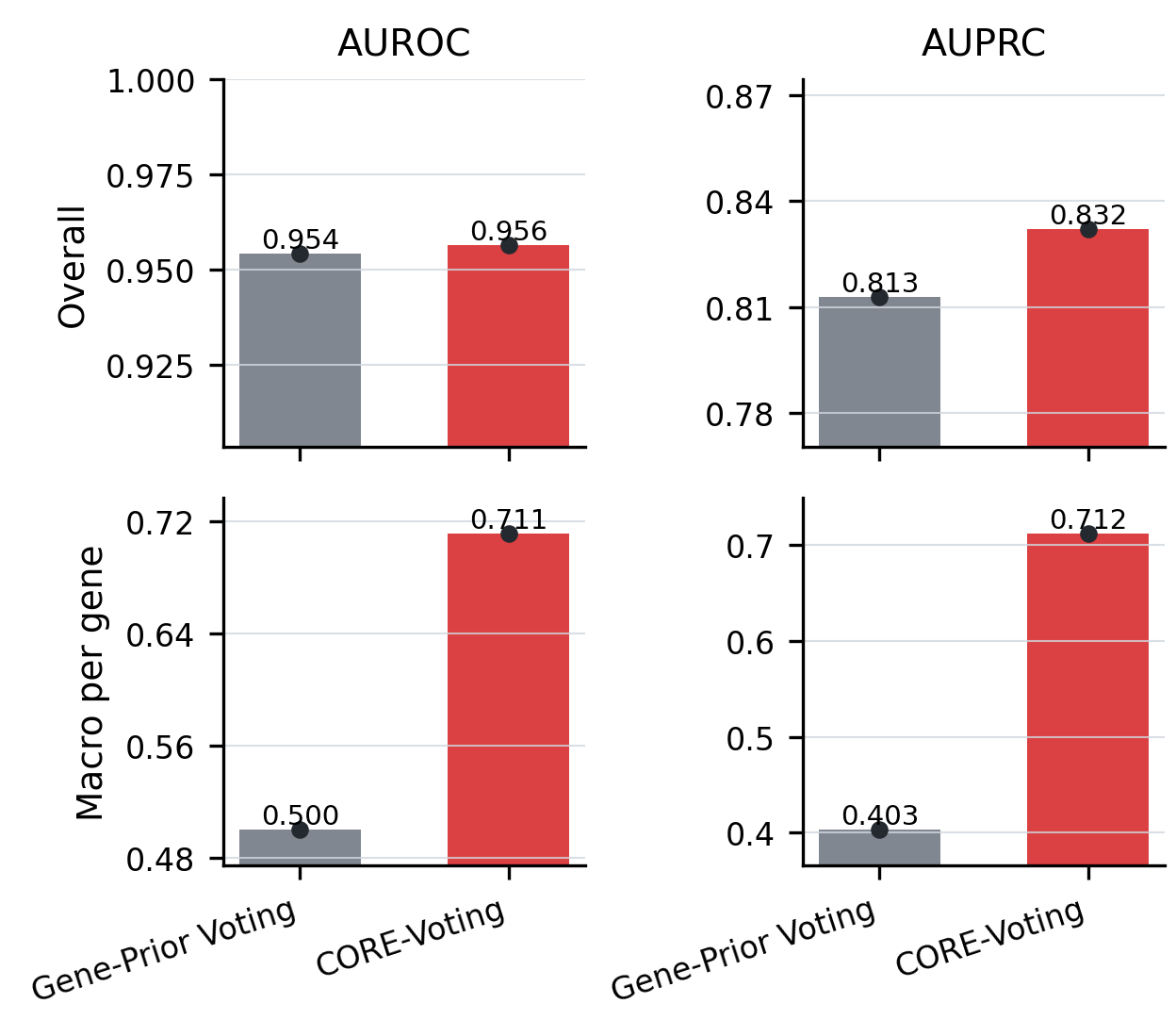}
        \caption{Full-support PerturbQA K562.}
        \label{fig:k562_core_voting_auroc_auprc}
    \end{subfigure}
    \caption{
    Comparison of {\geneprior} and \corevoting across support sweeps.
    {\geneprior} can score well on aggregate metrics but is uninformative within gene, whereas {\corevoting} improves macro-per-gene performance by using contrastive same-gene support evidence.
    }
    \label{fig:support_sweep}
    \vspace{-8pt}
\end{figure*}
\begin{figure*}[t]
\vspace{-6pt}
    \centering
    \includegraphics[width=1.0\textwidth]{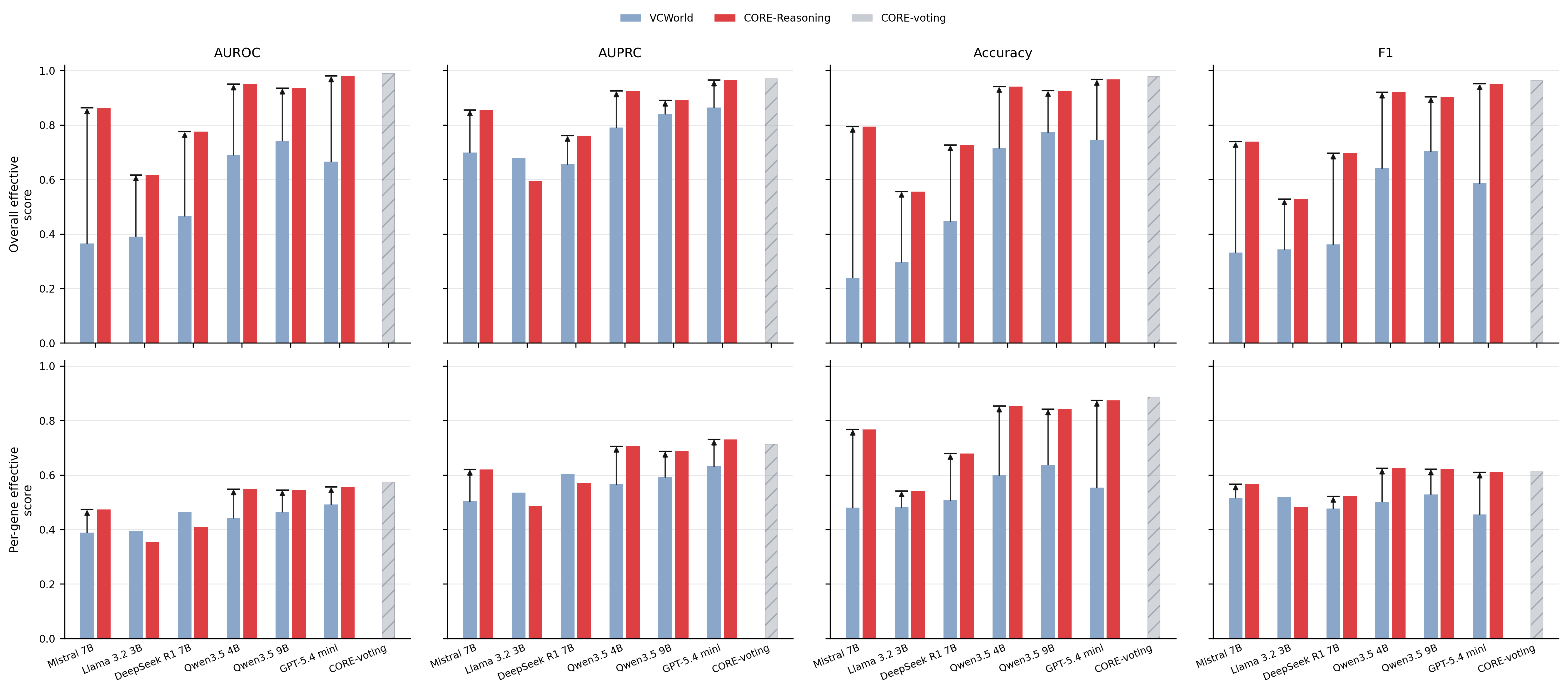}
    \caption{
    LLM comparison  at fixed-budget (\(k=50,m=200\)) on Tahoe100M C32.
    \corereasoning substantially improves aggregate metrics across LLM backbones and improves macro per-gene discrimination most clearly for     Qwen3.5 models and GPT-5.4 mini, while \corevoting remains a strong non-LLM
    reference.
    }
    \vspace{-12pt}
    \label{fig:c32_llm_metrics}
\end{figure*}

\vspace{-3pt}
\subsection{Support Sweeps for \corevoting}
\vspace{-3pt}

Fig.~\ref{fig:support_sweep} isolates the value of contrastive support
evidence from free-form LLM reasoning. On Tahoe100M C32, {\geneprior} is already strong
under aggregate metrics: as the support budget grows, background response
frequencies alone drive overall AUROC and AUPRC toward saturation. However, the
same baseline has macro per-gene AUROC of 0.500 for every support setting,
since it assigns the same score to all perturbations for a fixed gene.

\corevoting preserves the strong aggregate behavior of the gene prior while
adding perturbation-specific discrimination. Using support 
\(S_{k=50,m=200}\), \corevoting improves macro per-gene AUROC from 0.500 to
0.574 and macro per-gene AUPRC from 0.652 to 0.713, while matching the
gene-prior baseline on aggregated metrics. Using full support from Tahoe100M C32,
aggregate metrics are nearly saturated for both methods, but the per-gene
gap remains: per-gene AUROC increases from 0.500 to 0.627 and per-gene AUPRC from
0.652 to 0.724. Thus, KG-organized same-gene evidence captures signal that is invisible to gene-frequency predictors.

On PerturbQA K562, 
\corevoting improves every reported metric over {\geneprior}, including overall
AUPRC from 0.813 to 0.832 and per-gene AUROC from 0.500 to 0.711
(Tab.~\ref{tab:k562-de-core-geneprior}). 

\vspace{-2pt}
\subsection{LLM Performance}
\vspace{-3pt}

\textbf{Tahoe100M C32 drug perturbations.}
Fig.~\ref{fig:c32_llm_metrics} and Tab.~\ref{tab:c32-llm-core-results}
compare standard VCWorld-style prompting with {\corereasoning}. {\corereasoning} improves effective AUROC,
accuracy, and F1 for every evaluated LLM backbone. The effect is especially
large for models whose standard prompts produce highly biased predictions:
Mistral-7B, for example, predicts ``yes'' on 92.1\% of answered examples under
the standard prompt despite a 29.0\% test positive rate, while {\corereasoning}
reduces the predicted-yes rate to 26.1\% and raises effective AUROC from 0.365
to 0.862. Similar calibration improvements are observed for Llama-3.2-3B and DeepSeek-R1-7B.

The Qwen3.5 runs show that {\corereasoning} gains persist for stronger LLMs. With Qwen3.5-9B, {\corereasoning} improves effective AUROC, Accuracy, and F1 by 26.0\%, 19.8\%, and 28.6\%, respectively, while keeping the predicted-yes rate close to the empirical prevalence. Macro-per-gene AUROC and AUPRC also improve by 17.2\% and 16.0\%. The same trend is visible for Qwen3.5-4B.

Per-gene gains are more selective than aggregate gains: smaller models often improve calibration and instance-level metrics without improving within-gene ranking. This reflects the metric distinction: aggregate metrics capture common response patterns, whereas macro-per-gene metrics test perturbation discrimination for the same gene. {\corevoting} remains competitive with the best LLM runs, reaching effective AUROC/AUPRC of 0.989/0.970 and per-gene AUROC/AUPRC of 0.574/0.713.

\textbf{PerturbQA CRISPRi generic perturbations.}
Fig.~\ref{fig:k562_llm_metrics} evaluates the same idea in the K562
PerturbQA setting, where the standard prompt is SUMMER-style. {\corereasoning} improves over
SUMMER-style prompting for several LLM backbones, most clearly for Mistral-7B,
DeepSeek-R1-7B, and Qwen3.5-4B. The gains appear in both aggregate and
macro-per-gene metrics, indicating that contrastive support evidence can help outside the drug perturbation setting.

Since PerturbQA K562 leverages the full and KG coverage for the queries is high, deterministic {\corevoting} is already
very strong, achieving 0.956 overall AUROC, 0.832 overall AUPRC, and 0.711
macro per-gene AUROC (Tab.~\ref{tab:k562-de-core-geneprior}). In this regime,
LLMs do not uniformly dominate the evidence aggregator; their main value is instead in interpreting and explaining contrastive evidence, while the numerical support signal itself is often sufficient.

We report the performance for the other three PerturbQA cell lines in App.~\ref{app:full_numerical_results}.

\begin{table*}[t]
\centering
\small
\setlength{\tabcolsep}{4.5pt}
\caption{
Comparison with strong single-cell perturbation baselines on Tahoe100M C32 and PerturbQA K562. 
The best value for each metric within each dataset is shown in bold. 
}
\label{tab:baseline_comparison}
\resizebox{\textwidth}{!}{%
\begin{tabular}{@{}llcccccccc@{}}
\toprule
Dataset & Method
& \multicolumn{4}{c}{Overall}
& \multicolumn{4}{c}{Macro per gene} \\
\cmidrule(lr){3-6}
\cmidrule(lr){7-10}
& 
& AUROC & AUPRC & Acc. & F1
& AUROC & AUPRC & Acc. & F1 \\
\midrule
C32  & STATE
& \textbf{0.992} & \textbf{0.968} & 0.769 & 0.342
& 0.492 & 0.652 & 0.449 & 0.126 \\
 & PerturbDiff
& 0.987 & 0.953 & 0.785 & 0.416
& 0.511 & 0.678 & 0.474 & 0.159 \\
  & VCWorld (GPT-5.4 mini) &  0.666 & 0.864 & 0.746 & 0.586 & 0.492 & 0.631 & 0.554 & 0.455 \\
 & CORE-Reasoning (GPT-5.4 mini)
& 0.980 & 0.964 & \textbf{0.967} & \textbf{0.951}
& \textbf{0.556} & \textbf{0.731} & \textbf{0.874} & \textbf{0.610} \\
\midrule
K562 & STATE
& 0.629 & 0.394 & 0.412 & 0.254
& 0.504 & 0.572 & 0.407 & 0.437 \\
 & PerturbDiff
& \textbf{0.888} & \textbf{0.663} & 0.525 & 0.349
& 0.495 & 0.575 & 0.432 & \textbf{0.497} \\


  & SUMMER (Qwen3.5 9B)  & 0.511 & 0.147 & 0.795 & 0.168 & 0.485 & 0.426 & 0.846 & 0.017 \\
 & {\corereasoning} (Qwen3.5 9B)   & {0.816} & {0.615} & \textbf{0.900} & \textbf{0.717} & \textbf{0.607} & \textbf{0.623} & \textbf{0.915} & {0.064} \\
\bottomrule
\end{tabular}%
}
\end{table*}

\textbf{Comparison with strong single-cell perturbation response prediction baselines.} Tab.~\ref{tab:baseline_comparison} shows that the LLM reasoning paradigm can provide a strong alternative to conventional single-cell perturbation modeling, even when evaluated against the latest strong baselines. CORE-Reasoning achieves the strongest overall performance, outperforming both prior LLM-based methods and recent state-of-the-art single-cell baselines on most metrics.

\vspace{-2pt}
\subsection{Case Studies}
\vspace{-3pt}

Tab.~\ref{tab:case_study_examples} shows representative C32 examples from
the Qwen3.5-9B runs. \corereasoning changes model
behavior by contrasting same-gene evidence: mostly negative supports correct false positives, while positive supports, anchor genes, or compact KG paths rescue false negatives. Mixed-evidence cases expose another advantage: \corereasoning surfaces disagreement among related
perturbations instead of treating all retrieved biological context as uniformly
supportive. For {\corereasoning}, failures occur when perturbation or gene mappings are weak or evidence is sparse, motivating coverage reporting and path-quality checks.


\vspace{-5pt}
\section{Related Work}
\vspace{-5pt}

\textbf{Single-cell perturbation response prediction.}
Perturbation prediction methods such as CPA~\cite{lotfollahi2023predicting}, GEARS~\cite{roohani2024predicting}, scGPT~\cite{cui2024scgpt}, and State~\cite{adduri2025state} model cellular responses to genetic or chemical interventions. Their objectives typically emphasize expression reconstruction, distributional shifts, or aggregate metrics rather than perturbation--gene discrimination. Recent benchmarks show that strong perturbation and foundation models can underperform simple baselines~\cite{ahlmanneltze2025deep}, while binary perturbation reasoning over perturbation--gene pairs remains challenging~\cite{wu2025perturbqa}. {\core} instead organizes relational evidence contrastively at this level.

\textbf{LLM-based perturbation reasoning.}
Recent LLM methods cast perturbation prediction as knowledge-grounded reasoning: PerturbQA defines differential-expression, direction, and enrichment QA tasks~\cite{wu2025perturbqa}; CellVerse extends language-centric evaluation to single-cell multi-omics~\cite{zhang2025cellverse}; SUMMER combines biological summaries, retrieved evidence, and prompting~\cite{wu2025perturbqa}; and VCWorld integrates structured knowledge with iterative virtual-cell reasoning~\cite{wei2026vcworld}. These methods largely assess perturbation--gene pairs in isolation; {\core} compares related perturbations with positive and negative outcomes for the same target gene.

\textbf{In-context learning in LLMs.}
In-context learning (ICL) enables task adaptation through instructions and demonstrations without parameter updates~\cite{brown2020language}. Prior work shows that performance depends strongly on demonstration relevance, retrieval, ordering, and label or format structure~\cite{liu2022makes,rubin2022learning,lu2022fantastically,min2022rethinking}. For perturbation reasoning, examples should define the relevant comparison, not merely provide plausible support. {\core} therefore uses ICL contrastively by presenting matched same-gene positive and negative outcomes, encouraging comparative rather than plausibility-only reasoning.

\vspace{-5pt}
\section{Conclusion, Limitations and Broader Impact}
\vspace{-5pt}
We show that biologically plausible reasoning is not sufficient for reliable perturbation prediction. By organizing retrieved relational evidence into same-gene positive and negative support sets, CORE reframes prediction as evidence-grounded comparison rather than isolated plausibility assessment. Across LLM-based and non-LLM settings, this contrastive organization improves calibration and perturbation-specific discrimination, suggesting that reliable virtual-cell reasoning depends as much on evidence structure as on biological knowledge coverage.

\textbf{Limitations.}
This study focuses on binary differential-expression prediction in fixed cellular contexts, leaving dose, time, protocol variation, expression magnitude, and DEG uncertainty for future work. CORE also depends on knowledge-graph and support-set coverage, so sparse or biased evidence can weaken retrieval and reasoning.

\textbf{Broader impact.}
CORE may help prioritize experiments and make LLM-assisted perturbation reasoning more auditable by exposing positive and negative support evidence. Its predictions should be used for hypothesis generation, not as a substitute for expert review, experimental validation, or clinical decision making.

\begin{ack}
This project is supported by the Canada CIFAR AI Chair Program.
\end{ack}

\clearpage
\bibliographystyle{plain}
\bibliography{references}
\clearpage
\appendix
\onecolumn

\label{app:overview}

\section{Dataset Details}
\label{app:dataset-details}

\begin{table}[h]
\centering
\small
\caption{Dataset statistics for the primary binary DE settings. ``Support''
denotes the training split available to retrieval, voting, and prompt
construction; ``test'' denotes held-out perturbation--gene pairs used for
evaluation.}
\begin{adjustbox}{max width=\textwidth}
\begin{tabular}{llrrrrr}
\toprule
Dataset & Split & Pairs & Perturbations & Genes & Positives & Positive rate \\
\midrule
C32 Tahoe100M (k=50,m=200) & all &
\(84{,}642\) & 315 & 2,000 & \(24{,}525\) & 28.97\% \\
 & support &
\(10{,}000\) & 50 & 1,962 & \(2{,}883\) & 28.83\% \\
 & test &
\(74{,}642\) & 265 & 2,000 & \(21{,}642\) & 28.99\% \\
\midrule
K562 PerturbQA & all &
\(157{,}679\) & 1,831 & 4,136 & \(19{,}980\) & 12.67\% \\
 & support &
\(134{,}467\) & 1,564 & 4,136 & \(16{,}861\) & 12.54\% \\
 & test &
\(23{,}212\) & 267 & 4,118 & \(3{,}119\) & 13.44\% \\

\midrule

HepG2 PerturbQA & all &
  \(126{,}889\) & 1,364 & 7,435 & \(17{,}860\) & 14.08\% \\
   & support &
  \(101{,}140\) & 1,086 & 7,435 & \(14{,}257\) & 14.10\% \\
   & test &
  \(25{,}749\) & 278 & 7,096 & \(3{,}603\) & 13.99\% \\
  \midrule
  Jurkat PerturbQA & all &
  \(142{,}822\) & 1,540 & 6,842 & \(20{,}058\) & 14.04\% \\
   & support &
  \(113{,}684\) & 1,227 & 6,842 & \(15{,}937\) & 14.02\% \\
   & test &
  \(29{,}138\) & 313 & 6,698 & \(4{,}121\) & 14.14\% \\
  \midrule
  RPE1 PerturbQA & all &
  \(187{,}089\) & 2,002 & 4,760 & \(26{,}652\) & 14.25\% \\
   & support &
  \(149{,}147\) & 1,596 & 4,760 & \(21{,}287\) & 14.27\% \\
   & test &
  \(37{,}942\) & 406 & 4,754 & \(5{,}365\) & 14.14\% \\
  
\bottomrule
\end{tabular}
\end{adjustbox}
\label{tab:dataset_statistics}
\end{table}

\textbf{Dataset sources.}
We evaluate binary differential-expression prediction in five fixed cellular
contexts. The C32 Tahoe100M setting uses single-cell expression profiles from C32
melanoma cells treated with small-molecule perturbations and DMSO controls from
Tahoe-100M~\cite{zhang2025tahoe100m}. Each
example is a perturbation--gene pair \((p,g)\), because the cell context is fixed
to C32. 

The PerturbQA settings include four cell lines: K562, HepG2, Jurkat, and RPE1, using the released perturbation-disjoint splits from PerturbQA~\cite{wu2025perturbqa}. Its perturbations include genetic interventions, and the pipeline preserves the benchmark-provided
binary labels and train/test assignment.

\textbf{C32 preprocessing and DEG labels.}
For C32, we first restrict the expression matrix to the target cell line and to
the 2,000 genes used by the benchmark. Counts are normalized to a total count of
\(10^4\) per cell and log-transformed. For each non-control perturbation, genes
are ranked against the DMSO control group using a Wilcoxon rank-sum test with
Benjamini--Hochberg correction~\cite{wilcoxon1945individual,benjamini1995controlling}. A perturbation--gene pair is labeled positive
\((y=1)\) if its adjusted \(p\)-value is below \(0.05\) and its absolute
log-fold change is greater than \(0.25\). Negative examples \((y=0)\) are drawn
from genes with unadjusted \(p>0.1\), with up to 200 negative genes sampled per
perturbation before support-budget subsampling. Direction labels, when used in
auxiliary analyses, are defined only on positive DEGs by the sign of the
log-fold change; the main results in this paper use the binary DE label.

\textbf{Train/test split strategy.}
All reported splits are perturbation-disjoint: a perturbation appears in either
the support split or the held-out test split, but not both. For the C32 support
sweeps, we use a fixed held-out perturbation set containing 265 perturbations
and \(74{,}642\) test pairs. The remaining 114 perturbations form the maximal
support pool. A support-budget dataset is parameterized by \((k,m)\), where
\(k\) is the number of support perturbations selected from this pool and \(m\)
is the maximum number of labeled genes sampled per support perturbation. If a
support perturbation has fewer than \(m\) eligible labeled genes after the DE
and negative-sampling step, all available labels are retained. The LLM operating
point used for the C32 results is \(k=50,m=200\), giving \(10{,}000\) support
pairs and the same \(74{,}642\) held-out test pairs. In the sharded inference
pipeline, each shard carries all support rows and a disjoint subset of the test
rows, so retrieval and prompting always condition on the same support pool.

For K562 PerturbQA, we use the benchmark's perturbation-disjoint train/test
split directly. We do not re-call DEGs or re-sample support perturbations for
this dataset; the full training split serves as the support pool, and evaluation
is performed on the released held-out perturbations.

\begin{table}[t]
\centering
\small
\caption{Dataset statistics for all Tahoe100M C32 DE configurations. Each row reports the statistics of the full dataset for one $k,m$ configuration; \texttt{Full-support} corresponds to the all available support examples for C32 split. The train/test ratio is computed as the number of training examples divided by the number of test examples. }
\label{tab:c32-all-config-de-stats}
\begin{tabular}{lrrrrr}
\toprule
Configuration & $k$ & $m$ & \#Train examples & \#Test examples & Train/Test (\%) \\
\midrule
\multirow{18}{*}{Sub-sampled}
 & 1   & 40   & 40     & 74,642 & 0.05  \\
 & 1   & 200  & 200    & 74,642 & 0.27  \\
 & 1   & 1000 & 261    & 74,642 & 0.35  \\
 & 2   & 40   & 80     & 74,642 & 0.11  \\
 & 2   & 200  & 400    & 74,642 & 0.54  \\
 & 2   & 1000 & 533    & 74,642 & 0.71  \\
 & 5   & 40   & 200    & 74,642 & 0.27  \\
 & 5   & 200  & 1,000  & 74,642 & 1.34  \\
 & 5   & 1000 & 1,419  & 74,642 & 1.90  \\
 & 20  & 40   & 800    & 74,642 & 1.07  \\
 & 20  & 200  & 4,000  & 74,642 & 5.36  \\
 & 20  & 1000 & 5,674  & 74,642 & 7.60  \\
 & 50  & 40   & 2,000  & 74,642 & 2.68  \\
 & 50  & 200  & 10,000 & 74,642 & 13.40 \\
 & 50  & 1000 & 14,165 & 74,642 & 18.97 \\
 & 114 & 40   & 4,560  & 74,642 & 6.11  \\
 & 114 & 200  & 22,800 & 74,642 & 30.55 \\
 & 114 & 1000 & 32,082 & 74,642 & 42.98 \\
\texttt{Full-support} & -- & -- & 32,082 & 74,642 & 42.98 \\
\bottomrule
\end{tabular}
\end{table}

\textbf{C32 support-budget sweep.}
For deterministic support-scaling experiments, we vary
\(k\in\{1,2,5,20,50,114\}\) and \(m\in\{40,200,1000\}\) while keeping the same
265 held-out perturbations fixed. The realized number of support pairs ranges
from 40 at \(k=1,m=40\) to \(32{,}082\) at \(k=114,m=1000\), the latter using
all available labels from the maximal support pool. The test set is unchanged
throughout the sweep, which makes differences across \((k,m)\) reflect the
amount of observable support evidence rather than a changing evaluation set. We summarize the data statistics for all sweeps in Tab.~\ref{tab:c32-all-config-de-stats}.

\section{Knowledge Graph Resources}
\label{app:kg-resources}

\textbf{KG sources and harmonization.}
CORE uses a harmonized biomedical knowledge graph assembled from VCWorld,
PrimeKG, HetioNet, PertKGE, and BioKG resources~\cite{wei2026vcworld,chandak2023building,himmelstein2017systematic,ni2024pertkge,walsh2020biokg}. Nodes are canonicalized across
sources before edge construction. Drugs and compounds are matched through
source-provided identifiers and normalized names, genes and gene products
through gene symbols and cross-references such as NCBI Gene, Ensembl, and
UniProt identifiers, and functional entities through ontology or pathway
identifiers such as GO, Reactome, and source-native disease, phenotype, and
anatomy identifiers. When multiple sources resolve to the same biomedical
entity, CORE keeps a single canonical node and preserves source-specific
aliases, descriptions, cross-references, and provenance fields.

\begin{figure*}[t]
    \centering
    \includegraphics[width=0.95\textwidth]{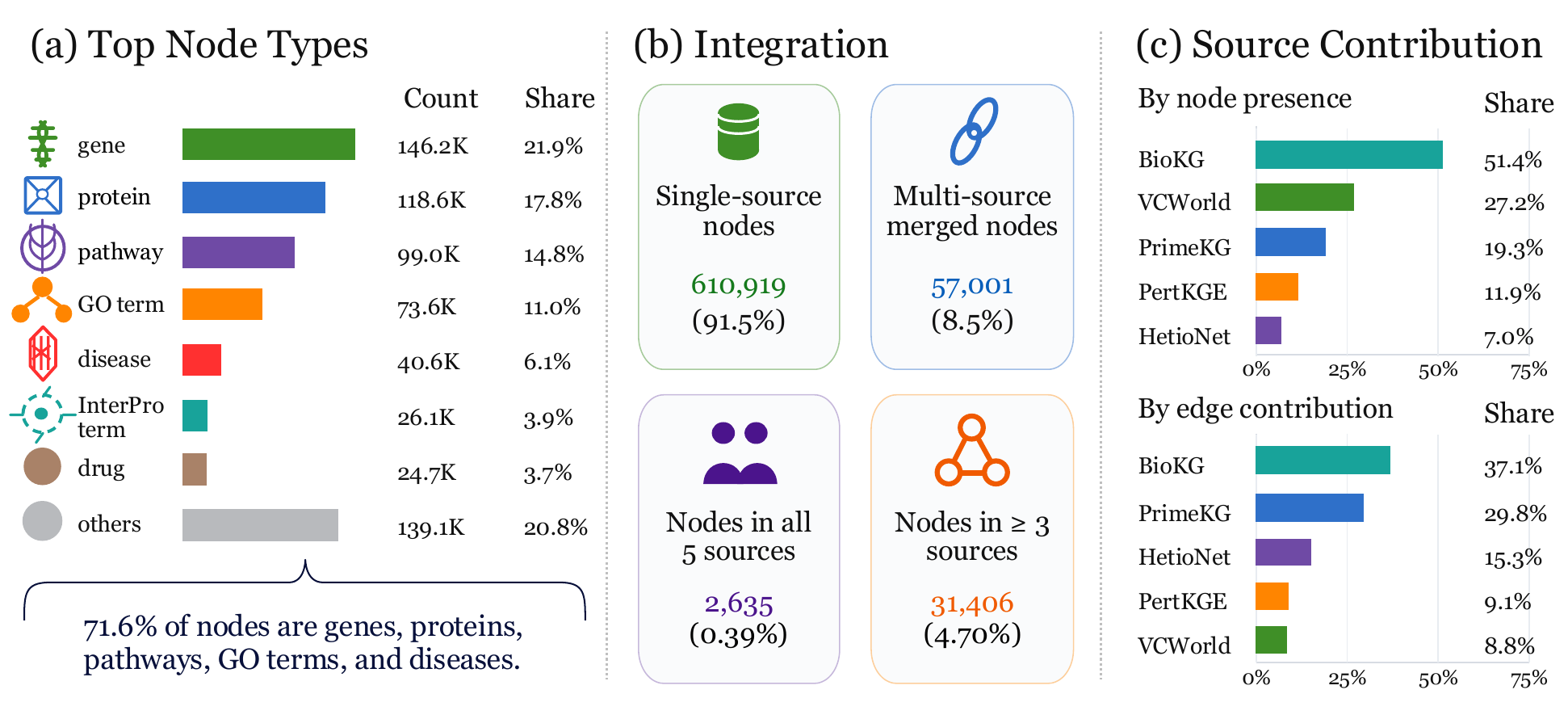}
    \caption{ReasonKG node composition and source integration. The graph is centered on genes, proteins, pathways, GO terms, and diseases, with multi-source nodes serving as cross-resource anchors.}
    \label{fig:reasonkg-node-composition}
\end{figure*}

\textbf{Node type composition.}  Fig.~\ref{fig:reasonkg-node-composition} gives a high-level summary of node type composition and source attributions. Most nodes are core biological entities: genes, proteins, pathways, GO terms, and diseases together account for 71.6\% of all nodes. Although most nodes appear in only one source, 57,001 nodes are merged across multiple sources, including 31,406 nodes found in at least three sources and 2,635 nodes shared by all five sources. This shows that the graph combines broad coverage with a smaller set of highly supported cross-source entities. BioKG provides the largest contribution, while VCWorld, PrimeKG, PertKGE, and HetioNet add complementary nodes and edges.

\textbf{Graph construction.}
The retrieval graph used in the experiments contains \(667{,}920\) canonical
nodes and \(20{,}292{,}372\) directed edge records after source harmonization
and retrieval filtering. Edge provenance is retained; the final edge records
come from PrimeKG (39.9\%), BioKG (25.1\%), PertKGE (12.1\%), VCWorld
(11.8\%), and HetioNet (11.1\%). The graph contains \(57{,}001\) multi-source
nodes (8.5\% of all canonical nodes), reflecting cross-resource merges. For
path retrieval, stored edge directions are preserved and reverse traversals are
shown with inverse-relation markers. For distance features, the audit searches
the traversal graph up to depth six and records whether a perturbation--gene
pair has a usable path within that budget.

\begin{figure*}[t]
    \centering
    \includegraphics[width=0.85\textwidth]{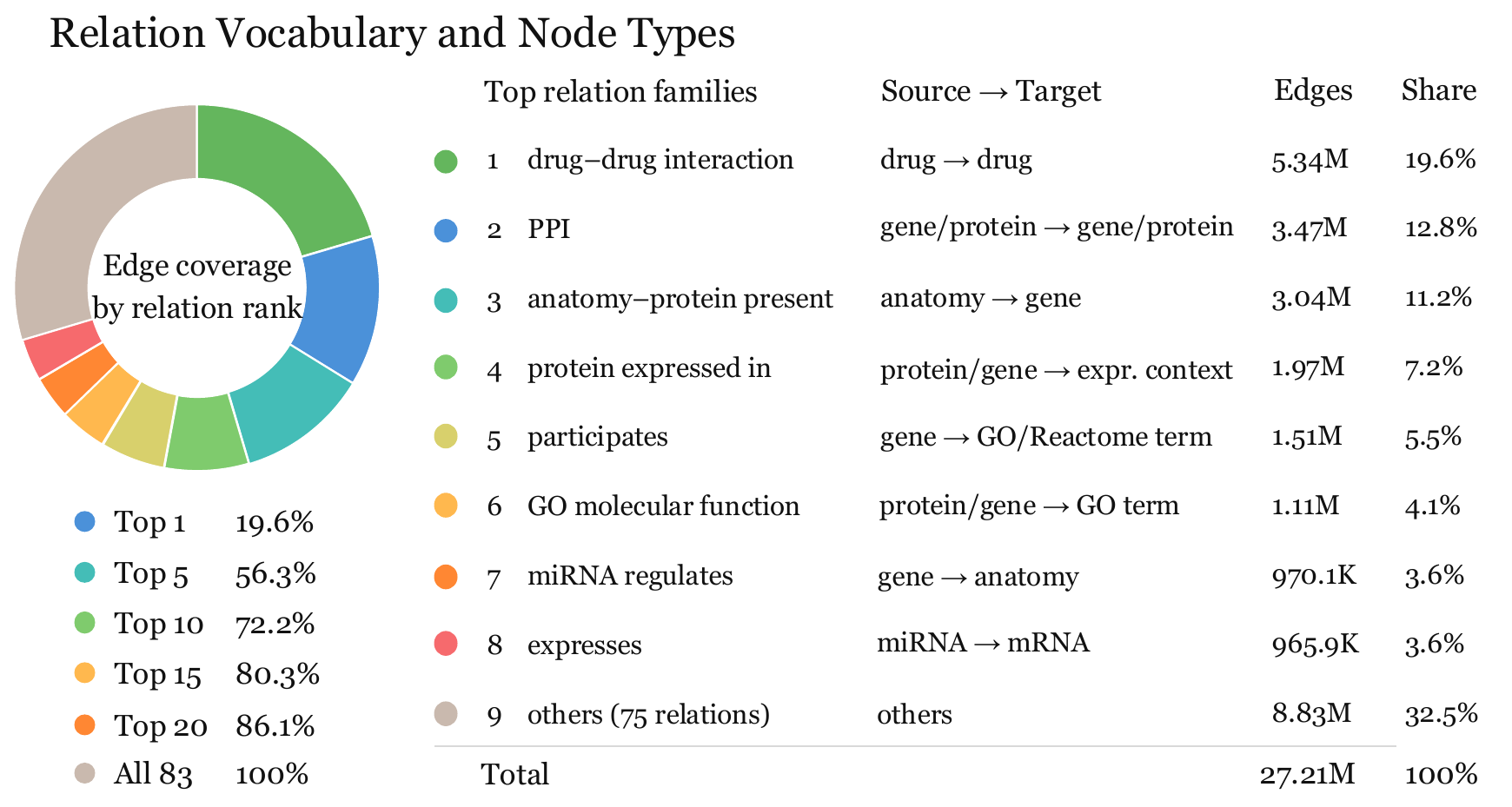}
    \caption{
    ReasonKG relation vocabulary and edge coverage.
    The harmonized graph contains 83 relation families and 27.21M edges, with the largest families including drug--drug interactions, PPI, anatomy--protein links, expression-context links, and pathway/GO participation.
    The edge-coverage summary shows that a small number of high-volume relation families accounts for much of the graph, while the long tail contributes diverse biomedical context.
    }
    \label{fig:reasonkg-relation-vocabulary}
\end{figure*}

\textbf{Relation normalization.}
Raw edge labels are grouped into broad relation families for retrieval auditing
and prompt presentation. These groups preserve biologically meaningful
distinctions while avoiding a long tail of source-specific relation names. The
dominant families are expression, drug--drug interaction, protein--protein
interaction, gene--GO annotation, regulatory interaction, gene--pathway
membership, sequence-domain membership, drug-effect, disease-phenotype, and
disease-gene association. Path prompts display the concrete relation sequence
whenever a shortest path is shown, rather than only the coarse relation family. We show the relation vocabulary and representative edge coverage in Fig.~\ref{fig:reasonkg-relation-vocabulary}.

\begin{figure*}[t]
    \centering
    \includegraphics[width=0.8\textwidth]{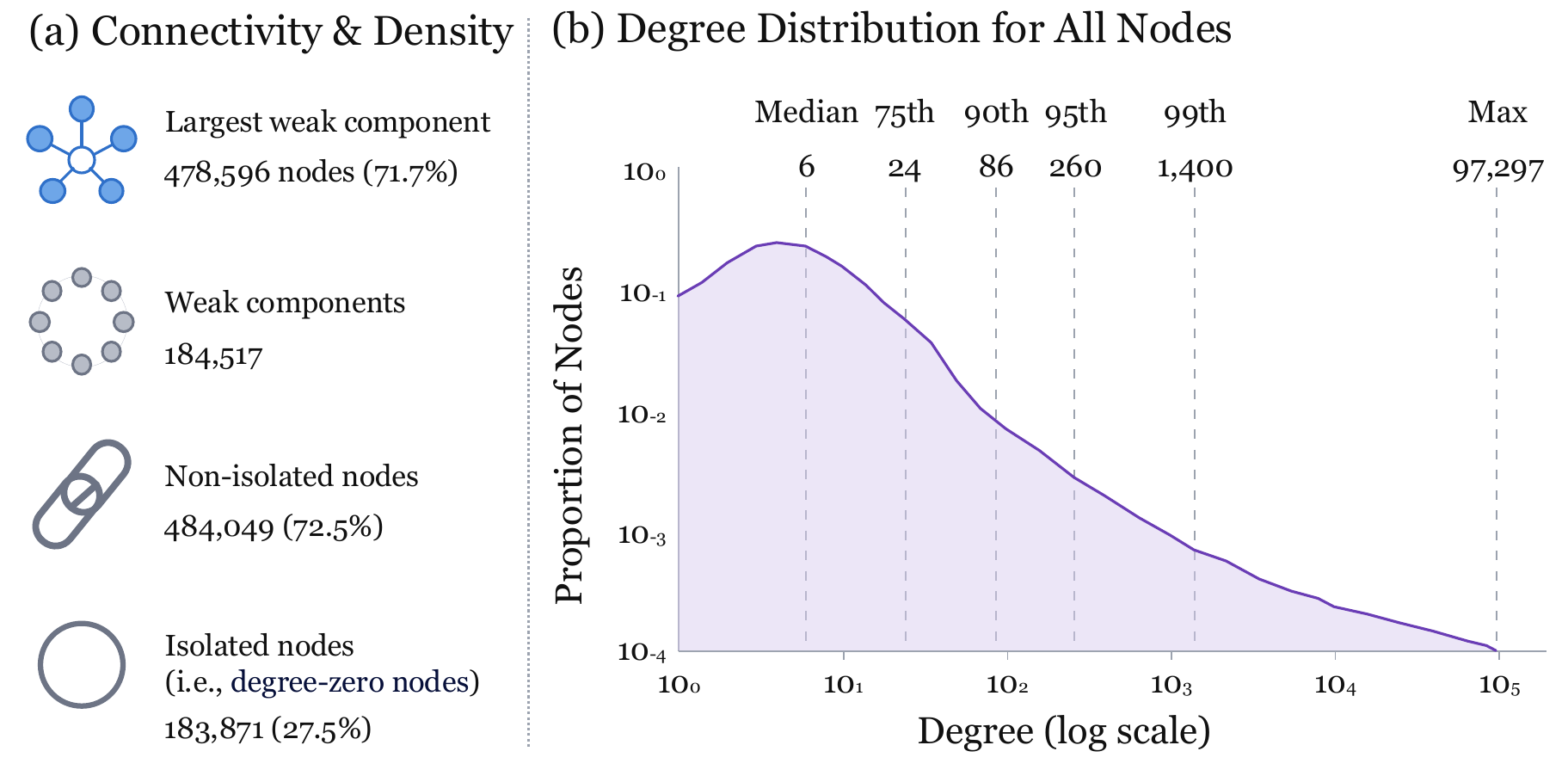}
    \caption{
    Connectivity and degree distribution of ReasonKG.
    The largest weakly connected component contains 478{,}596 nodes, covering 71.7\% of the graph, while 27.5\% of nodes are isolated.
    The degree distribution is heavy-tailed, with many low-degree nodes and a small number of highly connected biomedical hubs. We further analyze the isolated node source attribution in Fig.~\ref{fig:reasonkg-isolated-nodes}.
    }
    \label{fig:reasonkg-connectivity}
\end{figure*}

\textbf{Connectivity and degree distribution.} Fig.~\ref{fig:reasonkg-connectivity} summarizes the connectivity structure of ReasonKG. The largest connected component contains \(478{,}596\) nodes, corresponding to
71.7\% of the retrieval graph. The graph also contains 484,049 non-isolated nodes, and \(183{,}872\) isolated
nodes (27.5\%) (\emph{i.e.}, zero-degree nodes), mostly from source-specific entities that are retained for
identifier coverage but do not contribute to path evidence. The degree distribution is highly long-tailed: most nodes have small degree, with a median degree of 6, but a small number of hub nodes connect to many entities, reaching a maximum degree of 97,297. This structure indicates that the graph contains one large connected biomedical backbone together with many sparse or isolated entities, which is expected when integrating heterogeneous resources with uneven coverage across entity types and sources.

\textbf{Isolated node distribution and source attribution.}

\begin{figure*}[t]
    \centering
    \includegraphics[width=0.68\textwidth]{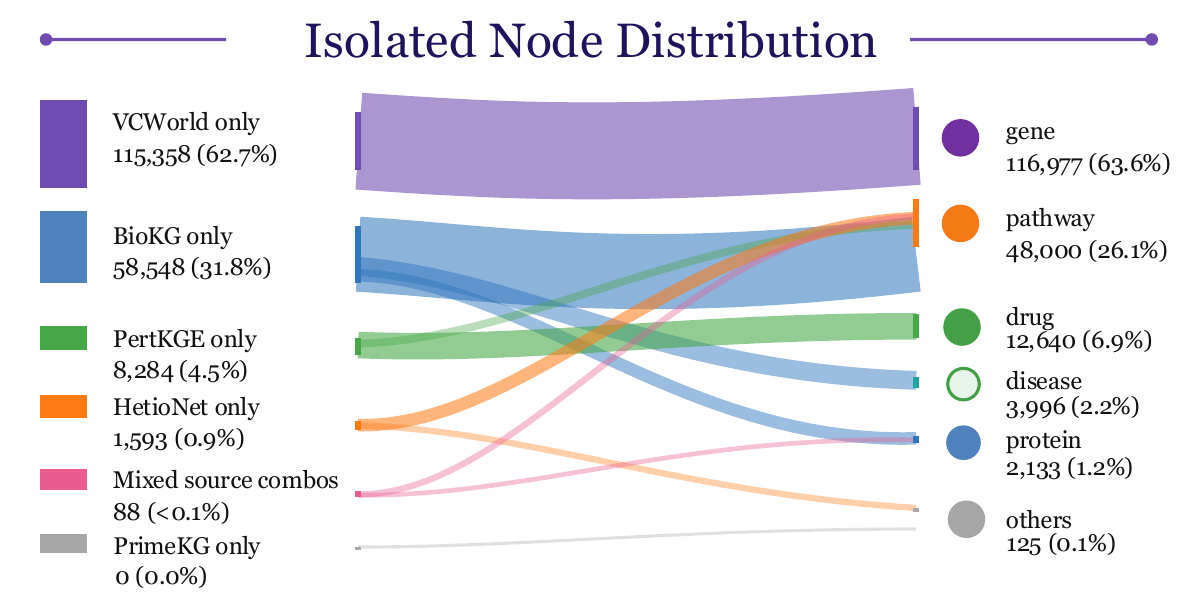}
    \caption{
    Source and type distribution of isolated nodes in ReasonKG.
    Most isolated nodes originate from source-specific entries that lack cross-resource bridging edges, and the majority are genes or pathways.
    This audit identifies regions where KG-based retrieval may be sparse and motivates reporting coverage alongside perturbation predictions.
    }
    \label{fig:reasonkg-isolated-nodes}
\end{figure*}

Fig.~\ref{fig:reasonkg-isolated-nodes} shows where isolated nodes come from and what entity types they represent. Most isolated nodes are source-specific: VCWorld-only nodes account for the largest share, followed by BioKG-only and PertKGE-only nodes, while mixed-source isolated nodes are rare. By type, isolated nodes are dominated by genes and pathways, with smaller contributions from drugs, diseases, proteins, and other entities. This suggests that isolation is mainly caused by source-specific coverage rather than failed cross-source merging: many entities are included because they are present in individual resources, but they do not have edges after integration under the retained relation schema.

\textbf{Benchmark coverage.}
Tab.~\ref{tab:kg_coverage_statistics} summarizes KG endpoint coverage for the
primary binary DE settings. A pair is endpoint-covered when both the query
perturbation and target gene can be grounded to the harmonized graph. A usable
bridge is an endpoint-covered perturbation--gene pair for which the audit finds
at least one path of length at most six. K562 is almost fully covered by the
KG. C32 has high perturbation coverage but much lower gene coverage because
many target genes in the 2,000-gene benchmark panel are absent from the
harmonized KG or are represented only by identifiers that cannot be safely
matched to a canonical gene node.

\begin{table}[h]
\centering
\small
\caption{KG grounding and path-audit coverage for the primary binary DE
settings. Usable bridges are pairs with an audited KG path of length at most
six. Percentages are computed relative to the corresponding dataset split.}
\begin{adjustbox}{max width=\textwidth}
\begin{tabular}{llrrrr}
\toprule
Dataset & Split & Perturbations grounded & Genes grounded
& Endpoint-covered pairs & Usable bridges \\
\midrule
C32 Tahoe100M (k=50,m=200) & support &
\(41/50\) (82.0\%) & \(528/1{,}962\) (26.9\%) &
\(3{,}346/10{,}000\) (33.5\%) &
\(2{,}988/10{,}000\) (29.9\%) \\
C32 Tahoe100M (k=50,m=200) & test &
\(212/265\) (80.0\%) & \(548/2{,}000\) (27.4\%) &
\(24{,}650/74{,}642\) (33.0\%) &
\(23{,}737/74{,}642\) (31.8\%) \\
\midrule
K562 PerturbQA & support &
\(1{,}561/1{,}564\) (99.8\%) & \(4{,}135/4{,}136\) (100.0\%) &
\(134{,}193/134{,}467\) (99.8\%) &
\(118{,}464/134{,}467\) (88.1\%) \\
K562 PerturbQA & test &
\(267/267\) (100.0\%) & \(4{,}117/4{,}118\) (100.0\%) &
\(23{,}206/23{,}212\) (100.0\%) &
\(22{,}655/23{,}212\) (97.6\%) \\
\midrule

HepG2 PerturbQA & support &
  \(1{,}086/1{,}086\) (100.0\%) & \(7{,}432/7{,}435\) (100.0\%) &
  \(101{,}109/101{,}140\) (100.0\%) &
  \(89{,}535/101{,}140\) (88.5\%) \\

 HepG2 PerturbQA & test &
  \(278/278\) (100.0\%) & \(7{,}093/7{,}096\) (100.0\%) &
  \(25{,}737/25{,}749\) (100.0\%) &
  \(25{,}407/25{,}749\) (98.7\%) \\
\midrule
  Jurkat PerturbQA & support &
  \(1{,}225/1{,}227\) (99.8\%) & \(6{,}840/6{,}842\) (100.0\%) &
  \(113{,}465/113{,}684\) (99.8\%) &
  \(100{,}478/113{,}684\) (88.4\%) \\

  Jurkat PerturbQA & test &
  \(313/313\) (100.0\%) & \(6{,}696/6{,}698\) (100.0\%) &
  \(29{,}129/29{,}138\) (100.0\%) &
  \(28{,}519/29{,}138\) (97.9\%) \\
\midrule
  RPE1 PerturbQA & support &
  \(1{,}594/1{,}596\) (99.9\%) & \(4{,}759/4{,}760\) (100.0\%) &
  \(148{,}928/149{,}147\) (99.9\%) &
  \(132{,}280/149{,}147\) (88.7\%) \\

  RPE1 PerturbQA & test &
  \(406/406\) (100.0\%) & \(4{,}753/4{,}754\) (100.0\%) &
  \(37{,}933/37{,}942\) (100.0\%) &
  \(37{,}203/37{,}942\) (98.1\%) \\
  
\bottomrule
\end{tabular}
\end{adjustbox}

\label{tab:kg_coverage_statistics}
\end{table}

\textbf{Handling missing KG evidence.}
Missing KG evidence is represented explicitly rather than imputed as negative
biological evidence. If the gene endpoint is not grounded, CORE omits
query-pair path evidence for that gene and relies on same-gene support evidence
when available, gene-anchor smoothing through response-similar grounded genes,
and the support-set gene prior. If the perturbation endpoint is not grounded,
CORE does not fabricate a direct KG path. The evidence packet marks the missing
endpoint and the LLM prompt states that absent KG connectivity should be treated
as missing information, not as evidence that the perturbation cannot affect the
gene.

\textbf{Handling disconnected or unmapped perturbations.}
Some perturbation names do not resolve to a canonical KG drug node, especially
when salts, hydrates, stereochemical qualifiers, or benchmark-specific naming
variants are present. For these cases, CORE uses a weak name-based fallback:
the query perturbation is mapped to the five most similar KG-covered evaluation
perturbations using character \(n\)-gram TF-IDF similarity, with weights
proportional to the fourth power of the lexical similarity. This fallback is
used only to transfer CORE scores from nearby covered perturbations and is
flagged as indirect evidence in the prompt. It is not treated as a curated
drug-target or mechanism edge.

\begin{figure*}[t]
    \centering
    \includegraphics[width=0.98\textwidth]{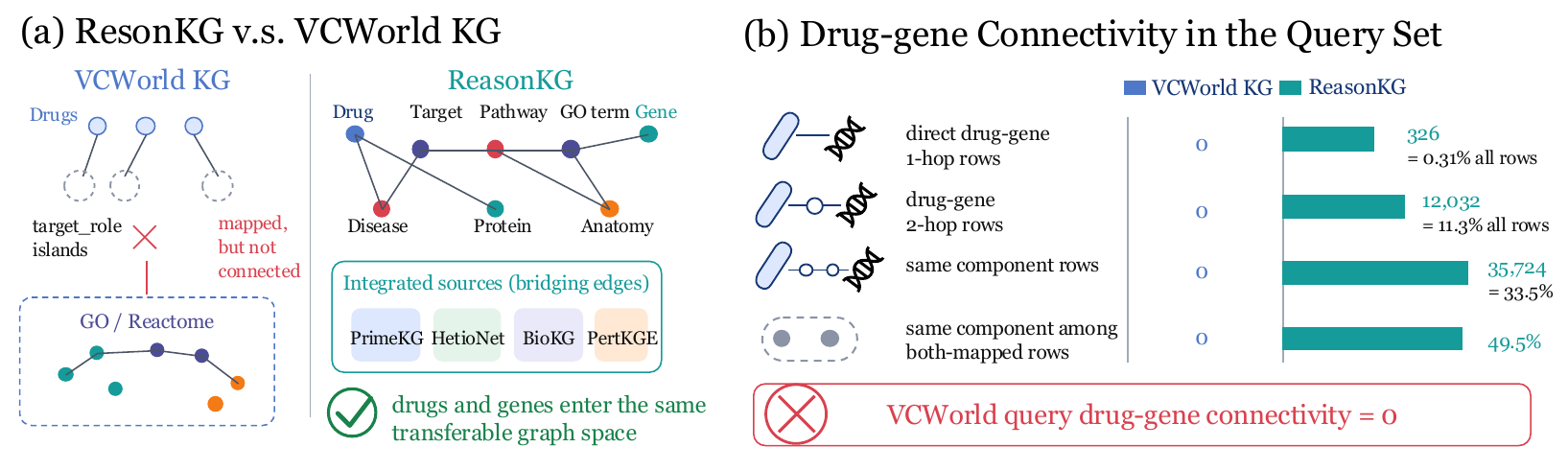}
    \caption{
    Query-set connectivity comparison between ReasonKG and the VCWorld KG.
    In the analyzed perturbation--gene queries, the VCWorld KG leaves query drugs and target genes disconnected, whereas ReasonKG places drugs, genes, pathways, GO terms, diseases, proteins, and anatomy nodes in a shared graph space.
    This integration enables direct, two-hop, and same-component evidence for a substantial fraction of query pairs.
    }
    \label{fig:reasonkg-vcworld-connectivity}
\end{figure*}

\textbf{Comparison between ReasonKG and VCWorld's KG.} Fig.~\ref{fig:reasonkg-vcworld-connectivity} explains why integrating multiple knowledge sources is necessary for perturbation reasoning. In VCWorld KG, drugs and genes can be mapped into the graph but remain disconnected, leaving query drug--gene pairs with no usable graph paths. ReasonKG resolves this by merging VCWorld with additional sources such as PrimeKG, HetioNet, BioKG, and PertKGE, which introduce bridging edges across drugs, diseases, proteins, pathways, GO terms, anatomy, and genes. As a result, ReasonKG connects many query pairs: 326 rows have direct drug--gene links, 12,032 have two-hop links, and 35,724 lie in the same connected component, while VCWorld alone has zero connected query rows. This shows that the integrated graph provides the transferable graph space needed for evidence retrieval and perturbation-specific reasoning.

\section{Method Details}
\label{app:method-details}

\textbf{Notation and inputs.}
For a fixed cell context, let
\(\mathcal S_{k,m}=\{(p,g,y_{p,g})\}\) denote the labeled support set under
support budget \((k,m)\), where \(p\) is a perturbation, \(g\) is a target gene,
and \(y_{p,g}\in\{0,1\}\) indicates whether \(g\) is differentially expressed
under \(p\). Let \(\mathcal P_{\mathrm{sup}}\) be the set of support
perturbations, \(\mathcal P_{\mathrm{eval}}\) be the set of held-out
perturbations to score, and \(\mathcal G\) be the benchmark gene set. CORE uses
ReasonKG only as an evidence-organizing structure: KG paths, distances, and
entity mappings are not treated as labels. The gold label for a held-out query
is never included in the evidence packet.

\subsection{ReasonKG Grounding and Distance Audit}
\label{app:kg-distance-audit}

For every support or evaluation perturbation that can be grounded in ReasonKG,
we audit its graph distance to each grounded benchmark gene. Let
\(d(p,h)\in\{1,\ldots,6,\infty\}\) denote the shortest audited path length from
perturbation \(p\) to gene \(h\), with \(d(p,h)=\infty\) when no path is found
within depth six. The audit is performed separately for support and evaluation
splits so that held-out perturbations remain perturbation-disjoint from the
support labels. These distances serve two purposes: they define sparse
perturbation signatures for support transfer, and they provide compact
shortest-path evidence for \corereasoning prompts.

The empirical association between distance and support labels is estimated only
from the support split:
\[
\rho(d)
=
\frac{
\sum_{(p,h)\in\mathcal S_{k,m}}
\mathbf 1[d(p,h)=d]\,y_{p,h}
}{
\sum_{(p,h)\in\mathcal S_{k,m}}
\mathbf 1[d(p,h)=d]
}.
\]
If a distance bucket is absent in the support split, CORE uses a fixed fallback
bucket value rather than estimating from held-out labels.

\subsection{KG-Derived Perturbation Signatures}
\label{app:kg-signatures}

For each KG-covered perturbation \(p\), CORE constructs a sparse signature
\(\phi(p)\) over grounded genes. For each grounded gene
\(h\in\mathcal G_{\mathrm{KG}}\), the signature includes a distance-identity
feature and a proximity-rate feature:
\[
\phi(p)
=
\left\{
\mathbf 1[\mathrm{dist}:h:d(p,h)],
\quad
\rho(d(p,h))\,\mathbf 1[\mathrm{prox}:h]
\right\}_{h\in\mathcal G_{\mathrm{KG}}}.
\]
In the implementation these are dictionary features such as
\(\texttt{dist:HES1:2}=1\) and \(\texttt{prox:HES1}=\rho(2)\), vectorized into
a sparse feature matrix. This representation lets perturbations be similar
when they have comparable graph-distance patterns to response-relevant genes,
not merely when their names are textually similar.

\subsection{Perturbation Similarity and Contrastive Retrieval}
\label{app:contrastive-retrieval}

Given a held-out perturbation \(p\) and a support perturbation
\(p'\in\mathcal P_{\mathrm{sup}}\), CORE computes a clipped cosine similarity
between KG-derived signatures:
\[
\mathrm{sim}(p,p')
=
\max\{0,\cos(\phi(p),\phi(p'))\}.
\]
The similarity is converted into a nonnegative vote weight
\[
w(p,p')=\mathrm{sim}(p,p')^\eta,
\]
with \(\eta=0.5\) in the reported experiments. The main text denotes the
retrieved perturbation set by \(\mathcal R_k(p)\). Operationally, the numerical
{\corevoting} score aggregates all support perturbations with observed labels and
nonzero weights; the prompt displays only the top \(L=3\) positive and top
\(L=3\) negative cases for readability.

For a query \((p,g)\), CORE keeps the target gene fixed and partitions the
weighted support evidence by the observed outcome for that same gene:
\[
\mathcal H^+_{p,g}
=
\{p'\in\mathcal P_{\mathrm{sup}}:
m_{p',g}=1,\; y_{p',g}=1,\; w(p,p')>0\},
\]
\[
\mathcal H^-_{p,g}
=
\{p'\in\mathcal P_{\mathrm{sup}}:
m_{p',g}=1,\; y_{p',g}=0,\; w(p,p')>0\},
\]
where \(m_{p',g}\) indicates whether the support label for \((p',g)\) is
observed. This fixed-gene partition is the key contrastive step: positive cases
show related perturbations that affected \(g\), while negative cases show
related perturbations that did not.

\subsection{{\corevoting}}
\label{app:core-voting-details}

CORE summarizes the contrastive support evidence through positive and negative
vote masses:
\[
V^+_{p,g}
=
\sum_{p'\in\mathcal H^+_{p,g}} w(p,p'),
\qquad
V^-_{p,g}
=
\sum_{p'\in\mathcal H^-_{p,g}} w(p,p').
\]
The same-gene contrast score is the weighted positive fraction:
\[
s_{\mathrm{contrast}}(p,g)
=
\begin{cases}
\dfrac{V^+_{p,g}}{V^+_{p,g}+V^-_{p,g}},
& V^+_{p,g}+V^-_{p,g}>0,\\[1.2ex]
\pi_g, & \text{otherwise}.
\end{cases}
\]
Here \(\pi_g\) is the support-set gene prior,
\[
\pi_g
=
\frac{
\sum_{p'\in\mathcal P_{\mathrm{sup}}}
m_{p',g}y_{p',g}
}{
\sum_{p'\in\mathcal P_{\mathrm{sup}}}
m_{p',g}
},
\]
with a global support positive rate used only when \(g\) has no observed
support labels. The evidence packet records \(V^+_{p,g}\), \(V^-_{p,g}\), and
their sum as the observed vote mass, because low vote mass indicates that the
same-gene contrast score is weakly supported.

\subsection{Gene-Anchor Smoothing}
\label{app:gene-anchor-smoothing}

To improve robustness when same-gene support is sparse, CORE also borrows
evidence from response-similar genes. For each gene \(g\), define a signed
support response profile over support perturbations:
\[
r_g(p')
=
\begin{cases}
2y_{p',g}-1, & \text{if } (p',g,y_{p',g})\in\mathcal S_{k,m},\\
0, & \text{otherwise}.
\end{cases}
\]
Gene-gene similarity is computed as clipped cosine similarity:
\[
\mathrm{sim}(g,h)
=
\max\{0,\cos(r_g,r_h)\}.
\]
For each target gene, CORE selects the top \(H\) grounded anchor genes,
\[
\mathcal A_H(g)
=
\operatorname{TopH}_{h\in\mathcal G_{\mathrm{KG}}}
\mathrm{sim}(g,h).
\]
We use \(H=10\) for the numerical anchor score. If the target gene itself is
KG-grounded, it is forced to be the strongest anchor with similarity \(1\).

The anchor-smoothed score is
\[
s_{\mathrm{anchor}}(p,g)
=
\frac{
\sum_{h\in\mathcal A_H(g)}
\mathrm{sim}(g,h)\,s_{\mathrm{contrast}}(p,h)
}{
\sum_{h\in\mathcal A_H(g)}
\mathrm{sim}(g,h)
}.
\]
If the anchor weights sum to zero, CORE sets
\(s_{\mathrm{anchor}}(p,g)=s_{\mathrm{contrast}}(p,g)\). The final
{\corevoting} score is a convex combination:
\[
s_{\mathrm{CORE}}(p,g)
=
(1-\alpha)s_{\mathrm{contrast}}(p,g)
+\alpha s_{\mathrm{anchor}}(p,g).
\]
We use \(\alpha=0.5\) in the reported experiments. For prompting, CORE displays
the top three anchor genes, while the numerical anchor score uses all \(H=10\)
anchors.

\subsection{Fallback for Unmapped Perturbations}
\label{app:unmapped-perturbations}

If a raw evaluation perturbation \(p\) cannot be grounded directly in ReasonKG,
CORE cannot construct \(\phi(p)\). In this case, it uses a weak name-based
fallback to transfer scores from KG-covered evaluation perturbations rather
than inventing KG paths for \(p\). Character \(n\)-gram TF-IDF similarity is
used to find the top five KG-covered evaluation perturbations:
\[
\mathcal M_5(p)
=
\operatorname{Top5}_{q\in\mathcal P_{\mathrm{eval}}^{\mathrm{KG}}}
\mathrm{sim}_{\mathrm{char}}(p,q).
\]
The mapped perturbations receive sharpened lexical weights
\[
\beta(p,q)
=
\frac{
\max\{\mathrm{sim}_{\mathrm{char}}(p,q),10^{-6}\}^{4}
}{
\sum_{q'\in\mathcal M_5(p)}
\max\{\mathrm{sim}_{\mathrm{char}}(p,q'),10^{-6}\}^{4}
}.
\]
The fallback contrast score is then
\[
s_{\mathrm{contrast}}(p,g)
=
\sum_{q\in\mathcal M_5(p)}
\beta(p,q)s_{\mathrm{contrast}}(q,g),
\]
and the same weighted transfer is applied to the anchor and final CORE scores.
This mapping is explicitly marked in \corereasoning prompts as weak indirect
evidence: character similarity is not treated as chemical similarity,
target-profile similarity, or a curated KG edge.

\subsection{{\corereasoning} Path Evidence}
\label{app:kg-path-evidence}

\corereasoning uses the same CORE evidence packet but serializes it for an LLM.
KG paths are retrieved for the query pair \((p,g)\), not for each support
perturbation. When both endpoints are grounded, the path module searches for
shortest paths between the query perturbation endpoint and the target gene
endpoint. If an endpoint is not directly available, the prompt may display a
path through a mapped perturbation endpoint or an anchor-gene endpoint; such
paths are labeled as indirect evidence.

Each displayed path \(\gamma\) receives a compactness score
\[
\mathrm{pathscore}(\gamma)=\frac{1}{|\gamma|},
\]
where \(|\gamma|\) is the number of edges. This value is used only to rank
paths for display. It is not calibrated as a probability, is not part of the
{\corevoting} score, and the prompt explicitly warns the LLM not to treat shorter
paths as sufficient positive evidence.

Thus, {\corevoting} and \corereasoning share evidence construction but use it
differently. {\corevoting} directly aggregates the support evidence numerically,
whereas \corereasoning asks the LLM to interpret the same contrastive evidence,
judge path specificity, and make only query-specific calibrated adjustments to
the default CORE score.

\subsection{Retrieved KG Path Examples}
\label{app:kg-path-examples}

The KG path component is used for explanation and LLM context, not for the
numerical {\corevoting} score. Tab.~\ref{tab:kg_path_examples} shows two
representative path formats. A direct-endpoint path uses the query perturbation
and query gene as the path endpoints. A mapped-endpoint path is used only when
one endpoint cannot be directly grounded; in that case the prompt explicitly
states the substituted endpoint and treats the path as indirect evidence.

\begin{table}[h]
\centering
\small
\caption{Representative KG paths displayed in {\corereasoning} prompts. Path
relations are shown in the stored graph direction. Inverse-relation markers
indicate that the traversed path uses the reverse of a stored directed edge.}
\label{tab:kg_path_examples}
\begin{adjustbox}{max width=\textwidth}
\begin{tabular}{p{0.22\textwidth}p{0.26\textwidth}p{0.47\textwidth}}
\toprule
Query pair & Endpoint type & Example displayed path \\
\midrule
Canagliflozin (hemihydrate), NUP58
& Direct query drug and direct query gene
& \(\mathrm{mRNA{:}NUP58}\)--[\(\mathrm{INV{:}{:}upregulates}\)]--\(\mathrm{PERT{:}CLOFARABINE}\)--[\(\mathrm{drug\_drug\_interaction}\)]--\(\mathrm{PERT{:}CANAGLIFLOZIN}\). \\

Bestatin (hydrochloride), HSP90B1
& Query drug mapped to a grounded perturbation; direct query gene
& \(\mathrm{mRNA{:}HSP90B1}\)
  --[\(\mathrm{INV{:}{:}ppi}\)]--
  \(\mathrm{mRNA{:}AKT1}\)
  --[\(\mathrm{INV{:}{:}downregulates}\)]--
  \(\mathrm{PERT{:}CRIZOTINIB}\). \\
\bottomrule
\end{tabular}
\end{adjustbox}
\end{table}

For each query, up to five compacted shortest paths are displayed. Direction
variants with the same mapped endpoints and intermediate entities are merged
before presentation. The path-length rank score \(1/|\gamma|\) is shown only to
rank paths by compactness; the prompt explicitly warns that this score is not a
probability and should not override the same-gene support evidence.

\subsection{Prompt Design for Evidence Serialization}
\label{app:prompt-design}

\corereasoning uses a structured evidence packet rather than a free-form
retrieval dump. The packet contains the following fields.

\begin{enumerate}
    \item \textbf{Task and query context.} The prompt states the fixed cell
    context, query perturbation, and query gene, together with textual
    descriptions of the perturbation and gene when available.
    \item \textbf{Gene prior.} The support-set positive rate for the target gene
    is displayed as a calibration reference, followed by an explicit warning
    that the prior alone is insufficient because the prediction must distinguish
    perturbations for the same gene.
    \item \textbf{Same-gene support evidence.} The packet reports the
    same-gene contrast score, positive vote mass, negative vote mass, and
    observed vote mass. It then lists up to three top-weighted positive and up
    to three top-weighted negative support perturbations for the same target
    gene, including their KG similarity and vote weight.
    \item \textbf{KG path evidence.} The packet includes up to five shortest KG
    paths between the grounded or mapped query endpoints. Mapped endpoints are
    explicitly marked so that the LLM can down-weight indirect evidence.
    \item \textbf{Gene-anchor evidence.} The packet displays the anchor score,
    the final blended score, and the top displayed anchor genes. In the main
    experiments, the prompt displays three anchor genes, while the numerical
    anchor score is computed from the top ten anchors.
    \item \textbf{Output contract.} Before the final deterministic yes/no
    sentence, the model must output a one-line JSON object with fields
    \texttt{probability}, \texttt{label}, \texttt{default\_final\_score},
    \texttt{support\_score}, \texttt{anchor\_score},
    \texttt{probability\_adjustment}, and \texttt{decisive\_evidence}.
\end{enumerate}

The prompt asks the model to compare positive and negative same-gene cases
before constructing a biological rationale. This design is intended to prevent
the model from treating any plausible perturbation--gene path as sufficient
positive evidence. It also asks the model to preserve the within-gene ordering
implied by the CORE final score unless the evidence packet gives a clear reason
for a small calibrated adjustment.

\subsection{{\corereasoning} Evidence Packet Example}
\label{app:evidence-packet-example}

For the query Bestatin (hydrochloride)--HSP90B1, the evidence packet illustrates
how CORE separates numerical support transfer from KG-path explanation. The
same-gene contrast score is \(0.882\), with positive vote mass \(6.277\) and
negative vote mass \(0.839\). The gene-anchor score is \(0.977\), giving a final
blended score of \(0.929\). The displayed positive same-gene supports include
Ponatinib and Cytarabine; the displayed negative supports include
Hydroxyfasudil and LY-2584702 (tosylate salt). The anchor block lists HSP90B1
itself and response-similar genes such as TPX2 and LINC01320. Because Bestatin
is not directly covered by the KG audit in this case, the KG path block uses a
mapped perturbation endpoint and explicitly marks the path as indirect evidence.

The resulting prompt asks the LLM to first decide whether the query perturbation
is closer to the positive or negative same-gene supports, then judge whether the
KG paths are mechanism-bearing or generic connectivity, and finally report a
calibrated probability. Thus, the LLM receives a constrained reasoning problem:
it may provide mechanistic interpretation, but the default probability and the
contrastive support cases are fixed before generation.

\section{Experimental Details}
\label{app:experimental-details}

\subsection{Benchmarking Settings}
\label{app:benchmarking-settings}

\textbf{Prediction score calculation.} For {\corevoting}, the reported
score is the deterministic blended probability. For {\corereasoning}, the
reported score is the parsed probability emitted by the model, with the
{\corevoting} score retained separately as the default calibration value in the
prompt record.

\textbf{Train/Test split setting.} For the Tahoe100M C32 support-budget sweep, all rows use the same held-out perturbation
set and differ only in the support budget \((k,m)\). For PerturbQA K562 data, the
benchmark train/test split is fixed and the full training split is used as the
support pool. Appendix~\ref{app:dataset-details} provides the preprocessing details.

\subsection{LLM Inference Protocol}
\label{app:llm-inference-protocol}

\textbf{Prompt generation and sharding.}
For each dataset split and prompt variant, we first materialize all query
prompts before running model inference. The sharding procedure is label-safe:
every shard contains the full support set and a disjoint subset of held-out
test queries, so all queries in a run condition on the same support pool. The
primary C32 {\corereasoning} run uses the \(k=50,m=200\) support budget, the
test split only for evaluation queries. The random
seed is fixed to 42 for support sampling, retrieval-side sampling, and prompt
construction.

\textbf{Models and backend.}
The main local LLM results use Qwen3.5-9B~\cite{qwen2026qwen35} with SGLang
inference~\cite{zheng2024sglang}. Unless otherwise stated, each prompt is decoded once. The
SGLang engine is initialized with automatic weight dtype selection, tensor
parallel size 1, static memory fraction 0.80, and maximum context length
16,386 tokens. The submitted C32 runs use the Triton attention backend and
PyTorch sampling backend, with CUDA graph capture disabled for both full and
piecewise graphs. Each inference shard runs on one GPU with batch size 8.
Prompts are rendered with the model chat template, a generation prompt is
appended, and rendered inputs are truncated to at most 8,192 input tokens.

\textbf{Decoding parameters.}
For the sharded SGLang runs, we use temperature \(0.6\), top-\(p=0.9\), and a
maximum of 8,192 generated tokens. These settings are shared across the
standard VCWorld-style prompt and the {\corereasoning} prompt unless explicitly
reported otherwise. {\corevoting} does not invoke an LLM; it directly evaluates
the deterministic blended score described in the method details.

\textbf{Output format and parsing.}
{\corereasoning} prompts require the model to emit a one-line structured
prediction before the final deterministic answer. For binary DE prediction, the
valid labels are \texttt{yes}, \texttt{no}, and \texttt{insufficient}. The
parser first searches for a JSON object in the marked structured-prediction
block, fenced JSON blocks, or the final raw response. It then reads the
\texttt{label} field and a continuous \texttt{probability} field when present.
Probabilities reported on the \(0\)--\(100\) scale are normalized to
\([0,1]\); probabilities outside the valid range are ignored. If structured
JSON is absent or malformed, the parser falls back to the final-answer text and
then to explicit probability phrases in the response. A \texttt{yes} label is
mapped to prediction 1, a \texttt{no} label is mapped to prediction 0, and an
\texttt{insufficient} or unparsable response is treated as unanswered. When a
valid continuous probability is unavailable but a valid binary label is
available, the binary label is used as the score for thresholded metrics.

\subsection{Evaluation Metrics}
\label{app:evaluation-metrics}

\textbf{Instance-level binary metrics.}
Let \(y_i\in\{0,1\}\) be the held-out binary DE label for query pair \(i\),
\(\hat y_i\in\{0,1\}\) be the parsed discrete prediction, and \(s_i\in[0,1]\)
be the parsed or deterministic continuous score. {\corevoting} uses
\(s_i=\hat y_{\mathrm{CORE}}(p_i,g_i)\). {\corereasoning} uses the parsed LLM
probability when available. Accuracy, precision, recall, and F1 are computed on
answered examples only:
\[
\mathrm{precision}=\frac{\mathrm{TP}}{\mathrm{TP}+\mathrm{FP}},
\qquad
\mathrm{recall}=\frac{\mathrm{TP}}{\mathrm{TP}+\mathrm{FN}},
\qquad
\mathrm{F1}=\frac{2\,\mathrm{precision}\,\mathrm{recall}}
{\mathrm{precision}+\mathrm{recall}}.
\]
When a denominator is zero, the corresponding precision, recall, or F1 term is
set to zero. We also report balanced accuracy when thresholded probabilistic
predictions are evaluated.

\textbf{Ranking metrics.}
AUROC and AUPRC are computed from the continuous scores \(s_i\) on answered
examples. AUROC is the probability that a randomly selected positive example is
ranked above a randomly selected negative example, with ties counted as one
half. AUPRC is computed as average precision over the score-ranked test set.
These ranking metrics are reported only when the answered subset contains both
positive and negative labels. For deterministic voting methods, all held-out
examples are answered, so the ranking metrics are computed over the complete
test split.

\textbf{Coverage and effective metrics.}
Because LLM outputs can be malformed or explicitly abstain, we report the
answered rate
\[
r_{\mathrm{ans}}=\frac{|\mathcal A|}{N},
\]
where \(N\) is the number of labeled evaluation examples and \(\mathcal A\) is
the answered subset. The complementary abstention rate is
\[
q=1-r_{\mathrm{ans}}.
\]
For metrics computed on answered examples, we additionally report effective
metrics by multiplying the answered-only value by \(r_{\mathrm{ans}}\). This
prevents a method from appearing strong by answering only a small, easy subset
of queries.

\textbf{Macro per-gene metrics.}
In addition to micro-averaged instance metrics, we compute per-gene
discrimination metrics by grouping held-out examples by target gene. For each
gene with at least one positive and one negative answered example, AUROC, AUPRC,
accuracy, balanced accuracy, precision, recall, F1, specificity, predicted
positive rate, and mean probability are computed within that gene. Macro
per-gene values are the unweighted averages over evaluable genes. Genes with
only one class in the answered subset are excluded from per-gene discrimination
metrics and counted in the skipped-gene statistic.

\subsection{Extended Quantitative Results}
\label{app:extended-results}

\subsubsection{Performance on All Metrics on Tahoe100M C32 and PerturbQA K562}

\begin{figure*}[t]
    \centering
    \includegraphics[width=0.98\textwidth]{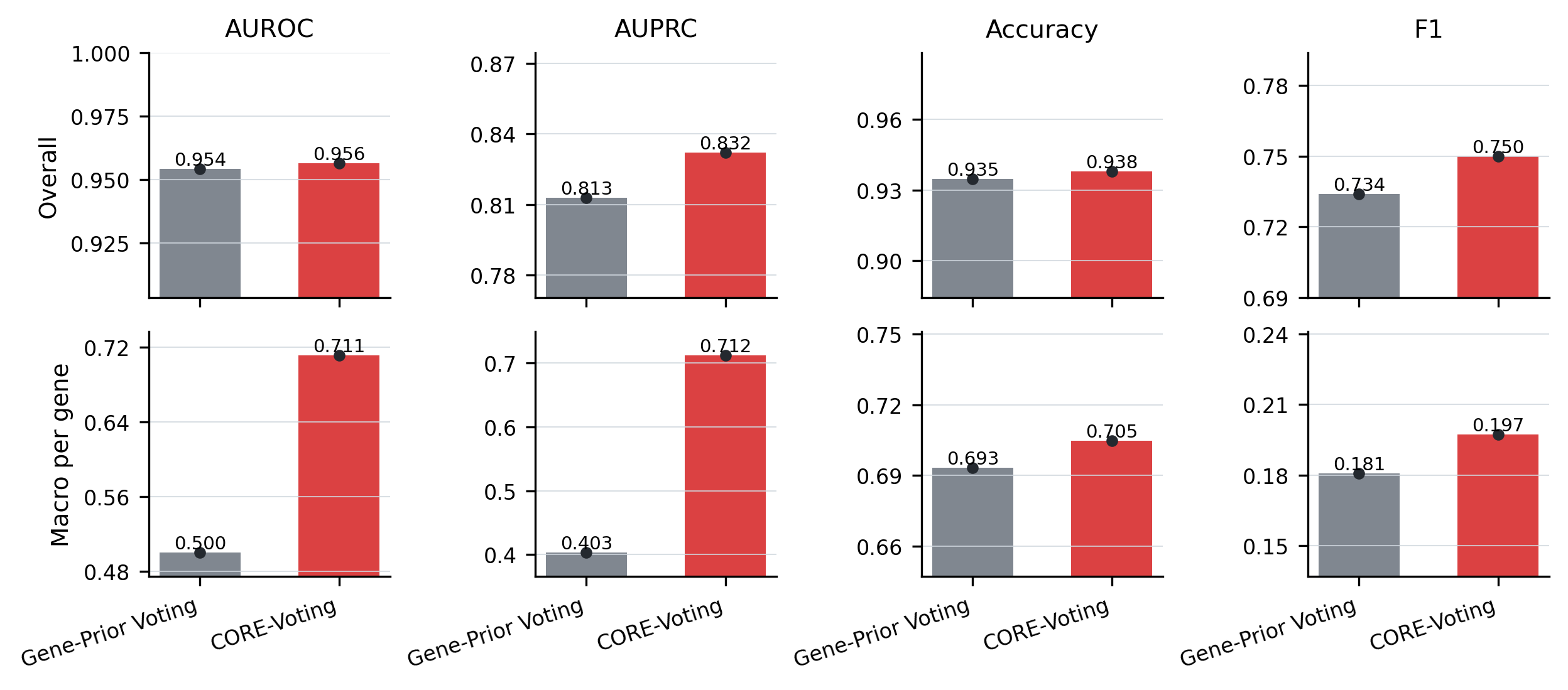}
    \caption{
    Full metrics for comparing {\geneprior} and {
    \corevoting} on PerturbQA K562 data (Partial results are also shown in Fig.~\ref{fig:k562_core_voting_auroc_auprc}).
    }
    \label{fig:k562_core_voting_all_metrics}
\end{figure*}

\textbf{Comparison between {\geneprior} and {\corevoting} on PerturbQA K562.}  Fig.~\ref{fig:k562_core_voting_all_metrics} additionally reports accuracy and F1 for comparison between {\geneprior} and {\corevoting} on PerturbQA K562, as a complement to Fig.~\ref{fig:k562_core_voting_auroc_auprc}. These metrics show the same direction of improvement, with {\corevoting} increasing overall accuracy from 0.935 to 0.938 and F1 from 0.734 to 0.750, and macro per-gene accuracy from 0.693 to 0.705 and F1 from 0.181 to 0.197. The smaller magnitude of these gains, compared with AUROC and AUPRC, suggests that the main benefit of {\corevoting} is improved ranking and calibration of perturbation--gene evidence, while thresholded performance is partly saturated by the strong gene-prior baseline.

\begin{figure*}[t]
    \centering
    \includegraphics[width=0.98\textwidth]{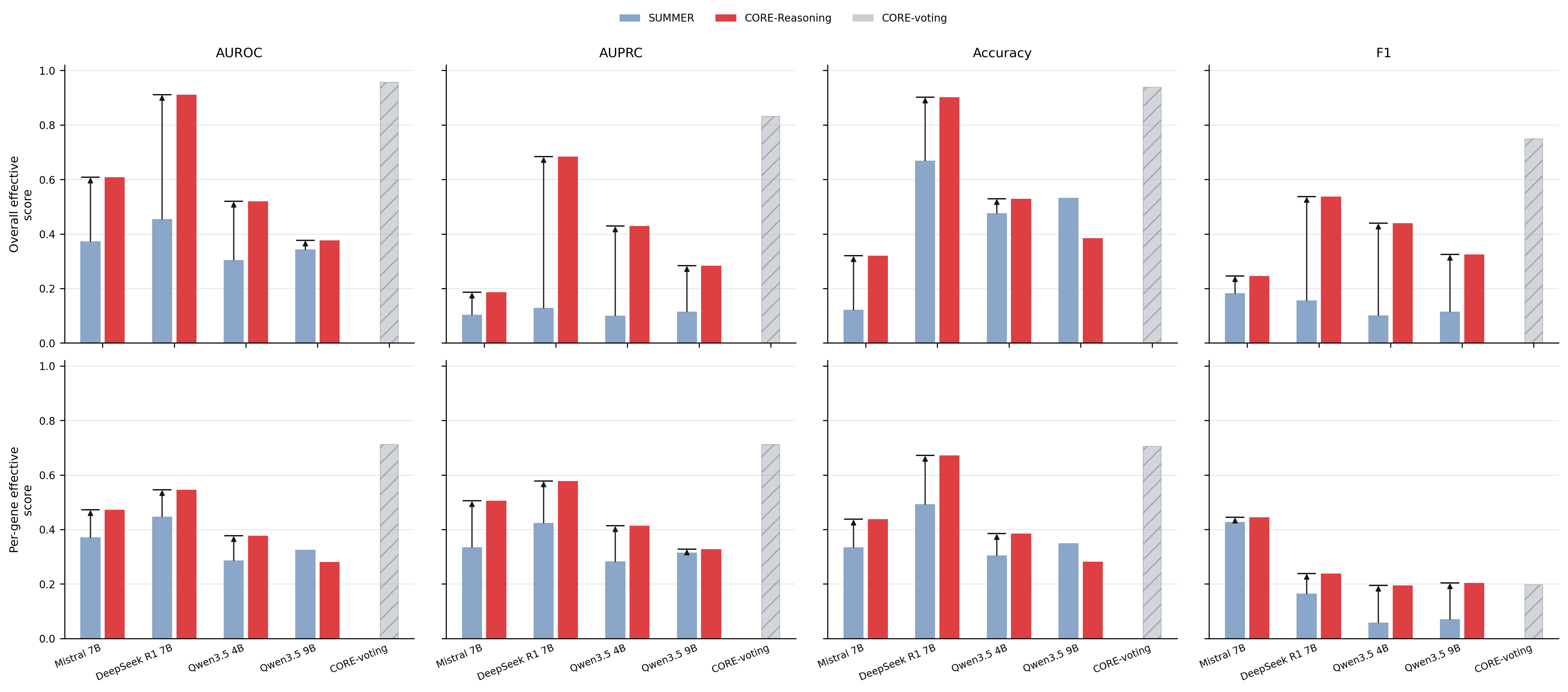}
    \caption{
    PerturbQA K562 comparison between SUMMER-style prompting,
    \corereasoning, and \corevoting. \corereasoning improves the standard prompt
    for several backbones, especially on ranking-oriented metrics, but gains are
    less uniform than in C32. The deterministic \corevoting baseline is strongest
    in this high-coverage, full-support setting.
    }
    \label{fig:k562_llm_metrics}
\end{figure*}

\textbf{Comparison between {\corereasoning} and SUMMER across LLM variants for PerturbQA K562.} Fig.~\ref{fig:k562_llm_metrics} shows that on PerturbQA K562, {\corereasoning} improves SUMMER-style prompting for several backbones, especially on AUROC and AUPRC. This suggests that contrastive same-gene evidence helps LLMs rank perturbation-gene effects more reliably than isolated plausibility prompts. However, improvements are not uniform across all metrics: thresholded metrics such as Accuracy and F1 can still degrade for some backbones, indicating remaining calibration sensitivity. The strong {\corevoting} results further show that, in this high-coverage full-support setting, much of the predictive signal is already present in the retrieved support labels; the LLM mainly helps when it can interpret that evidence without introducing calibration noise.

\begin{figure}[t]
\centering
\begin{subfigure}[t]{0.45\linewidth}
    \centering
    \includegraphics[width=\linewidth]{figures/experiment/k50_m200_de_yes_ratio_with_core.png}
    \caption{Predicted-positive rate.}
\end{subfigure}
\hfill
\begin{subfigure}[t]{0.45\linewidth}
    \centering
    \includegraphics[width=\linewidth]{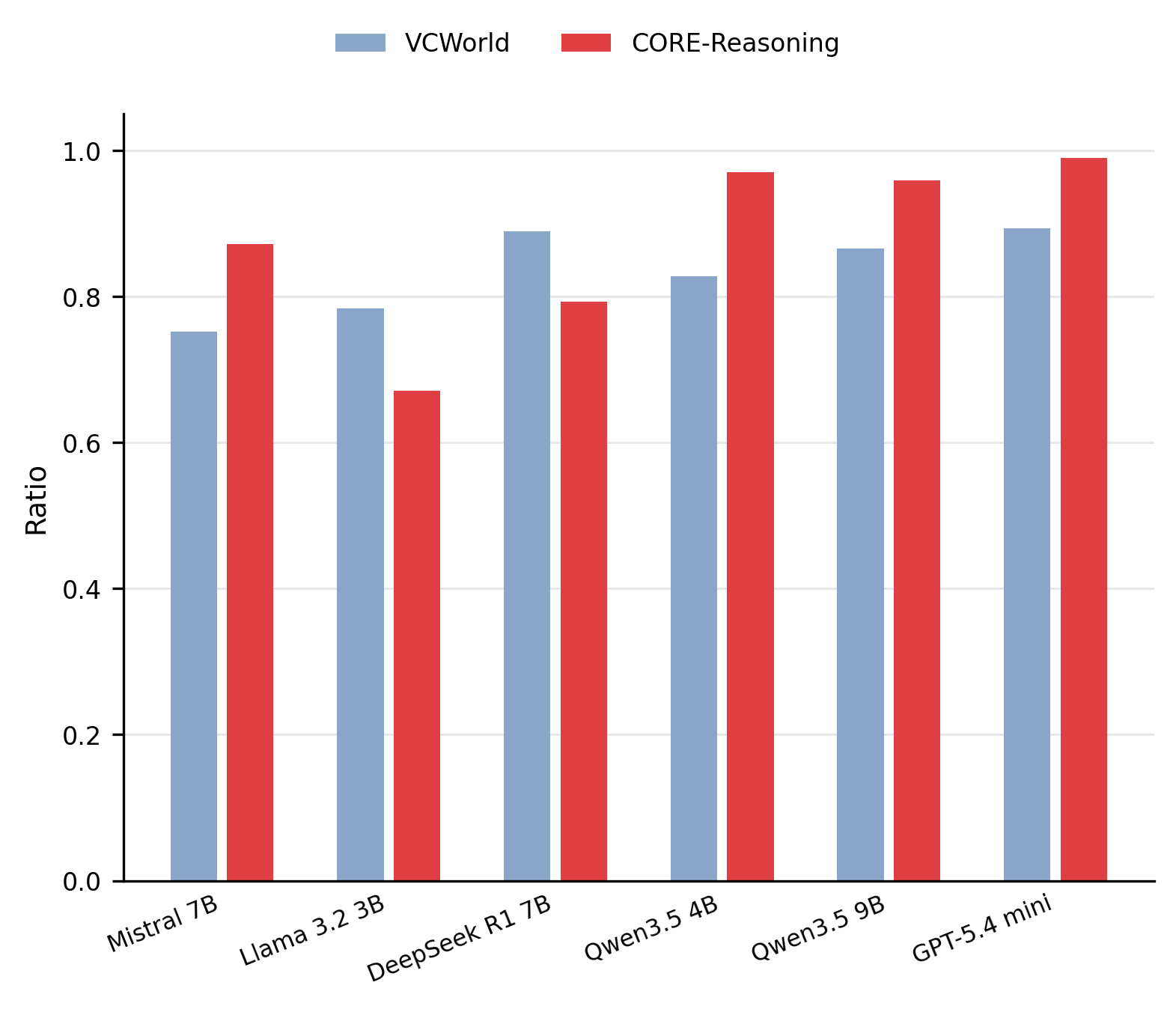}
    \caption{Effective answer rate.}
\end{subfigure}
\caption{
Calibration and output coverage for C32 fixed-budget LLM runs.
The dashed line in the left panel is the empirical test positive rate.
Standard VCWorld-style prompts can be strongly biased toward positive
predictions, whereas \corereasoning moves the predicted-yes rate toward the
empirical prevalence while maintaining high answer rates for the Qwen3.5 runs.
}
\label{fig:c32_llm_calibration}
\end{figure}

\begin{figure}[t]
\centering
\begin{subfigure}[t]{0.45\linewidth}
    \centering
    \includegraphics[width=\linewidth]{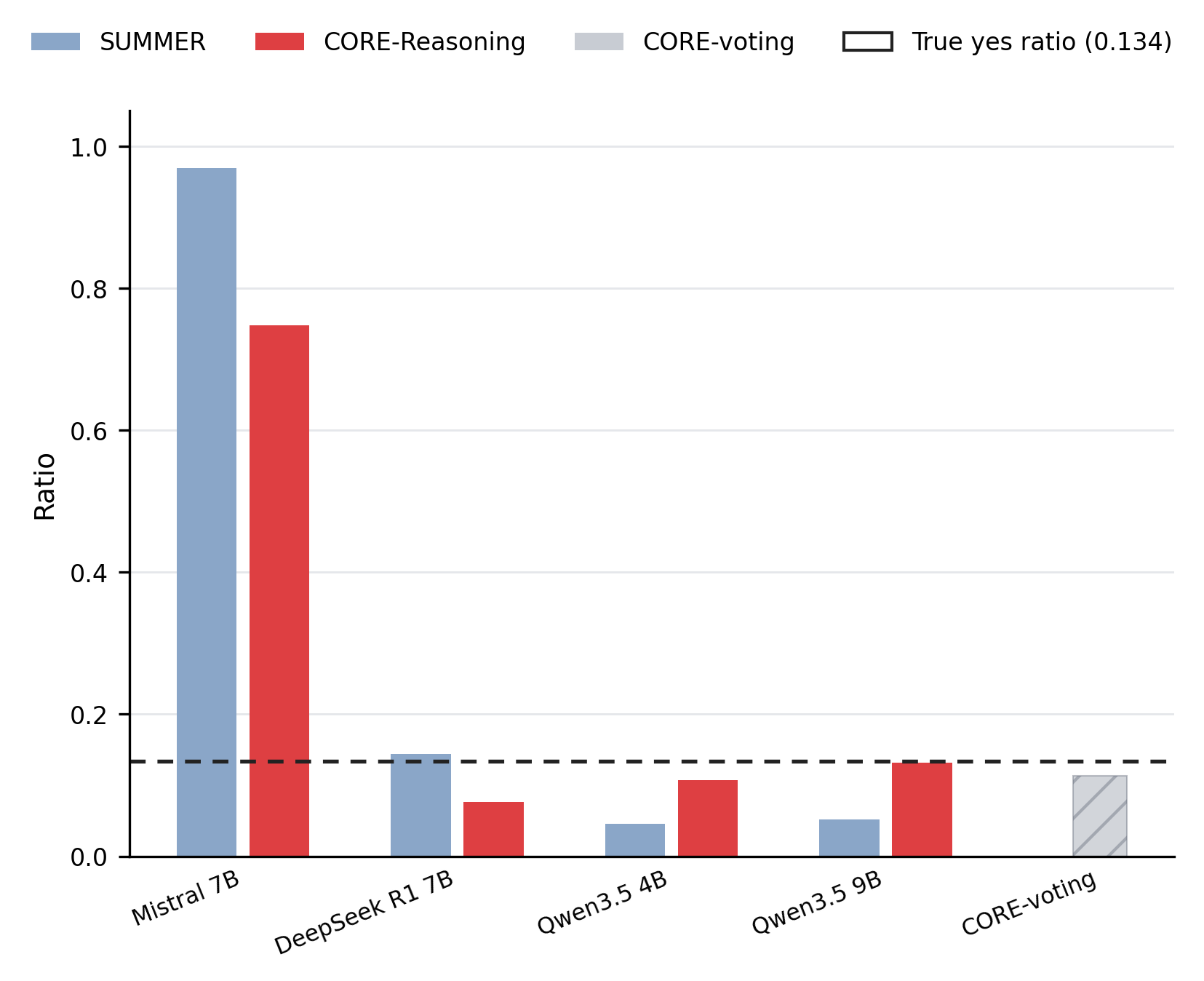}
    \caption{Predicted-positive rate.}
\end{subfigure}
\hfill
\begin{subfigure}[t]{0.45\linewidth}
    \centering
    \includegraphics[width=\linewidth]{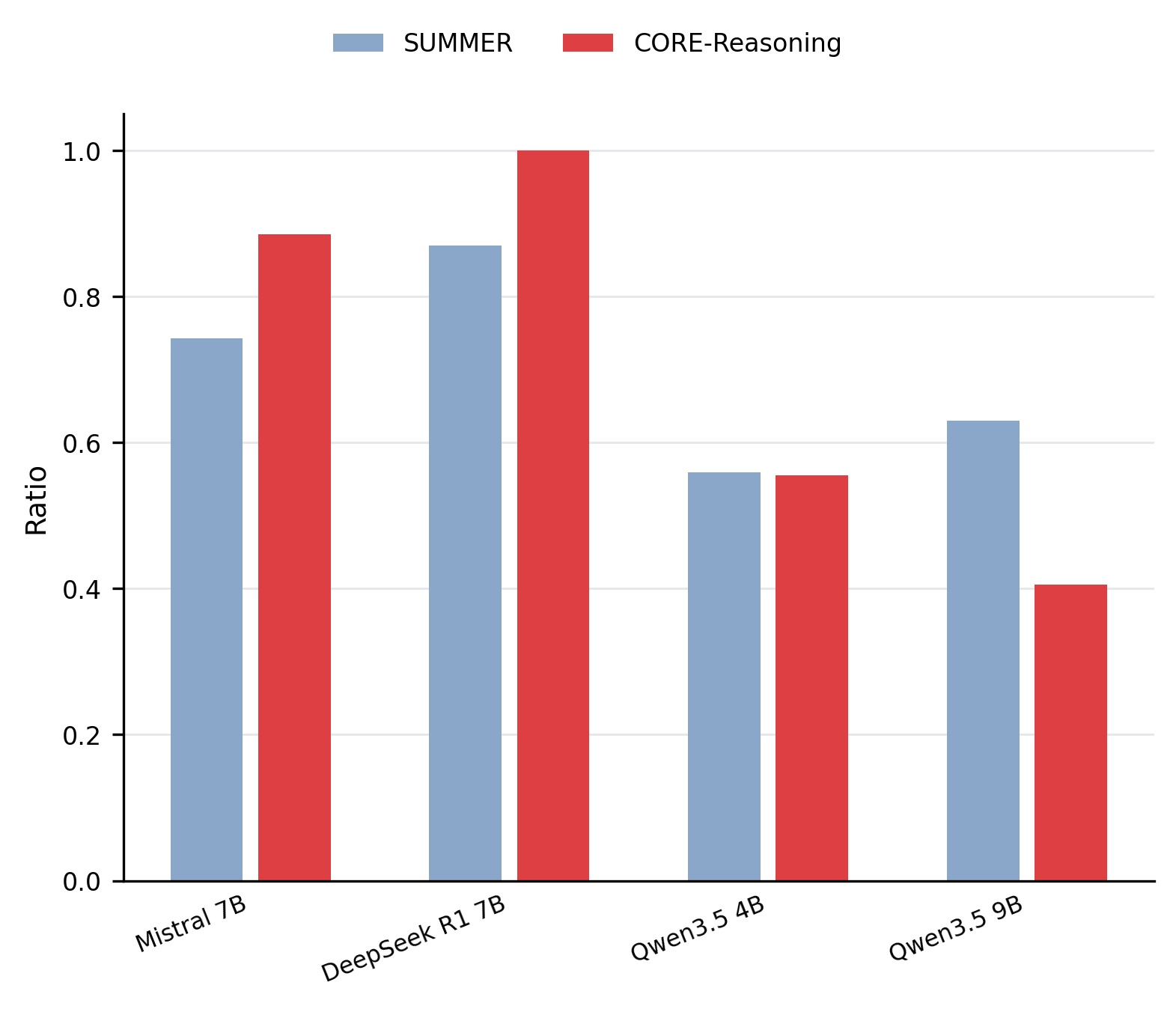}
    \caption{Effective answer rate.}
\end{subfigure}
\caption{
Calibration and output coverage for PerturbQA K562 cell line. CORE evidence changes
both the predicted-positive prevalence and the parseable answer rate, indicating
that the evidence packet affects not only final accuracy but also the model's
decision policy.
}
\label{fig:k562_llm_calibration}
\end{figure}

\textbf{Positive rate and effective answer rate for LLM variants.}  Fig.~\ref{fig:c32_llm_calibration} shows that on Tahoe100M C32, isolated VCWorld-style prompts often produce too many positive predictions relative to the empirical positive rate of 29.0\%. This is most visible for Mistral-7B, Llama-3.2-3B, and DeepSeek-R1-7B, whose predicted-positive rates are far above the dashed prevalence line. {\corereasoning} reduces this bias across most backbones, bringing the Qwen3.5 models and GPT-5.4 mini close to the empirical rate. Importantly, this shift is not simply caused by refusing more queries: the stronger backbones retain high answer rates, indicating that contrastive evidence improves calibration while preserving output coverage.

Fig.~\ref{fig:k562_llm_calibration} shows tha on PerturbQA K562, where the empirical positive rate is only 13.4\%, positive-rate calibration is harder. SUMMER-style prompting shows severe yes bias for Mistral-7B, while {\corereasoning} substantially lowers the predicted-yes rate, though it remains above the empirical prevalence. For DeepSeek-R1-7B and Qwen3.5 models, predicted-yes rates are closer to the target, but CORE evidence shifts predictions in different directions depending on the backbone. The answer-rate changes are also non-uniform: CORE improves coverage for most models. This suggests that CORE affects not only final labels, but also the model’s decision and parsing behavior.

\subsubsection{Full Numerical Results}
\label{app:full_numerical_results}

\begin{table*}[t]
\centering
\scriptsize
\setlength{\tabcolsep}{3pt}
\caption{Scaling results for C32 drug differential-expression prediction using {\geneprior}and {\corevoting}. Results are reported for each support size setting, with micro-averaged overall metrics and macro-averaged per-gene metrics. Bold values indicate the better method for each setting and metric. The final rows report averages over all $k,m$ settings. Higher is better for all metrics.}
\label{tab:c32-de-core-geneprior-scaling}
\begin{tabular}{rrlrrrrrrrr}
\toprule
$k$ & $m$ & Method & \multicolumn{4}{c}{Overall} & \multicolumn{4}{c}{Macro per gene} \\
\cmidrule(lr){4-7} \cmidrule(lr){8-11}
 & & & AUROC & AUPRC & Acc. & F1 & AUROC & AUPRC & Acc. & F1 \\
\midrule
1 & 40 & {\geneprior}& \textbf{0.545} & 0.346 & \textbf{0.732} & \textbf{0.142} & \textbf{0.500} & 0.652 & \textbf{0.382} & \textbf{0.035} \\
 & & {\corevoting} & 0.543 & \textbf{0.359} & \textbf{0.732} & \textbf{0.142} & 0.483 & \textbf{0.653} & \textbf{0.382} & \textbf{0.035} \\
1 & 200 & {\geneprior}& \textbf{0.708} & 0.558 & \textbf{0.816} & \textbf{0.542} & \textbf{0.500} & 0.652 & \textbf{0.512} & \textbf{0.174} \\
 & & {\corevoting} & 0.706 & \textbf{0.578} & \textbf{0.816} & \textbf{0.542} & 0.483 & \textbf{0.653} & \textbf{0.512} & \textbf{0.174} \\
1 & 1000 & {\geneprior}& 0.767 & 0.640 & \textbf{0.850} & \textbf{0.658} & 0.500 & 0.652 & \textbf{0.576} & \textbf{0.243} \\
 & & {\corevoting} & \textbf{0.769} & \textbf{0.657} & \textbf{0.850} & \textbf{0.658} & \textbf{0.505} & \textbf{0.656} & \textbf{0.576} & \textbf{0.243} \\
2 & 40 & {\geneprior}& \textbf{0.583} & 0.396 & \textbf{0.752} & \textbf{0.257} & \textbf{0.500} & 0.652 & \textbf{0.412} & \textbf{0.069} \\
 & & {\corevoting} & 0.581 & \textbf{0.413} & \textbf{0.752} & \textbf{0.257} & 0.483 & \textbf{0.653} & \textbf{0.412} & \textbf{0.069} \\
2 & 200 & {\geneprior}& 0.822 & 0.720 & \textbf{0.883} & \textbf{0.754} & 0.500 & 0.652 & \textbf{0.644} & \textbf{0.317} \\
 & & {\corevoting} & \textbf{0.824} & \textbf{0.728} & \textbf{0.883} & \textbf{0.754} & \textbf{0.505} & \textbf{0.655} & \textbf{0.644} & \textbf{0.317} \\
2 & 1000 & {\geneprior}& 0.875 & 0.795 & \textbf{0.914} & \textbf{0.831} & 0.500 & 0.652 & \textbf{0.706} & \textbf{0.382} \\
 & & {\corevoting} & \textbf{0.877} & \textbf{0.807} & \textbf{0.914} & \textbf{0.831} & \textbf{0.500} & \textbf{0.657} & \textbf{0.706} & \textbf{0.382} \\
5 & 40 & {\geneprior}& 0.700 & 0.553 & \textbf{0.816} & \textbf{0.544} & \textbf{0.500} & 0.652 & \textbf{0.528} & \textbf{0.197} \\
 & & {\corevoting} & \textbf{0.700} & \textbf{0.566} & \textbf{0.816} & \textbf{0.544} & 0.498 & \textbf{0.653} & \textbf{0.528} & \textbf{0.197} \\
5 & 200 & {\geneprior}& \textbf{0.926} & \textbf{0.872} & \textbf{0.947} & \textbf{0.902} & \textbf{0.500} & 0.652 & \textbf{0.788} & \textbf{0.479} \\
 & & {\corevoting} & 0.922 & 0.871 & 0.944 & 0.897 & 0.493 & \textbf{0.652} & 0.781 & 0.470 \\
5 & 1000 & {\geneprior}& \textbf{0.956} & \textbf{0.915} & \textbf{0.966} & \textbf{0.939} & \textbf{0.500} & 0.652 & \textbf{0.838} & \textbf{0.538} \\
 & & {\corevoting} & 0.947 & 0.911 & 0.961 & 0.931 & 0.497 & \textbf{0.653} & 0.824 & 0.521 \\
20 & 40 & {\geneprior}& \textbf{0.907} & 0.835 & \textbf{0.932} & \textbf{0.873} & 0.500 & 0.652 & \textbf{0.775} & \textbf{0.492} \\
 & & {\corevoting} & 0.906 & \textbf{0.846} & 0.929 & 0.865 & \textbf{0.502} & \textbf{0.657} & 0.768 & 0.481 \\
20 & 200 & {\geneprior}& 0.984 & 0.952 & \textbf{0.978} & \textbf{0.963} & 0.500 & 0.652 & \textbf{0.885} & \textbf{0.628} \\
 & & {\corevoting} & \textbf{0.985} & \textbf{0.963} & 0.975 & 0.958 & \textbf{0.537} & \textbf{0.675} & 0.874 & 0.607 \\
20 & 1000 & {\geneprior}& 0.987 & 0.959 & \textbf{0.978} & \textbf{0.963} & 0.500 & 0.652 & \textbf{0.884} & \textbf{0.630} \\
 & & {\corevoting} & \textbf{0.989} & \textbf{0.970} & 0.976 & 0.960 & \textbf{0.553} & \textbf{0.679} & 0.876 & 0.613 \\
50 & 40 & {\geneprior}& \textbf{0.973} & 0.933 & \textbf{0.971} & \textbf{0.951} & 0.500 & 0.652 & \textbf{0.847} & \textbf{0.598} \\
 & & {\corevoting} & 0.969 & \textbf{0.937} & 0.969 & 0.946 & \textbf{0.523} & \textbf{0.680} & 0.841 & 0.578 \\
50 & 200 & {\geneprior}& \textbf{0.990} & 0.965 & 0.978 & 0.963 & 0.500 & 0.652 & 0.885 & \textbf{0.619} \\
 & & {\corevoting} & 0.989 & \textbf{0.970} & \textbf{0.978} & \textbf{0.963} & \textbf{0.574} & \textbf{0.713} & \textbf{0.886} & 0.614 \\
50 & 1000 & {\geneprior}& \textbf{0.992} & 0.971 & \textbf{0.979} & \textbf{0.965} & 0.500 & 0.652 & \textbf{0.892} & \textbf{0.623} \\
 & & {\corevoting} & 0.991 & \textbf{0.972} & 0.979 & 0.964 & \textbf{0.623} & \textbf{0.739} & 0.888 & 0.615 \\
114 & 40 & {\geneprior}& \textbf{0.983} & \textbf{0.951} & \textbf{0.978} & \textbf{0.962} & 0.500 & 0.652 & \textbf{0.878} & \textbf{0.621} \\
 & & {\corevoting} & 0.981 & 0.950 & 0.972 & 0.952 & \textbf{0.537} & \textbf{0.678} & 0.858 & 0.579 \\
114 & 200 & {\geneprior}& \textbf{0.993} & \textbf{0.976} & \textbf{0.979} & \textbf{0.964} & 0.500 & 0.652 & \textbf{0.888} & \textbf{0.618} \\
 & & {\corevoting} & 0.993 & 0.976 & 0.978 & 0.963 & \textbf{0.601} & \textbf{0.707} & 0.885 & 0.605 \\
114 & 1000 & {\geneprior}& \textbf{0.994} & \textbf{0.978} & 0.979 & 0.964 & 0.500 & 0.652 & 0.889 & 0.621 \\
 & & {\corevoting} & 0.993 & 0.978 & \textbf{0.979} & \textbf{0.965} & \textbf{0.627} & \textbf{0.724} & \textbf{0.892} & \textbf{0.622} \\
\midrule
\multicolumn{2}{c}{Full-Support} & {\geneprior}& \textbf{0.994} & \textbf{0.978} & 0.979 & 0.964 & 0.500 & 0.652 & 0.889 & 0.621 \\
\multicolumn{2}{c}{} & {\corevoting} & 0.993 & 0.978 & \textbf{0.979} & \textbf{0.965} & \textbf{0.627} & \textbf{0.724} & \textbf{0.892} & \textbf{0.622} \\
\bottomrule
\end{tabular}
\end{table*}

\begin{table}[t]
\centering
\small
\caption{PerturbQA K562 differential-expression prediction results using {\geneprior} and {\corevoting}. The table reports micro-averaged overall metrics and macro-averaged per-gene metrics. Bold values indicate the better method for each setting and metric. Higher is better for all metrics. }
\label{tab:k562-de-core-geneprior}
\begin{tabular}{lrrrrrrrr}
\toprule
Method & \multicolumn{4}{c}{Overall} & \multicolumn{4}{c}{Macro per gene} \\
\cmidrule(lr){2-5} \cmidrule(lr){6-9}
 & AUROC & AUPRC & Acc. & F1 & AUROC & AUPRC & Acc. & F1 \\
\midrule
{\geneprior} & 0.954 & 0.813 & 0.935 & 0.734 & 0.500 & 0.403 & 0.693 & 0.181 \\
{\corevoting} & \textbf{0.956} & \textbf{0.832} & \textbf{0.938} & \textbf{0.750} & \textbf{0.711} & \textbf{0.712} & \textbf{0.705} & \textbf{0.197} \\
\bottomrule
\end{tabular}
\end{table}

We report the full numerical results for all performance metrics reported in Sec.~\ref{sec:experiment}.

\textbf{Comparison between {\geneprior} and {\corevoting}.}  Tab.~\ref{tab:c32-de-core-geneprior-scaling} reports the numerical results corresponding to Fig.~\ref{fig:c32_core_voting_sweep}, which shows that both voting methods scale strongly with larger support budgets on Tahoe100M C32, with overall AUROC approaching 0.99 under high-support settings. {\corevoting} most consistently improves ranking-oriented metrics, especially AUPRC and macro per-gene AUROC/AUPRC, while {\geneprior} often matches or slightly exceeds it on thresholded accuracy and F1. Under full support, the two methods are nearly tied overall, but {\corevoting} improves macro per-gene AUROC from 0.500 to 0.627 and macro AUPRC from 0.652 to 0.724, indicating better gene-level discrimination beyond the gene prior. The same pattern is stronger on K562 PerturbQA (Tab.~\ref{tab:k562-de-core-geneprior}), where {\corevoting} improves every reported metric, including overall AUPRC from 0.813 to 0.832 and macro per-gene AUROC from 0.500 to 0.711.

\begin{table*}[t]
\centering
\scriptsize
\caption{Tahoe100M C32 differential-expression prediction results for VCWorld prompting, {\corereasoning}, and {\corevoting}. All rows use the same $k = 50, m = 200$ support setting and evaluation set. Answer and predicted-yes rates are reported as percentages. Overall metrics are micro-averaged, and per-gene metrics are macro-averaged. Bold indicates the best value in each metric column, while the second best value is underlined.}
\label{tab:c32-llm-core-results}
\resizebox{\textwidth}{!}{%
\begin{tabular}{llrrrrrrrrrr}
\toprule
Model & Method & Ans. & Pred. yes & \multicolumn{4}{c}{Overall} & \multicolumn{4}{c}{Macro per gene} \\
\cmidrule(lr){5-8} \cmidrule(lr){9-12}
 & & (\%) & (\%) & AUROC & AUPRC & Acc. & F1 & AUROC & AUPRC & Acc. & F1 \\
\midrule
Mistral 7B & VCWorld & 75.1 & 92.1 & 0.365 & 0.699 & 0.239 & 0.331 & 0.389 & 0.503 & 0.480 & 0.516 \\
 & {\corereasoning} & 87.2 & 26.1 & 0.862 & 0.855 & 0.794 & 0.739 & 0.473 & 0.620 & 0.767 & 0.567 \\
Llama 3.2 3B & VCWorld & 78.4 & 80.2 & 0.391 & 0.678 & 0.297 & 0.343 & 0.395 & 0.536 & 0.482 & 0.520 \\
 & {\corereasoning} & 67.0 & 41.3 & 0.616 & 0.593 & 0.556 & 0.528 & 0.355 & 0.487 & 0.541 & 0.484 \\
DeepSeek R1 7B & VCWorld & 88.9 & 54.2 & 0.466 & 0.657 & 0.447 & 0.361 & 0.465 & 0.604 & 0.508 & 0.477 \\
 & {\corereasoning} & 79.3 & 34.6 & 0.775 & 0.761 & 0.726 & 0.696 & 0.408 & 0.570 & 0.678 & 0.521 \\
Qwen3.5 4B & VCWorld & 82.8 & 29.3 & 0.690 & 0.790 & 0.715 & 0.641 & 0.443 & 0.566 & 0.599 & 0.501 \\
 & {\corereasoning} & 97.0 & 30.0 & 0.950 & 0.925 & 0.941 & 0.920 & 0.548 & 0.705 & 0.852 & \textbf{0.625} \\
Qwen3.5 9B & VCWorld & 86.5 & 26.8 & 0.742 & 0.840 & 0.773 & 0.703 & 0.465 & 0.592 & 0.638 & 0.529 \\
 & {\corereasoning} & 95.9 & 30.4 & 0.935 & 0.890 & 0.926 & 0.904 & 0.545 & 0.687 & 0.842 & \underline{0.621} \\
GPT-5.4 mini & VCWorld & 89.3 & 17.3 & 0.666 & 0.864 & 0.746 & 0.586 & 0.492 & 0.631 & 0.554 & 0.455 \\
 & {\corereasoning} & 99.0 & 29.6 & \underline{0.980} & \underline{0.964} & \underline{0.967} & \underline{0.951} & \underline{0.556} & \textbf{0.731} & \underline{0.874} & 0.610 \\
/ & {\corevoting} & 100.0 & 29.5 & \textbf{0.989} & \textbf{0.970} & \textbf{0.978} & \textbf{0.963} & \textbf{0.574} & \underline{0.713} & \textbf{0.886} & 0.614 \\
\bottomrule
\end{tabular}%
}
\end{table*}

\begin{table*}[t]
\centering
\scriptsize
\caption{Overall and macro-per-gene performance on PerturbQA K562. 
For each model, better values between SUMMER and {\corereasoning} are shown in bold.}
\label{tab:perturbqa_k562_full}
\begin{tabular}{llccccccccc}
\toprule
\multirow{2}{*}{Model} & \multirow{2}{*}{Method} & \multirow{2}{*}{Ans.\%}
& \multicolumn{4}{c}{Overall}
& \multicolumn{4}{c}{Macro-per-gene} \\
\cmidrule(lr){4-7} \cmidrule(lr){8-11}
& & & Acc. & F1 & AUROC & AUPRC & Acc. & F1 & AUROC & AUPRC \\
\midrule
Mistral 7B & SUMMER & 74.2 & 0.122 & 0.182 & 0.373 & 0.103 & 0.075 & 0.068 & 0.397 & 0.373 \\
 & {\corereasoning} & \textbf{88.5} & \textbf{0.320} & \textbf{0.246} & \textbf{0.608} & \textbf{0.186} & \textbf{0.278} & \textbf{0.073} & \textbf{0.482} & \textbf{0.528} \\

Llama 3.2 3B & SUMMER & \textbf{96.1} & 0.583 & 0.211 & 0.504 & 0.139 & 0.596 & 0.055 & 0.491 & 0.485 \\
 & {\corereasoning} & 92.4 & \textbf{0.711} & \textbf{0.405} & \textbf{0.740} & \textbf{0.357} & \textbf{0.647} & \textbf{0.071} & \textbf{0.492} & \textbf{0.540} \\

DeepSeek R1 7B & SUMMER & 87.0 & 0.669 & 0.156 & 0.455 & 0.129 & 0.709 & 0.027 & 0.459 & 0.448 \\
 & {\corereasoning} & \textbf{100.0} & \textbf{0.902} & \textbf{0.537} & \textbf{0.911} & \textbf{0.683} & \textbf{0.937} & \textbf{0.041} & \textbf{0.546} & \textbf{0.590} \\


Qwen3.5 4B & SUMMER & 88.9 & 0.741 & 0.160 & 0.480 & 0.137 & 0.791 & 0.017 & 0.455 & 0.409 \\
  & {\corereasoning} & \textbf{97.7} & \textbf{0.916} & \textbf{0.727} & \textbf{0.845} & \textbf{0.678} & \textbf{0.931} & \textbf{0.047} & \textbf{0.582} & \textbf{0.607} \\

Qwen3.5 9B & SUMMER & 94.6 & 0.795 & 0.168 & 0.511 & 0.147 & 0.846 & 0.017 & 0.485 & 0.426 \\
 & {\corereasoning} & \textbf{96.3} & \textbf{0.900} & \textbf{0.717} & \textbf{0.816} & \textbf{0.615} & \textbf{0.915} & \textbf{0.064} & \textbf{0.607} & \textbf{0.623} \\


/ & {\geneprior} & 100.0 & 0.935 & 0.734 & 0.954 & 0.813 & 0.693 & 0.181 & 0.500 & 0.403 \\
/ & {\corevoting} & 100.0 & 0.938 & 0.750 & 0.956 & 0.832 & 0.705 & 0.197 & 0.711 & 0.712 \\
\bottomrule
\end{tabular}
\end{table*}

\textbf{Comparison between {\corereasoning} and {\corevoting} on PerturbQA across all four cell lines.}
On K562 cell line (Tab.~\ref{tab:perturbqa_k562_full}), {\corereasoning} with DeepSeek-R1-7B gives the strongest ranking metrics among LLMs, reaching near-complete answer coverage with 0.902 overall accuracy and 0.911 AUROC. This is substantially higher than the corresponding SUMMER baseline, which obtains 0.669 accuracy and 0.455 AUROC.


\begin{table*}[t]
\centering
\small
\caption{Overall and macro-per-gene performance on HepG2. For each model, the better value between SUMMER and {\corereasoning} is shown in bold.}
\label{tab:core_results_hepg2}
\resizebox{\textwidth}{!}{%
\begin{tabular}{llccccccccc}
\toprule
Model & Method & Ans.\%
& \multicolumn{4}{c}{Overall}
& \multicolumn{4}{c}{Macro-per-gene} \\
\cmidrule(lr){4-7} \cmidrule(lr){8-11}
& & & Acc. & F1 & AUROC & AUPRC & Acc. & F1 & AUROC & AUPRC \\
\midrule
Mistral 7B & SUMMER
& 72.6 & 0.122 & 0.182 & 0.366 & 0.106 & 0.068 & 0.054 & 0.405 & 0.458 \\
 & {\corereasoning}
& \textbf{87.4} & \textbf{0.359} & \textbf{0.271} & \textbf{0.609} & \textbf{0.192}
& \textbf{0.355} & \textbf{0.059} & \textbf{0.486} & \textbf{0.615} \\

Llama 3.2 3B & SUMMER
& \textbf{96.2} & 0.591 & 0.213 & 0.501 & 0.144 & 0.615 & 0.037 & 0.495 & 0.582 \\
 & {\corereasoning}
& 93.1 & \textbf{0.759} & \textbf{0.484} & \textbf{0.786} & \textbf{0.482}
& \textbf{0.692} & \textbf{0.055} & \textbf{0.526} & \textbf{0.653} \\

DeepSeek R1 7B & SUMMER
& 86.1 & 0.654 & 0.147 & 0.444 & 0.130 & 0.721 & 0.017 & 0.444 & 0.529 \\
 & {\corereasoning}
& \textbf{100.0} & \textbf{0.913} & \textbf{0.621} & \textbf{0.924} & \textbf{0.760}
& \textbf{0.950} & \textbf{0.032} & \textbf{0.524} & \textbf{0.672} \\



Qwen3.5 4B & SUMMER &  84.1 & 0.692 & 0.160 & 0.455 & 0.134 & 0.770 & 0.012 & 0.440 & 0.462 \\

 & {\corereasoning}   & \textbf{97.6} & \textbf{0.920} & \textbf{0.765} & \textbf{0.874} & \textbf{0.711} & \textbf{0.937} & \textbf{0.036} & \textbf{0.593} & \textbf{0.676} \\

Qwen3.5 9B & SUMMER & 90.8 & 0.756 & 0.186 & 0.499 & 0.152 & 0.828 & 0.014 & 0.473 & 0.493 \\
 & {\corereasoning}   & \textbf{96.6} & \textbf{0.908} & \textbf{0.747} & \textbf{0.840} & \textbf{0.661} & \textbf{0.929} & \textbf{0.040} & \textbf{0.603} & \textbf{0.690} \\

/ & {\geneprior}
& 100.0 & 0.943 & 0.799 & 0.957 & 0.844 & 0.652 & 0.312 & 0.500 & 0.482 \\
/ & {\corevoting}
& 100.0 & 0.945 & 0.799 & 0.958 & 0.866 & 0.664 & 0.291 & 0.684 & 0.744 \\
\bottomrule
\end{tabular}%
}
\end{table*}

On HepG2 cell line (Tab.~\ref{tab:core_results_hepg2}), {\corereasoning} again provides a strong LLM-only baseline: the DeepSeek-R1-7B run achieves 0.913 overall accuracy, 0.924 AUROC, and 0.760 AUPRC at nearly complete coverage. Compared with SUMMER, {\corereasoning} improves accuracy from 0.654 to 0.913 and AUROC from 0.444 to 0.924.


\begin{table*}[t]
\centering
\small
\caption{Overall and macro-per-gene performance on Jurkat. For each model, the better value between SUMMER and {\corereasoning} is shown in bold.}
\label{tab:core_results_jurkat}
\resizebox{\textwidth}{!}{%
\begin{tabular}{llccccccccc}
\toprule
Model & Method & Ans.\%
& \multicolumn{4}{c}{Overall}
& \multicolumn{4}{c}{Macro-per-gene} \\
\cmidrule(lr){4-7} \cmidrule(lr){8-11}
& & & Acc. & F1 & AUROC & AUPRC & Acc. & F1 & AUROC & AUPRC \\
\midrule
Mistral 7B & SUMMER
& 73.6 & 0.126 & 0.189 & 0.367 & 0.109 & 0.064 & 0.050 & 0.399 & 0.415 \\
 & {\corereasoning}
& \textbf{87.8} & \textbf{0.345} & \textbf{0.266} & \textbf{0.621} & \textbf{0.201}
& \textbf{0.315} & \textbf{0.053} & \textbf{0.458} & \textbf{0.549} \\

Llama 3.2 3B & SUMMER
& \textbf{96.3} & 0.598 & 0.231 & 0.518 & 0.152 & 0.617 & 0.039 & 0.499 & 0.525 \\
 & {\corereasoning}
& 93.2 & \textbf{0.755} & \textbf{0.474} & \textbf{0.773} & \textbf{0.480}
& \textbf{0.688} & \textbf{0.051} & \textbf{0.503} & \textbf{0.580} \\

DeepSeek R1 7B & SUMMER
& 86.8 & 0.664 & 0.186 & 0.468 & 0.145 & 0.721 & 0.020 & 0.459 & 0.478 \\
 & {\corereasoning}
& \textbf{100.0} & \textbf{0.914} & \textbf{0.623} & \textbf{0.926} & \textbf{0.772}
& \textbf{0.953} & \textbf{0.029} & \textbf{0.556} & \textbf{0.631} \\



Qwen3.5 4B & SUMMER  & 83.4 & 0.691 & 0.210 & 0.473 & 0.151 & 0.757 & 0.014 & 0.445 & 0.422 \\
 & {\corereasoning}    & \textbf{97.9} & \textbf{0.927} & \textbf{0.779} & \textbf{0.872} & \textbf{0.723} & \textbf{0.942} & \textbf{0.032} & \textbf{0.593} & \textbf{0.647} \\
Qwen3.5 9B & SUMMER  & 90.9 & 0.762 & 0.232 & 0.518 & 0.168 & 0.827 & 0.014 & 0.474 & 0.444 \\
 & {\corereasoning}    & \textbf{96.5} & \textbf{0.907} & \textbf{0.746} & \textbf{0.827} & \textbf{0.654} & \textbf{0.927} & \textbf{0.036} & \textbf{0.573} & \textbf{0.634} \\

/ & {\geneprior}
& 100.0 & 0.949 & 0.814 & 0.960 & 0.870 & 0.681 & 0.225 & 0.500 & 0.437 \\
/ & {\corevoting}
& 100.0 & 0.951 & 0.816 & 0.960 & 0.886 & 0.698 & 0.216 & 0.686 & 0.715 \\
\bottomrule
\end{tabular}%
}
\end{table*}

On Jurkat cell line (Tab.~\ref{tab:core_results_jurkat}), {\corereasoning} with DeepSeek-R1-7B is the top LLM-based method, achieving 0.914 overall accuracy and 0.926 AUROC, compared with 0.664 accuracy and 0.468 AUROC for SUMMER. {\corevoting} is strongest on overall AUROC/AUPRC and macro-per-gene AUROC/AUPRC, while thresholded macro F1/Acc remain mixed.

\begin{table}[t]
\centering
\scriptsize
\caption{Overall and macro-per-gene performance on RPE1. For each model, the better value between SUMMER and {\corereasoning} is shown in bold.}
\label{tab:core_results_rpe1}
\begin{tabular}{llccccccccc}
\toprule
Model & Method & Ans.\% 
& \multicolumn{4}{c}{Overall} 
& \multicolumn{4}{c}{Macro-per-gene} \\
\cmidrule(lr){4-7} \cmidrule(lr){8-11}
& & & Acc. & F1 & AUROC & AUPRC & Acc. & F1 & AUROC & AUPRC \\
\midrule
Mistral 7B & SUMMER 
& 73.9 & 0.124 & 0.188 & 0.372 & 0.107 & 0.066 & 0.057 & 0.391 & 0.346 \\
 & {\corereasoning} 
& \textbf{88.3} & \textbf{0.325} & \textbf{0.262} & \textbf{0.628} & \textbf{0.205} & \textbf{0.271} & \textbf{0.064} & \textbf{0.488} & \textbf{0.500} \\

Llama 3.2 3B & SUMMER 
& \textbf{96.4} & 0.587 & 0.204 & 0.491 & 0.140 & 0.605 & 0.044 & 0.482 & 0.451 \\
 & {\corereasoning} 
& 92.9 & \textbf{0.745} & \textbf{0.467} & \textbf{0.775} & \textbf{0.464} & \textbf{0.679} & \textbf{0.061} & \textbf{0.490} & \textbf{0.514} \\

DeepSeek R1 7B & SUMMER 
& 86.9 & 0.659 & 0.146 & 0.446 & 0.129 & 0.707 & 0.022 & 0.448 & 0.421 \\
 & {\corereasoning} 
& \textbf{100.0} & \textbf{0.915} & \textbf{0.633} & \textbf{0.942} & \textbf{0.794} & \textbf{0.952} & \textbf{0.040} & \textbf{0.544} & \textbf{0.556} \\



Qwen3.5 4B  & SUMMER & 86.9 & 0.716 & 0.162 & 0.469 & 0.140 & 0.773 & 0.017 & 0.450 & 0.397 \\
   & {\corereasoning}   & \textbf{98.1} & \textbf{0.933} & \textbf{0.801} & \textbf{0.890} & \textbf{0.778} & \textbf{0.945} & \textbf{0.043} & \textbf{0.627} & \textbf{0.622} \\
Qwen3.5 9B  & SUMMER & 92.7 & 0.770 & 0.166 & 0.499 & 0.149 & 0.830 & 0.015 & 0.474 & 0.405 \\
  & {\corereasoning}   & \textbf{96.3} & \textbf{0.910} & \textbf{0.763} & \textbf{0.838} & \textbf{0.678} & \textbf{0.922} & \textbf{0.049} & \textbf{0.590} & \textbf{0.595} \\

/ & {\geneprior} 
& 100.0 & 0.955 & 0.837 & 0.975 & 0.902 & 0.741 & 0.245 & 0.500 & 0.396 \\
/ & {\corevoting} 
& 100.0 & 0.956 & 0.843 & 0.976 & 0.911 & 0.750 & 0.250 & 0.729 & 0.703 \\
\bottomrule
\end{tabular}
\end{table}

On RPE1 cell line (Tab.~\ref{tab:core_results_rpe1}), {\corereasoning} with DeepSeek-R1-7B reaches 0.915 Overall accuracy, 0.942 AUROC, and 0.794 AUPRC with near-complete answer coverage, clearly outperforming SUMMER on the same backbone. {\corevoting} gives the strongest final predictions, and improves not only aggregate performance but also per-gene discriminative consistency.

Across the four cell lines, {\corereasoning} consistently produces the strongest LLM-based results, especially with DeepSeek-R1-7B, and it substantially improves over the corresponding SUMMER runs under the same model backbone. {\corevoting} is the strongest final prediction strategy: it gives the best or tied-best overall AUROC and AUPRC in every cell line and best macro-per-gene AUROC and AUPRC across all four cell lines; macro F1 is more threshold-sensitive and not uniformly best.

Together, these results show a complementary advantage: {\corevoting} provides robust deterministic scores from the same contrastive evidence, while {\corereasoning} provides interpretable LLM-based adjudication and generally improves prediction quality.

\subsection{Case Study}
\label{app:case-study}

We select qualitative case studies by first parsing the Qwen3.5-9B outputs and
joining them to the corresponding CORE evidence packets. The cases in
Tab.~\ref{tab:case_study_examples} are held-out C32 \(k=50,m=200\) examples
chosen to cover the main evidence regimes observed in the run. The VCWorld
column uses the same model checkpoint with the original prompt, whereas the
{\corereasoning} column reports the parsed label and probability from the
structured CORE output. Across the full held-out set, {\corereasoning} changes
\(5{,}451\) answered VCWorld errors into correct predictions, including
\(1{,}573\) false-positive corrections and \(3{,}878\) false-negative rescues;
it also turns \(1{,}999\) VCWorld abstentions on positive examples into correct
positive predictions. The examples below therefore illustrate recurring
correction modes rather than isolated anecdotes.

For each case, we report only evidence visible to the model: same-gene positive
and negative vote masses, top support perturbations, gene-anchor scores, and
displayed KG paths or endpoint mappings. This allows the qualitative analysis to
separate empirical support transfer from generic biomedical plausibility.

\begin{table}[h]
\centering
\scriptsize
\caption{Representative C32 case studies selected from Qwen3.5-9B predictions.
They cover false-positive correction, mixed-evidence positive rescue,
direct KG-path rescue, sparse anchor-driven behavior, and a weak-mapping
failure.}
\label{tab:case_study_examples}
\begin{adjustbox}{max width=\textwidth}
\begin{tabular}{p{0.15\textwidth}p{0.19\textwidth}p{0.07\textwidth}p{0.10\textwidth}p{0.12\textwidth}p{0.31\textwidth}}
\toprule
Regime & Query pair & True \(y\) & VCWorld & {\corereasoning} & Evidence snapshot \\
\midrule
False-positive correction
& Gemfibrozil--GADD45B
& 0
& yes
& no, \(p=0.095\)
& Negative same-gene evidence dominates the packet: vote mass \(3.346\) versus
\(0.382\) positive, with a low anchor score \(0.088\). The top negative supports
are Ritonavir, Clopidogrel, and Methylprednisolone succinate, while the only
displayed positive support is Oleic acid. Although the query endpoints are
directly covered and short paths connect GADD45B to Gemfibrozil through
glucocorticoid intermediates, {\corereasoning} assigns more weight to the
contrastive negative supports than to generic path plausibility. \\

Mixed-evidence positive rescue
& Mebendazole--KIF14
& 1
& no
& yes, \(p=0.550\)
& The evidence is deliberately not one-sided: positive vote mass is \(2.141\)
and negative vote mass is \(1.412\). Positives include Docetaxel, Artemether,
and Methylthiouracil; negatives include Methylprednisolone succinate, Ritonavir,
and Talc. The model makes only a modest positive update, consistent with the
displayed direct-endpoint path KIF14 -- AARS1 -- Mebendazole and the mixed
same-gene contrast. \\

Direct KG-path rescue
& Mitoxantrone (dihydrochloride)--GADD45A
& 1
& no
& yes, \(p=1.000\)
& Same-gene support is nearly balanced (\(0.996\) positive versus \(1.064\)
negative), so the retrieved support neighbors alone would not strongly favor a
positive label. The displayed KG evidence, however, contains a direct one-hop
query path, GADD45A -- Mitoxantrone, with relation
\texttt{INV::upregulates}. {\corereasoning} treats this endpoint-specific path as
decisive evidence for a positive prediction. \\

Sparse-evidence anchor rescue
& Homoharringtonine--RMEL3
& 1
& no
& yes, \(p=0.918\)
& No same-gene neighbor vote mass is observed and no KG path is displayed. CORE
therefore uses the documented fallback score \(0.867\) together with a high
gene-anchor score \(0.968\). The anchors A2M, IGF2, and TEX41 all support a
positive response under the query perturbation, allowing the model to recover a
positive label despite sparse direct evidence. \\

Weak-mapping failure
& Elimusertib hydrochloride--LINC00824
& 0
& yes
& yes, \(p=1.000\)
& The query perturbation is not directly covered and is mapped by character
\(n\)-gram similarity to alternative drug endpoints, including Crizotinib
hydrochloride. The default blended score is only \(0.409\), with weak positive
and negative vote masses (\(0.890\) versus \(0.500\)), but the LLM over-commits
to a positive label. The displayed paths use mapped Crizotinib endpoints and
generic anatomy nodes, illustrating a residual failure mode when endpoint
grounding is weak. \\
\bottomrule
\end{tabular}
\end{adjustbox}
\end{table}

\textbf{False-positive correction.}
These examples are negative DE pairs for which a baseline prompt predicts a
positive label from plausible but nonspecific biomedical associations, while
{\corereasoning} predicts \texttt{no}. The analysis emphasizes whether the model
uses negative same-gene supports and the calibrated CORE score to avoid treating
any KG path as sufficient causal evidence.

\textbf{False-negative rescue.}
These examples are positive DE pairs for which a baseline prompt predicts
\texttt{no} or abstains, while {\corereasoning} predicts \texttt{yes}. We use
them to examine whether contrastive same-gene support, endpoint-specific KG
paths, and anchor evidence recover positives that are difficult to infer from
generic perturbation or gene descriptions alone.

\textbf{Strong KG path case.}
These examples contain direct grounded endpoints and compact KG paths between
the query perturbation and gene. We use them to evaluate whether the LLM can
distinguish endpoint-specific path evidence from broad graph connectivity,
especially when same-gene support is mixed.

\textbf{Sparse-evidence anchor case.}
These examples have limited same-gene vote mass or weak KG grounding, so the
fallback score and anchor-smoothed evidence dominate the final prediction. They
highlight the behavior of CORE under sparse query-specific evidence and make
clear when the prediction is not supported by direct KG paths.

\textbf{Mixed-evidence calibrated case.}
These examples have comparable positive and negative vote masses, or a
high-confidence same-gene score but weaker anchor or path evidence. We use them
to inspect whether {\corereasoning} preserves uncertainty rather than converting
all biologically plausible cases into high-confidence positives.

\textbf{Failure cases.}
Failure cases are held-out examples where {\corereasoning} is confidently wrong
or where weak name-based mapping evidence leads to a misleading prediction. We
include these cases to characterize the limits of KG grounding, support
transfer, and LLM calibration under distribution shift.

\section{Full Prompt and Model Output Examples}
\label{app:full-prompt-output-examples}

This appendix section gives one complete held-out C32 \(k=50,m=200\) example
for comparing VCWorld-style prompting and \corereasoning. The prompt boxes are
shown exactly as rendered for inference. For model outputs, internal
\texttt{<think>...</think>} scratchpad content is omitted, while the visible
reasoning and final answer are kept verbatim.

The example query is Gemfibrozil--GADD45B with true label \(y=0\).

\textbf{VCWorld Prompt}
\begin{framed}
\begin{Verbatim}[fontsize=\tiny,breaklines=true,breakanywhere=true]
[Start of Prompt]
You are VCWorld, a sophisticated Biological World Model and Causal Reasoning Engine. Your task is to simulate and predict the cellular response to drug perturbations.

Goal: Determine if a perturbation of Gemfibrozil in the C32 cells cell line results in the differential expression (DE) of GADD45B.

Input Data:
- Drug (Gemfibrozil): description of drug that is to perturb the cell
- Gene (GADD45B): description of gene, the impact on which you wish to infer
- Biological Context: description of cell line in which the genes are expressed
  *(Instruction: You MUST explicitly utilize the description above AND enhance it with your internal biological knowledge regarding C32 cells's tissue of origin, hallmark mutations (e.g., KRAS, TP53 status), and signaling idiosyncrasies.)*
- Evidence Set: set of experimental observations that describe the impact of small molecule perturbations on related genes, to contextualize your answer

Reasoning Guidelines:
Do not rely on superficial text matching. Perform a stepwise biological simulation as follows.

Output: Provide a structured analysis answering the following steps.

1) **Mechanism & Analogue Identification:**
   Identify drugs in the evidence set that share the same *Mechanism of Action (MoA)* or target specific pathway nodes as Gemfibrozil.

2) **Specificity & Relevance Analysis (Drug-Gene-Cell Triad):**
   Analyze the potential associations between the Drug, Gene, and Cell Line:
   - **Specificity:** Is the drug's effect broad (e.g., general stress) or specific (e.g., targeted kinase inhibition)? Is the gene's expression tissue-specific?
   - **Relevance:** Given the enhanced cell line context (e.g., its mutations), is the drug's target relevant in this specific cellular environment? (e.g., Does the cell rely on the targeted pathway?)

3) **Downstream Signaling Cascade Simulation:**
   Trace the signaling cascade initiated by Gemfibrozil. When Gemfibrozil inhibits/activates its target, which specific downstream kinases, transcription factors, or stress responses are modulated?
   *Constraint:* Ensure this simulation aligns with the enhanced context of C32 cells.

4) **Causal Bridge & Evidence Synthesis:**
   Connect the drug's downstream effect to the gene's regulatory requirements.
   - Construct a logical bridge: Drug -> Target -> Pathway -> TF -> Gene.
   - Cite specific "Analogue Cases" that support this link.
   - *Soft Reference:* Briefly refer to "Contrast Cases" (if available) as supplementary context to see if they offer a different perspective or boundary condition, but focus primarily on constructing the positive mechanism.

5) **Final Deterministic Prediction:**
   Based on the analysis above, determine if the drug effectively perturbs the gene in this specific cell line.

   End your response with exactly one of the following options:
   - No. Perturbation of Gemfibrozil does not impact GADD45B.
   - Yes. Perturbation of Gemfibrozil results in differential expression of GADD45B.
   - There is insufficient evidence to determine how Perturbation of Gemfibrozil affects GADD45B.
[End of Prompt]

[Start of Input]
- Description of molecule drug (Gemfibrozil): Gemfibrozil activates the nuclear receptor peroxisome proliferator-activated receptor-alpha (PPARalpha), which alters lipid metabolism. This activation increases lipoprotein lipase activity, inhibits apolipoprotein B synthesis, and increases the clearance of apolipoprotein B. This perturbation may affect the expression of genes containing a PPARalpha response element that are involved in fatty acid oxidation and lipoprotein metabolism.
- Description of gene of interest (GADD45B): is a nuclear and cytoplasmic protein that regulates the cell cycle and promotes apoptosis. It functions by interacting with cell cycle components like CDK1 and positively regulating stress-activated JNK and p38MAPK signaling cascades. Its expression and activity could be impacted by perturbations that induce cellular stress or DNA damage, which are known to activate upstream MAPK pathways.
- Context: C32 is a human amelanotic melanoma cell line derived from skin and widely used in cancer biology and drug discovery. It harbors hallmark alterations including BRAF V600E mutation, PTEN deletion, CDKN2A mutation, and a TERT promoter mutation, reflecting key oncogenic pathways in melanoma. With a doubling time of ~53 hours, microsatellite stability (MSS), and well-defined HLA typing, C32 serves as a valuable model for immunological studies and dependency analyses in large-scale omics resources such as CCLE and DepMap. Its predominantly European genetic ancestry further contextualizes its relevance for investigating melanoma pathogenesis, signaling dependencies, and targeted therapeutic strategies.
- Examples: Example 1:
- Drug: Methylprednisolone succinate
- Gene: GADD45B
- Drug Description: Drug 'Methylprednisolone succinate' description not found
- Gene Description: is a nuclear and cytoplasmic protein that regulates the cell cycle and promotes apoptosis. It functions by interacting with cell cycle components like CDK1 and positively regulating stress-activated JNK and p38MAPK signaling cascades. Its expression and activity could be impacted by perturbations that induce cellular stress or DNA damage, which are known to activate upstream MAPK pathways.
- Result: A) Perturbation of this drug does not impact the gene of interest.

Example 2:
- Drug: AZD1390
- Gene: GADD45B
- Drug Description: AZD1390 is a potent inhibitor of Ataxia-Telangiectasia Mutated (ATM) kinase. This action prevents the activation of the DNA damage checkpoint and disrupts DNA repair processes. Perturbation with AZD1390 may affect the expression of genes involved in the DNA damage response and cell cycle checkpoints regulated by ATM kinase.
- Gene Description: is a nuclear and cytoplasmic protein that regulates the cell cycle and promotes apoptosis. It functions by interacting with cell cycle components like CDK1 and positively regulating stress-activated JNK and p38MAPK signaling cascades. Its expression and activity could be impacted by perturbations that induce cellular stress or DNA damage, which are known to activate upstream MAPK pathways.
- Result: B) Perturbation of this drug results in differential expression of the gene of interest.

Example 3:
- Drug: Relugolix
- Gene: GADD45B
- Drug Description: Relugolix is a competitive antagonist of the gonadotropin-releasing hormone (GnRH) receptor. By blocking GnRH receptors in the pituitary gland, it decreases the release of luteinizing hormone and follicle-stimulating hormone, ultimately reducing testosterone production. This perturbation may affect the expression of androgen-responsive genes in hormone-sensitive tissues by lowering circulating androgen levels.
- Gene Description: is a nuclear and cytoplasmic protein that regulates the cell cycle and promotes apoptosis. It functions by interacting with cell cycle components like CDK1 and positively regulating stress-activated JNK and p38MAPK signaling cascades. Its expression and activity could be impacted by perturbations that induce cellular stress or DNA damage, which are known to activate upstream MAPK pathways.
- Result: A) Perturbation of this drug does not impact the gene of interest.

Example 4:
- Drug: Ritonavir
- Gene: GADD45B
- Drug Description: Ritonavir is an HIV protease inhibitor that prevents the cleavage of the viral gag-pol polyprotein, resulting in the formation of immature, non-infectious viral particles. In host cells, it functions as a potent inhibitor of the cytochrome P450 enzyme CYP3A4 and as an activator of the nuclear receptor NR1I2. As an activator of the nuclear receptor NR1I2, ritonavir perturbation may affect the expression of target genes involved in xenobiotic metabolism.
- Gene Description: is a nuclear and cytoplasmic protein that regulates the cell cycle and promotes apoptosis. It functions by interacting with cell cycle components like CDK1 and positively regulating stress-activated JNK and p38MAPK signaling cascades. Its expression and activity could be impacted by perturbations that induce cellular stress or DNA damage, which are known to activate upstream MAPK pathways.
- Result: A) Perturbation of this drug does not impact the gene of interest.

Example 5:
- Drug: Indacaterol (maleate)
- Gene: GADD45B
- Drug Description: Indacaterol is a beta-2 adrenergic receptor agonist that stimulates intracellular adenyl cyclase, leading to increased cyclic AMP levels. This causes the relaxation of bronchial smooth muscle. This perturbation may affect the expression of genes regulated by transcription factors downstream of the beta-2 adrenergic signaling pathway, such as CREB.
- Gene Description: is a nuclear and cytoplasmic protein that regulates the cell cycle and promotes apoptosis. It functions by interacting with cell cycle components like CDK1 and positively regulating stress-activated JNK and p38MAPK signaling cascades. Its expression and activity could be impacted by perturbations that induce cellular stress or DNA damage, which are known to activate upstream MAPK pathways.
- Result: A) Perturbation of this drug does not impact the gene of interest.

Example 6:
- Drug: Clopidogrel
- Gene: GADD45B
- Drug Description: Drug 'Clopidogrel' description not found
- Gene Description: is a nuclear and cytoplasmic protein that regulates the cell cycle and promotes apoptosis. It functions by interacting with cell cycle components like CDK1 and positively regulating stress-activated JNK and p38MAPK signaling cascades. Its expression and activity could be impacted by perturbations that induce cellular stress or DNA damage, which are known to activate upstream MAPK pathways.
- Result: A) Perturbation of this drug does not impact the gene of interest.

Example 7:
- Drug: Ouabain (Octahydrate)
- Gene: GADD45B
- Drug Description: Ouabain inhibits the Na-K-ATPase membrane pump, which leads to an increase in intracellular sodium and calcium concentrations. This perturbation may affect the expression of genes regulated by calcium-dependent signaling pathways involved in cellular stress and muscle contraction.
- Gene Description: is a nuclear and cytoplasmic protein that regulates the cell cycle and promotes apoptosis. It functions by interacting with cell cycle components like CDK1 and positively regulating stress-activated JNK and p38MAPK signaling cascades. Its expression and activity could be impacted by perturbations that induce cellular stress or DNA damage, which are known to activate upstream MAPK pathways.
- Result: B) Perturbation of this drug results in differential expression of the gene of interest.

Example 8:
- Drug: L-Thyroxine (sodium salt pentahydrate)
- Gene: GADD45B
- Drug Description: Drug 'L-Thyroxine (sodium salt pentahydrate)' description not found
- Gene Description: is a nuclear and cytoplasmic protein that regulates the cell cycle and promotes apoptosis. It functions by interacting with cell cycle components like CDK1 and positively regulating stress-activated JNK and p38MAPK signaling cascades. Its expression and activity could be impacted by perturbations that induce cellular stress or DNA damage, which are known to activate upstream MAPK pathways.
- Result: A) Perturbation of this drug does not impact the gene of interest.

Example 9:
- Drug: Oleic acid
- Gene: GADD45B
- Drug Description: Drug 'Oleic acid' description not found
- Gene Description: is a nuclear and cytoplasmic protein that regulates the cell cycle and promotes apoptosis. It functions by interacting with cell cycle components like CDK1 and positively regulating stress-activated JNK and p38MAPK signaling cascades. Its expression and activity could be impacted by perturbations that induce cellular stress or DNA damage, which are known to activate upstream MAPK pathways.
- Result: B) Perturbation of this drug results in differential expression of the gene of interest.

Example 10:
- Drug: Isocorydine
- Gene: GADD45B
- Drug Description: Isocorydine induces G2/M phase cell cycle arrest and apoptosis in cancer cells. It can increase the expression of cyclin B1 and p-CDK1. Perturbation with Isocorydine may affect the expression of genes that regulate the G2/M cell cycle transition and apoptosis.
- Gene Description: is a nuclear and cytoplasmic protein that regulates the cell cycle and promotes apoptosis. It functions by interacting with cell cycle components like CDK1 and positively regulating stress-activated JNK and p38MAPK signaling cascades. Its expression and activity could be impacted by perturbations that induce cellular stress or DNA damage, which are known to activate upstream MAPK pathways.
- Result: B) Perturbation of this drug results in differential expression of the gene of interest.
[End of Input]

[Start of Output]
1)
2)
3)
4)
5)
[End of Output]

\end{Verbatim}
\end{framed}

\textbf{VCWorld Model Output}
\begin{framed}
\begin{Verbatim}[fontsize=\tiny,breaklines=true,breakanywhere=true]
1) **Mechanism & Analogue Identification:**
Gemfibrozil functions as a potent agonist of the nuclear receptor Peroxisome Proliferator-Activated Receptor-alpha (PPARalpha). Its primary biological effect is the modulation of lipid metabolism, specifically increasing fatty acid oxidation and lipoprotein clearance. To identify relevant analogues within the provided Evidence Set, we look for perturbations affecting lipid metabolism or metabolic stress. **Example 9 (Oleic acid)** is the most critical analogue. While the specific description for Oleic acid is missing in the evidence set, Oleic acid is a long-chain monounsaturated fatty acid and a known endogenous ligand for PPARalpha. Biologically, both Gemfibrozil and Oleic acid converge on the PPARalpha signaling axis to regulate lipid homeostasis. Additionally, **Example 2 (AZD1390)** and **Example 7 (Ouabain)** represent perturbations that induce cellular stress (DNA damage and ion pump stress, respectively), which are mechanistically distinct but share the outcome of activating stress-response genes like GADD45B.

2) **Specificity & Relevance Analysis (Drug-Gene-Cell Triad):**
*   **Specificity:** Gemfibrozil's effect is specific to the PPARalpha pathway, whereas GADD45B is a downstream effector of cellular stress and DNA damage (regulated by p38/JNK MAPKs). While PPARalpha is primarily metabolic, its activation in cancer cells can trigger metabolic stress responses.
*   **Relevance in C32:** The C32 cell line is a human amelanotic melanoma with specific hallmark mutations: **BRAF V600E** (constitutive MAPK/ERK pathway activation), **PTEN deletion** (PI3K/AKT pathway activation), **CDKN2A mutation** (loss of p16INK4a cell cycle checkpoint), and **TERT promoter mutation**.
    *   **BRAF V600E** implies the cell is already under proliferative pressure via the ERK pathway.
    *   **CDKN2A mutation** removes the p16 checkpoint, making the cell reliant on other mechanisms (like p53/GADD45B) to regulate the cell cycle.
    *   **PTEN deletion** often leads to metabolic reprogramming and lipid accumulation.
    *   In this context, perturbing lipid metabolism via PPARalpha (Gemfibrozil) is highly relevant. Melanoma cells with high metabolic flux are sensitive to lipid perturbations, which can induce mitochondrial stress or ROS, thereby engaging the GADD45B-mediated stress response to counteract the oncogenic drive.

3) **Downstream Signaling Cascade Simulation:**
Upon perturbation by Gemfibrozil:
1.  **Activation:** Gemfibrozil binds and activates **PPARalpha**.
2.  **Metabolic Shift:** PPARalpha transcriptional activity increases the expression of genes involved in fatty acid oxidation (e.g., *CPT1A*, *ACO*).
3.  **Stress Induction:** In the context of C32 melanoma cells (which have altered lipid handling due to oncogenic signaling), this shift can lead to an accumulation of Reactive Oxygen Species (ROS) or mitochondrial uncoupling.
4.  **MAPK Activation:** ROS and metabolic stress activate the **p38 and JNK MAPK pathways**. This is distinct from the constitutive ERK activation driven by BRAF V600E.
5.  **Gene Regulation:** Activated p38/JNK phosphorylate transcription factors (e.g., ATF2, MEF2) that bind to the **GADD45B** promoter.
6.  **Outcome:** GADD45B expression increases, leading to cell cycle arrest (via CDK1 interaction) and apoptosis, serving as a counter-regulatory mechanism to the metabolic stress induced by the drug.

4) **Causal Bridge & Evidence Synthesis:**
*   **Logical Bridge:** Gemfibrozil (PPARalpha Agonist) -> Lipid Metabolism Perturbation -> Metabolic Stress/ROS -> p38/JNK MAPK Activation -> GADD45B Upregulation.
*   **Analogue Cases:** **Example 9 (Oleic acid)** provides direct support. As a fatty acid ligand for PPARalpha, Oleic acid resulted in differential expression of GADD45B ("Result: B"). This confirms that lipid metabolism perturbations in this system trigger GADD45B. **Example 7 (Ouabain)** further supports that ion/membrane stress leads to GADD45B expression ("Result: B").
*   **Contrast Cases:** **Example 1 (Methylprednisolone)** and **Example 3 (Relugolix)** resulted in no impact ("Result: A"). These drugs target steroid and GnRH receptors, respectively, which do not directly couple to metabolic stress or MAPK activation in the same way PPARalpha does. This distinction highlights that the lipid/metabolic stress axis is the critical driver for GADD45B in this context, not general nuclear receptor activation.
*   **Synthesis:** Given that Oleic acid (a direct PPARalpha ligand) causes GADD45B DE, and Gemfibrozil is a synthetic PPARalpha agonist, the mechanistic link is preserved. The C32 cell line's reliance on metabolic flexibility and its specific mutations (BRAF/PTEN) make it susceptible to the metabolic stress induced by PPARalpha activation, reinforcing the likelihood of GADD45B induction.

5) **Final Deterministic Prediction:**
Based on the strong mechanistic alignment between Gemfibrozil and Oleic acid (Example 9), the specific metabolic stress pathway triggered in C32 melanoma cells (BRAF/PTEN background), and the downstream activation of p38/JNK MAPK pathways known to regulate GADD45B, the perturbation is predicted to successfully induce differential expression of the gene.

Yes. Perturbation of Gemfibrozil results in differential expression of GADD45B.
\end{Verbatim}
\end{framed}

\textbf{{\corereasoning} Prompt}
\begin{framed}
\begin{Verbatim}[fontsize=\tiny,breaklines=true,breakanywhere=true]
[Start of Prompt]
You are {\corereasoning}, a same-gene contrastive evidence organization framework. Your task is to simulate and predict the cellular response to drug perturbations.

Goal: Determine if a perturbation of Gemfibrozil in the C32 cells cell line results in the differential expression (DE) of GADD45B.

Input Data:
- Drug (Gemfibrozil): description of drug that is to perturb the cell
- Gene (GADD45B): description of gene, the impact on which you wish to infer
- Biological Context: description of cell line in which the genes are expressed
  *(Instruction: You MUST explicitly utilize the description above AND enhance it with your internal biological knowledge regarding C32 cells's tissue of origin, hallmark mutations (e.g., KRAS, TP53 status), and signaling idiosyncrasies.)*
- Structured Evidence Packet: set of experimental observations that describe the impact of small molecule perturbations on related genes, to contextualize your answer

Reasoning Guidelines:
Do not rely on superficial text matching or gene prior alone. Perform a stepwise biological simulation as follows.

Output: Provide a structured analysis answering the following steps.

1) **Same-Gene Contrast:**
   Compare Gemfibrozil against the positive and negative support perturbations for the same fixed gene GADD45B. Explain why Gemfibrozil looks closer to one side.

2) **Specificity & Relevance Analysis (Drug-Gene-Cell Triad):**
   Analyze the potential associations between the Drug, Gene, and Cell Line:
   - **Specificity:** Is the drug's effect broad (e.g., general stress) or specific (e.g., targeted kinase inhibition)? Is the gene's expression tissue-specific?
   - **Relevance:** Given the enhanced cell line context (e.g., its mutations), is the drug's target relevant in this specific cellular environment? (e.g., Does the cell rely on the targeted pathway?)

3) **KG Path Quality Judgment:**
   Trace the signaling cascade initiated by Gemfibrozil using the provided KG path evidence. Distinguish mechanism-bearing paths from generic/noisy graph connectivity.
   *Constraint:* Ensure this simulation aligns with the enhanced context of C32 cells.

4) **Gene-Anchor Adjustment:**
   If direct same-gene evidence is weak or conflicting, use the gene-anchor evidence to explain whether biologically similar genes suggest a positive or negative response under Gemfibrozil.
   Treat GeneAnchor as mathematical smoothing, not only as narrative support: explicitly compare support_score, anchor_score, and final_score = 0.5 * support_score + 0.5 * anchor_score.

5) **Causal Bridge & Evidence Synthesis:**
   Connect the drug's downstream effect to the gene's regulatory requirements.
   - Construct a logical bridge: Drug -> Target -> Pathway -> TF -> Gene.
   - Cite specific same-gene support cases and KG paths that are decisive.
   - Explicitly mention whether the final judgment is driven more by same-gene support transfer, KG path specificity, or gene-anchor adjustment.

6) **Calibrated Final Prediction:**
   Based on the analysis above, estimate a calibrated probability that GADD45B is differentially expressed under Gemfibrozil in this cell line, then convert it into a deterministic answer.
   This probability will be evaluated as a continuous score for within-gene ranking across perturbations. Use the provided final blended score as the default probability, preserve its within-gene ordering unless the evidence gives a clear reason for a small adjustment, and avoid collapsing many cases to the same generic value.

   Before the final deterministic answer, output exactly one single-line JSON object:
   {"probability": <number from 0 to 1>, "label": "yes|no|insufficient", "default_final_score": <number from evidence or null>, "support_score": <number from evidence or null>, "anchor_score": <number from evidence or null>, "probability_adjustment": <probability - default_final_score or 0>, "decisive_evidence": "<short phrase>"}

   End your response with exactly one of the following options:
   - No. Perturbation of Gemfibrozil does not impact GADD45B.
   - Yes. Perturbation of Gemfibrozil results in differential expression of GADD45B.
   - There is insufficient evidence to determine how Perturbation of Gemfibrozil affects GADD45B.
[End of Prompt]

[Start of Input]
- Description of molecule drug (Gemfibrozil): Gemfibrozil activates the nuclear receptor peroxisome proliferator-activated receptor-alpha (PPARalpha), which alters lipid metabolism. This activation increases lipoprotein lipase activity, inhibits apolipoprotein B synthesis, and increases the clearance of apolipoprotein B. This perturbation may affect the expression of genes containing a PPARalpha response element that are involved in fatty acid oxidation and lipoprotein metabolism.
- Description of gene of interest (GADD45B): is a nuclear and cytoplasmic protein that regulates the cell cycle and promotes apoptosis. It functions by interacting with cell cycle components like CDK1 and positively regulating stress-activated JNK and p38MAPK signaling cascades. Its expression and activity could be impacted by perturbations that induce cellular stress or DNA damage, which are known to activate upstream MAPK pathways.
- Context: C32 is a human amelanotic melanoma cell line derived from skin and widely used in cancer biology and drug discovery. It harbors hallmark alterations including BRAF V600E mutation, PTEN deletion, CDKN2A mutation, and a TERT promoter mutation, reflecting key oncogenic pathways in melanoma. With a doubling time of ~53 hours, microsatellite stability (MSS), and well-defined HLA typing, C32 serves as a valuable model for immunological studies and dependency analyses in large-scale omics resources such as CCLE and DepMap. Its predominantly European genetic ancestry further contextualizes its relevance for investigating melanoma pathogenesis, signaling dependencies, and targeted therapeutic strategies.
- Structured Evidence Packet: Structured Evidence Packet:

Task:
- Predict whether gene GADD45B is differentially expressed under perturbation of Gemfibrozil in the specified cell line.

Gene prior:
- In training data, GADD45B is positive in 4/11 perturbations (rate=0.3636; majority label=0).
- Warning: gene prior alone is not sufficient; reason within the fixed gene.

KG drug-neighbor evidence:
- KG support score before gene anchors: 0.103
- Positive vote mass: 0.382
- Negative vote mass: 3.346
- Observed vote mass: 3.728
- Covered by KG directly: yes; gene-prior fallback used: no.
Top positive same-gene supports:
1. Oleic acid | KG similarity=0.146 | vote weight=0.382 | label=positive
Top negative same-gene supports:
1. Ritonavir | KG similarity=0.401 | vote weight=0.633 | label=negative
2. Clopidogrel | KG similarity=0.343 | vote weight=0.585 | label=negative
3. Methylprednisolone succinate | KG similarity=0.327 | vote weight=0.572 | label=negative

KG path evidence:
- Shortest paths are searched between grounded query endpoints; if an endpoint is not in KG, the displayed path explicitly uses the KG-support mapped drug or anchor gene endpoint. Mapped-endpoint paths are indirect evidence for the original query.
- Shortest retrieved mapped bridge length: 2
Supporting relation patterns:
1. INV::upregulates | INV::drug_drug_interaction (count=2)
2. INV::upregulates | drug_drug_interaction (count=1)
Mapped endpoint shortest paths:
Path 1:
- KG path: mRNA:GADD45B -[INV::upregulates]-> PERT:BUDESONIDE -[drug_drug_interaction]-> PERT:GEMFIBROZIL
- Endpoint mapping: Gemfibrozil -> Gemfibrozil (query_drug); GADD45B -> GADD45B (query_gene).
- Evidence scope: direct query endpoints.
- Path length: 2 hops.
- Path-length rank score: 0.500 (1 / path length; not a probability).
- Note: Precomputed audit shortest path between the directly grounded query drug and gene.

Path 2:
- KG path: mRNA:GADD45B -[INV::upregulates]-> PERT:BETAMETHASONE -[INV::drug_drug_interaction]-> PERT:GEMFIBROZIL
- Endpoint mapping: Gemfibrozil -> Gemfibrozil (query_drug); GADD45B -> GADD45B (query_gene).
- Evidence scope: direct query endpoints.
- Path length: 2 hops.
- Path-length rank score: 0.500 (1 / path length; not a probability).
- Compaction: represents 2 direction-variant paths with the same mapped endpoints and intermediate entities. Relation variants: INV::upregulates | INV::drug_drug_interaction; INV::upregulates | drug_drug_interaction.
- Note: Precomputed audit shortest path between the directly grounded query drug and gene.

Path 3:
- KG path: mRNA:GADD45B -[INV::upregulates]-> PERT:AZACITIDINE -[INV::drug_drug_interaction]-> PERT:GEMFIBROZIL
- Endpoint mapping: Gemfibrozil -> Gemfibrozil (query_drug); GADD45B -> GADD45B (query_gene).
- Evidence scope: direct query endpoints.
- Path length: 2 hops.
- Path-length rank score: 0.500 (1 / path length; not a probability).
- Compaction: represents 2 direction-variant paths with the same mapped endpoints and intermediate entities. Relation variants: INV::upregulates | INV::drug_drug_interaction; INV::upregulates | drug_drug_interaction.
- Note: Precomputed audit shortest path between the directly grounded query drug and gene.

Gene anchor evidence:
- Gene-anchor score under the query perturbation: 0.088
- Final blended score: 0.095
- Default GeneAnchor smoothing rule: final_score = 0.5 * support_score + 0.5 * anchor_score = 0.5 * 0.103 + 0.5 * 0.088 = 0.095.
- Treat final_score as the default calibrated probability unless the evidence packet gives a clear, query-specific reason to make a small adjustment.
1. GADD45B | similarity=1.000 | predicted score under current drug=0.103
2. SPATA19 | similarity=0.452 | predicted score under current drug=0.000
3. TRBV20OR9-2 | similarity=0.452 | predicted score under current drug=0.000

Contrasting fixed-gene examples:
Example 1:
- Drug: Oleic acid
- Gene: GADD45B
- Drug Description: Drug 'Oleic acid' description not found
- Gene Description: is a nuclear and cytoplasmic protein that regulates the cell cycle and promotes apoptosis. It functions by interacting with cell cycle components like CDK1 and positively regulating stress-activated JNK and p38MAPK signaling cascades. Its expression and activity could be impacted by perturbations that induce cellular stress or DNA damage, which are known to activate upstream MAPK pathways.
- Result: B) Perturbation of this drug results in differential expression of the gene of interest.

Example 2:
- Drug: Ritonavir
- Gene: GADD45B
- Drug Description: Ritonavir is an HIV protease inhibitor that prevents the cleavage of the viral gag-pol polyprotein, resulting in the formation of immature, non-infectious viral particles. In host cells, it functions as a potent inhibitor of the cytochrome P450 enzyme CYP3A4 and as an activator of the nuclear receptor NR1I2. As an activator of the nuclear receptor NR1I2, ritonavir perturbation may affect the expression of target genes involved in xenobiotic metabolism.
- Gene Description: is a nuclear and cytoplasmic protein that regulates the cell cycle and promotes apoptosis. It functions by interacting with cell cycle components like CDK1 and positively regulating stress-activated JNK and p38MAPK signaling cascades. Its expression and activity could be impacted by perturbations that induce cellular stress or DNA damage, which are known to activate upstream MAPK pathways.
- Result: A) Perturbation of this drug does not impact the gene of interest.

Example 3:
- Drug: Clopidogrel
- Gene: GADD45B
- Drug Description: Drug 'Clopidogrel' description not found
- Gene Description: is a nuclear and cytoplasmic protein that regulates the cell cycle and promotes apoptosis. It functions by interacting with cell cycle components like CDK1 and positively regulating stress-activated JNK and p38MAPK signaling cascades. Its expression and activity could be impacted by perturbations that induce cellular stress or DNA damage, which are known to activate upstream MAPK pathways.
- Result: A) Perturbation of this drug does not impact the gene of interest.

Example 4:
- Drug: Methylprednisolone succinate
- Gene: GADD45B
- Drug Description: Drug 'Methylprednisolone succinate' description not found
- Gene Description: is a nuclear and cytoplasmic protein that regulates the cell cycle and promotes apoptosis. It functions by interacting with cell cycle components like CDK1 and positively regulating stress-activated JNK and p38MAPK signaling cascades. Its expression and activity could be impacted by perturbations that induce cellular stress or DNA damage, which are known to activate upstream MAPK pathways.
- Result: A) Perturbation of this drug does not impact the gene of interest.

Instruction:
- First compare the query drug against the positive and negative same-gene support perturbations.
- When KG paths are present, judge whether they are specific mechanistic evidence or generic background connectivity.
- Do not interpret path-length rank scores as probabilities or confidence scores.
- Down-weight KG paths that rely on weak name-based drug mapping or anchor-gene substitution.
- Use gene-anchor evidence as mathematical smoothing, not only as narrative evidence: compare the support score, anchor score, and final blended score explicitly.
- Your probability will be ranked against other perturbations for the same fixed gene. Preserve the within-gene ordering implied by final_score; do not collapse most cases to generic values like 0.85 or 0.95.
- Do not rely on gene prior alone.
- Provide a continuous probability in [0, 1]. Use the same calibration scale across queries for the same gene.
- Before the final deterministic answer line, output a JSON object on one line using keys: probability, label, default_final_score, support_score, anchor_score, probability_adjustment, decisive_evidence.
[End of Input]

[Start of Output]
1)
2)
3)
4)
5)
6)
[End of Output]

\end{Verbatim}
\end{framed}

\textbf{{\corereasoning} Model Output}
\begin{framed}
\begin{Verbatim}[fontsize=\tiny,breaklines=true,breakanywhere=true]
1) **Same-Gene Contrast:**
Gemfibrozil is mechanistically closer to the positive support perturbation **Oleic acid** than the negative supports (Ritonavir, Clopidogrel, Methylprednisolone). Both Gemfibrozil and Oleic acid modulate lipid metabolism (Gemfibrozil via PPARalpha activation, Oleic acid as a fatty acid ligand for PPARs). In contrast, the negative supports target distinct pathways (HIV protease inhibition, antiplatelet aggregation, glucocorticoid receptor modulation). However, the KG vote mass heavily favors the negative side (3.346 negative vs. 0.382 positive), indicating that despite the mechanistic overlap with Oleic acid, the broader evidence base associates Gemfibrozil with a lack of differential expression for GADD45B.

2) **Specificity & Relevance Analysis (Drug-Gene-Cell Triad):**
- **Specificity:** Gemfibrozil is a specific PPARalpha agonist affecting lipid clearance. GADD45B is a stress/DNA damage response gene regulated by MAPK/JNK pathways. There is no direct mechanistic link between PPARalpha activation and GADD45B induction in the provided context.
- **Relevance:** C32 cells harbor BRAF V600E and PTEN deletion. Gemfibrozil does not directly target these oncogenic drivers. While Gemfibrozil can induce metabolic stress, C32 cells are adapted to high proliferation rates. The lack of a direct pathway dependency (e.g., Gemfibrozil -> MAPK -> GADD45B) in the C32 context weakens the relevance of a stress response.

3) **KG Path Quality Judgment:**
The retrieved KG paths are of length 2 (Gene -> Drug -> Drug). For example, `GADD45B -[INV::upregulates]-> Budesonide -[drug_drug_interaction]-> Gemfibrozil`. These paths rely on `drug_drug_interaction` relations, which represent generic connectivity rather than specific mechanistic causation. The directionality `GADD45B -> Budesonide` is biologically ambiguous (likely graph noise or indirect association). Consequently, these paths are treated as low-quality evidence supporting neither strong positive nor negative regulation, contributing minimally to the support score.

4) **Gene-Anchor Adjustment:**
- **Support Score:** 0.103 (Low).
- **Anchor Score:** 0.088 (Low).
- **Final Blended Score:** 0.095 (Calculated as 0.5 * 0.103 + 0.5 * 0.088).
- **Analysis:** The gene anchor score is very low, and the similar genes (SPATA19, TRBV20OR9-2) show 0.000 predicted scores under Gemfibrozil. This mathematical smoothing reinforces the lack of evidence. The low support score indicates that Gemfibrozil is not a strong driver for GADD45B DE in the dataset.

5) **Causal Bridge & Evidence Synthesis:**
- **Bridge:** Gemfibrozil -> PPARalpha -> Lipid Metabolism -> [Weak Link] -> Cellular Stress -> GADD45B.
- **Evidence:** The only strong positive same-gene support is Oleic acid (Lipid/PPAR), but the negative vote mass is 9x higher. The KG paths are noisy. The C32 context (BRAF/PTEN) does not amplify a PPARalpha-GADD45B link.
- **Decision:** The evidence is driven more by the **low KG support score and negative vote mass** than by same-gene support transfer. The mechanistic similarity to Oleic acid is insufficient to overcome the statistical weight of negative perturbations.

6) **Calibrated Final Prediction:**
The default final score is 0.095. Given the low support and negative bias, I will preserve this score as it accurately reflects the weak evidence and within-gene ranking.
Probability: 0.095.
Label: no.

{"probability": 0.095, "label": "no", "default_final_score": 0.095, "support_score": 0.103, "anchor_score": 0.088, "probability_adjustment": 0.0, "decisive_evidence": "Low KG support score and negative vote mass outweigh mechanistic similarity to Oleic acid"}

No. Perturbation of Gemfibrozil does not impact GADD45B.
\end{Verbatim}
\end{framed}




\end{document}